\documentclass[10pt,twocolumn,letterpaper]{article}

\usepackage{titling}
\usepackage{cvpr}
\usepackage{times}
\usepackage{epsfig}
\usepackage{graphicx}
\usepackage{amsmath}
\usepackage{amssymb}
\usepackage[keeplastbox]{flushend}
\usepackage{newtxtt}

\usepackage{visinfdefs}
\usepackage[super]{nth}
\usepackage{etoolbox,siunitx}
\robustify\bfseries
\newcommand{\myparagraph}[1]{\smallskip\noindent\textbf{#1}}

\usepackage{pifont}
\newcommand{\cmark}{\ding{51}}

\usepackage{booktabs}
\usepackage{tabularx}
\usepackage{multirow}

\usepackage[format=plain,labelformat=simple,labelsep=period,font=small,skip=4pt,compatibility=false]{caption}

\usepackage[font=scriptsize,skip=2pt]{subcaption}

\makeatletter 
\@namedef{ver@everyshi.sty}{} 
\makeatother 
\usepackage{tikz}

\usepackage[pagebackref=true,breaklinks=true,letterpaper=true,colorlinks,filecolor=red,bookmarks=false]{hyperref}
\usepackage[capitalize]{cleveref}
\crefname{section}{Sec.}{Section}

\cvprfinalcopy
\def\cvprPaperID{1830}

\usepackage{fancyhdr}
\usepackage{setspace}

\begin{document}


\title{Self-Supervised Monocular Scene Flow Estimation}
\author{Junhwa Hur \qquad\qquad Stefan Roth \\ Department of Computer Science, TU Darmstadt}


{
\twocolumn[{
\renewcommand\twocolumn[1][]{#1}
\maketitle
\thispagestyle{fancy} 
\begin{center}
\centering  \includegraphics[width=0.99\linewidth]{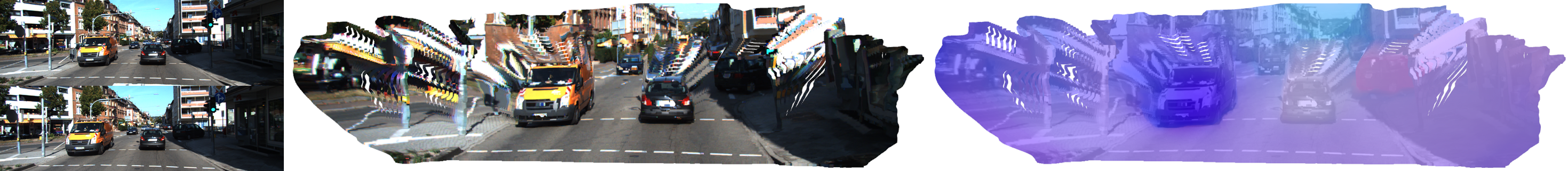}\\[-0.5mm]
\label{fig:intro-teaser}
\captionof{figure}{\textbf{Results of our monocular scene flow approach on the KITTI dataset \cite{Geiger:2012:AWR}}. Given two consecutive images \emph{(left)}, our method jointly predicts depth \emph{(middle)} and scene flow \emph{(right)}. ($x$,$z$)-coordinates of 3D scene flow are visualized using an optical flow color coding.\\
}
\end{center}
}]
}


\begin{abstract}
Scene flow estimation has been receiving increasing attention for 3D environment perception.
Monocular scene flow estimation -- obtaining 3D structure and 3D motion from two temporally consecutive images -- is a highly ill-posed problem, and practical solutions are lacking to date.
We propose a novel monocular scene flow method that yields competitive accuracy \emph{and} real-time performance.
By taking an inverse problem view, we design a single convolutional neural network (CNN) that successfully estimates depth and 3D motion simultaneously from a classical optical flow cost volume.
We adopt self-supervised learning with 3D loss functions and occlusion reasoning to leverage unlabeled data.
We validate our design choices, including the proxy loss and augmentation setup.
Our model achieves state-of-the-art accuracy among unsupervised/self-supervised learning approaches to monocular scene flow, and yields competitive results for the optical flow and monocular depth estimation sub-tasks.
Semi-supervised fine-tuning further improves the accuracy and yields promising results in real-time.

\end{abstract}

\section{Introduction}
\label{sec:introduction}

Scene flow estimation is the task of obtaining $3$D structure and $3$D motion of dynamic scenes, which is crucial to environment perception, \eg, in the context of autonomous navigation.
Consequently, many scene flow approaches have been proposed recently, based on different types of input data, such as stereo images \cite{Huguet:2007:AVM,Schuster:2018:SFF,Vogel:2013:PRS,Wedel:2011:3SF,Zhang:2001:O3S}, 3D point clouds \cite{Gu:2019:HPL,Liu:2019:FN3}, or a sequence of RGB-D images \cite{Hadfield:2011:KDP,Hornacek:2014:SF6,Lv:2018:LRD,Qiao:2018:SFN,Quiroga:2014:DSR,Thakur:2018:SED}.
However, each sensor configuration has its own limitations, \eg requiring stereo calibration for a stereo rig, expensive sensing devices (\eg, LiDAR) for measuring 3D points, or being limited to indoor usage (\ie, RGB-D camera).
We here consider \emph{monocular 3D scene flow estimation}, aiming to overcome these limitations.

Monocular scene flow estimation, however, is a highly ill-posed problem since both monocular depth (also called single-view depth) and per-pixel 3D motion need to be estimated from consecutive monocular frames, here two consecutive frames.
Comparatively few approaches have been suggested so far \cite{Brickwedde:2019:MSF,Xiao:2017:MSF}, none of which achieves both reasonable accuracy and real-time performance.

Recently, a number of CNN approaches \cite{Chen:2019:SSL,Liu:2019:ULS,Luo:2019:EPC,Ranjan:2019:CCJ,Yang:2018:EPC,Zou:2018:DFN} have been proposed to jointly estimate depth, flow, and camera ego-motion in a monocular setup.
This makes it possible to recover 3D motion from the various outputs, however with important limitations.
The depth--scale ambiguity \cite{Ranjan:2019:CCJ,Zou:2018:DFN} and the impossibility of estimating depth in occluded regions \cite{Chen:2019:SSL,Liu:2019:ULS,Luo:2019:EPC,Yang:2018:EPC} significantly limit the ability to obtain accurate 3D scene flow across the entire image.

In this paper, we propose a monocular scene flow approach that yields competitive accuracy \emph{and} real-time performance by exploiting CNNs.
To the best of our knowledge, our method is the first monocular scene flow method that directly predicts 3D scene flow from a CNN. 
Due to the scarcity of 3D motion ground truth and the domain over-fitting problem when using synthetic datasets \cite{Butler:2012:NOS,Mayer:2016:ALD}, we train directly on the target domain in a self-supervised manner to leverage large amounts of unlabeled data.
Optional semi-supervised fine-tuning on limited quantities of ground-truth data can further boost the accuracy.

We make three main technical contributions:
\emph{(i)} We propose to approach this ill-posed problem by taking an \emph{inverse problem view}.
Noting that optical flow is the 2D projection of a 3D point and its 3D scene flow, we take the inverse direction and estimate scene flow in the monocular setting by \emph{decomposing a classical optical flow cost volume into scene flow and depth} using a \emph{single joint decoder}.
We use a standard optical flow pipeline (PWC-Net \cite{Sun:2018:PWC}) as basis and adapt it for monocular scene flow.
We verify our architectural choice and motivation by comparing with multi-task CNN approaches.
\emph{(ii)} We demonstrate that solving the monocular scene flow task with a single joint decoder actually \emph{simplifies} joint depth and flow estimation methods \cite{Chen:2019:SSL,Liu:2019:ULS,Luo:2019:EPC,Ranjan:2019:CCJ,Yang:2018:EPC,Zou:2018:DFN}, and yields competitive accuracy despite a simpler network.
Existing multi-task CNN methods have multiple modules for the various tasks and often require complex training schedules due to the instability of training multiple CNNs jointly.
In contrast, our method only uses a single network that outputs scene flow and depth (as well as optical flow after projecting to 2D) with a \emph{simpler training setup} and better accuracy for depth and scene flow.
\emph{(iii)} We introduce a \emph{self-supervised loss function} for monocular scene flow as well as a suitable \emph{data augmentation scheme}.
We introduce a view synthesis loss, a $3$D reconstruction loss, and an occlusion-aware loss, all validated in an ablation study.
Interestingly, we find that the geometric augmentations of the two tasks conflict one another and determine a suitable compromise using an ablation study.

After training on unlabeled data from the KITTI raw dataset \cite{Geiger:2013:VMR}, we evaluate on the KITTI Scene Flow dataset \cite{Menze:2015:J3E,Menze:2018:OSF} and demonstrate highly competitive accuracy compared to previous unsupervised/self-supervised learning approaches to monocular scene flow \cite{Luo:2019:EPC,Yang:2018:EPC,Yin:2018:GNU}, increasing the accuracy by 34.0\%.
The accuracy of our fine-tuned network moves even closer to that of the semi-supervised method of \cite{Brickwedde:2019:MSF}, while being orders of magnitude faster.
\section{Related Work}
\label{sec:relatedwork}

\paragraph{Scene flow.}
Scene flow is commonly defined as a dense 3D motion field for each point in the scene, and was first introduced by Vedula \etal~\cite{Vedula:1999:TDS,Vedula:2005:TDS}. 
The most common setup is to jointly estimate 3D scene structure and 3D motion of each point given a sequence of \emph{stereo} images \cite{Huguet:2007:AVM,Schuster:2018:SFF,Vogel:2014:VC3,Vogel:2013:PRS,Vogel:2015:3SF,Wedel:2011:3SF,Zhang:2001:O3S}.
Early approaches were mostly based on standard variational formulations and energy minimization, yielding limited accuracy and incurring long runtime \cite{Basha:2013:MVS,Huguet:2007:AVM,Vogel:2011:3SF,Wedel:2011:3SF,Zhang:2001:O3S}.
Later, Vogel \etal~\cite{Vogel:2014:VC3,Vogel:2013:PRS,Vogel:2015:3SF} introduced an explicit piecewise planar surface representation with a rigid motion model, which brought significant accuracy improvements especially in traffic scenarios. 
Exploiting semantic knowledge by means of rigidly moving objects yielded further accuracy boosts \cite{Behl:2017:BBS,Ma:2019:DRI,Menze:2015:OSF,Ren:2017:CSF}.

Recently, CNN models have been introduced as well.
Supervised approaches \cite{Ilg:2018:OMD,Jiang:2019:SAS,Mayer:2016:ALD,Saxena:2019:PDO} rely on large synthetic datasets and limited in-domain data to achieve state-of-the-art accuracy with real-time performance.
Un-/self-supervised learning approaches \cite{Lee:2019:LRF,Liu:2019:ULS,Wang:2019:UOS} have been developed to circumvent the difficulty of obtaining ground-truth data, but their accuracy has remained behind.

Another category of approaches estimates scene flow from a sequence of RGB-D images \cite{Hadfield:2011:KDP,Hornacek:2014:SF6,Lv:2018:LRD,Qiao:2018:SFN,Quiroga:2014:DSR,Thakur:2018:SED} or 3D points clouds \cite{Gu:2019:HPL,Liu:2019:FN3},  exploiting the given 3D structure cues.
In contrast, our approach is based on a more challenging setup that jointly estimates 3D scene structure and 3D scene flow from a \emph{sequence of monocular images}.

\myparagraph{Monocular scene flow.}
Xiao \etal~\cite{Xiao:2017:MSF} introduced a variational approach to monocular scene flow given an initial depth cue, but without competitive accuracy.
Brickwedde \etal~\cite{Brickwedde:2019:MSF} proposed an integrated pipeline by combining CNNs and an energy-based formulation.
Given depth estimates from a monocular depth CNN, trained on pseudo-labeled data, the method jointly estimates 3D plane parameters and the 6D rigid motion of a piecewise rigid scene representation, achieving state-of-the-art accuracy.
In contrast to \cite{Brickwedde:2019:MSF}, our approach is purely CNN-based, runs in real-time, and is trained in an end-to-end self-supervised manner, which allows to exploit a large amount of unlabeled data (\cf~\cite{Xiao:2017:MSF}).

\myparagraph{Joint estimation of optical flow and depth.}
Given two depth maps and optical flow between two temporally consecutive frames, 3D scene flow can be simply calculated \cite{Schuster:2018:CSD} by relating two 3D points from optical flow. 
However, this pipeline has a critical limitation; it cannot estimate the 3D motion for occluded pixels since their depth value in the second frame is not known.
Several recent methods \cite{Chen:2019:SSL,Lai:2019:BSM,Ranjan:2019:CCJ,Yang:2018:EPC,Yin:2018:GNU,Zhu:2019:RMD,Zou:2018:DFN} utilized multi-task CNN models to jointly estimate depth, optical flow, camera motion, and moving object masks from a monocular sequence in an unsupervised/self-supervised setting.
While it may be possible to reconstruct scene flow from their outputs, these methods \cite{Luo:2019:EPC,Yang:2018:EPC} yield limited scene flow accuracy due to being limited to non-occluded regions.
In contrast, our method directly estimates 3D scene flow with a CNN so that we naturally bypass this problem.


\begin{figure}[t]
\centering
\subcaptionbox{Projecting scene flow into 2D space.\label{fig:proj_sf}}{\includegraphics[width=0.8\linewidth]{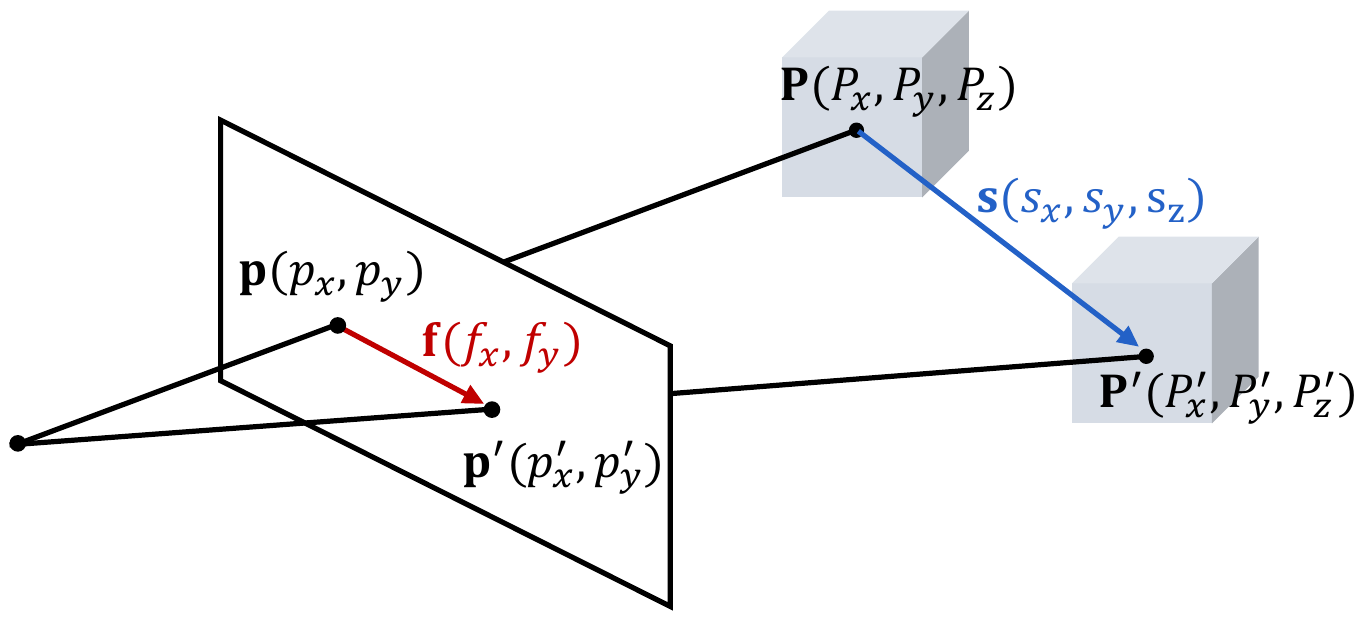}} \\[1mm]
\subcaptionbox{Back-projecting optical flow into 3D space.\label{fig:backproj_op}}{\includegraphics[width=0.8\linewidth]{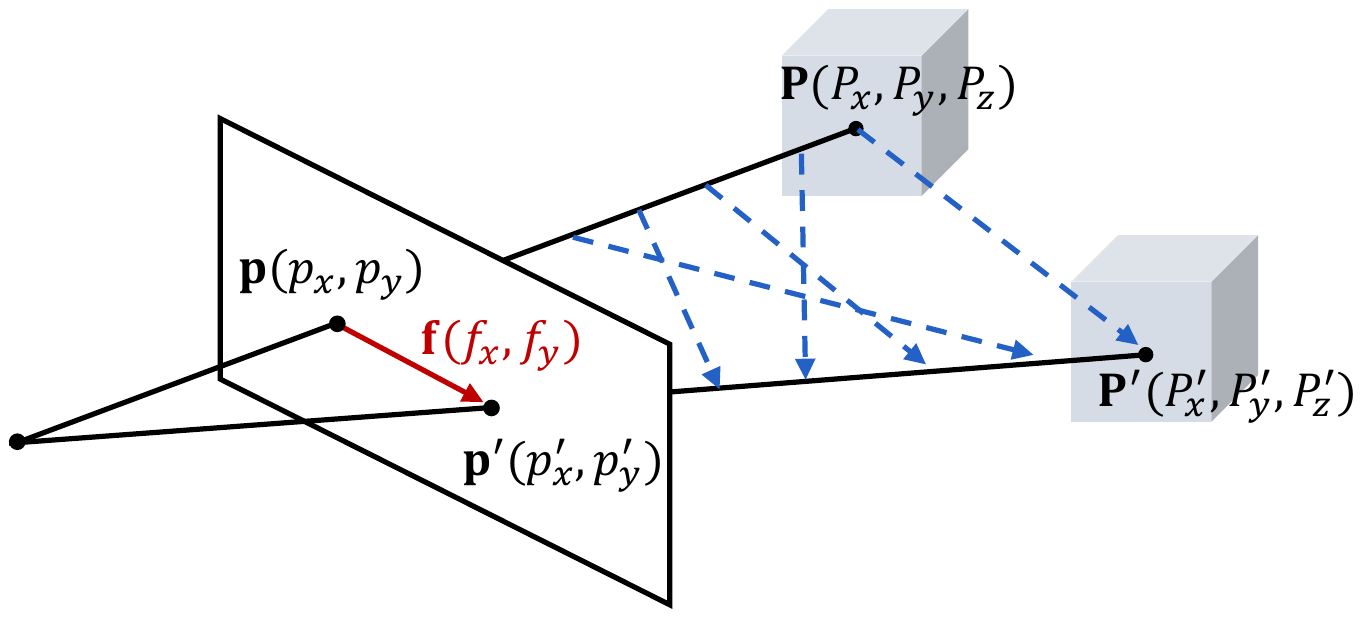}}
\caption{\textbf{Relating monocular scene flow estimation to optical flow}: \emph{(a)} Projection of scene flow into the image plane yields optical flow \cite{Yan:2016:SFE}. \emph{(b)} Back-projection of optical flow leaves an ambiguity in jointly determining depth and scene flow.}
\label{fig:concept_mnsf}
\vspace{-0.5em}
\end{figure}

\section{Self-Supervised Monocular Scene Flow}
\label{sec:mono_sf_estimation}

\subsection{Problem formulation}
\label{subsec:problem_formulation}
For each pixel $\mv{p}=(p_x, p_y)$ in the reference frame $\mv{I}_t$, our main objective is to estimate the corresponding 3D point $\mv{P}=(P_x, P_y, P_z)$ and its (forward) scene flow $\mv{s}=(s_x, s_y, s_z)$ to the target frame $\mv{I}_{t+1}$, as illustrated in \cref{fig:proj_sf}.
The scene flow is defined as 3D motion with respect to the camera, and its projection onto the image plane becomes the optical flow $\mv{f}=(f_x, f_y)$.

To estimate scene flow in the monocular camera setting, we take an inverse problem approach: we use CNNs to estimate a classical \emph{optical flow cost volume} as intermediate representation, which is then \emph{decomposed} with a \emph{learned decoder} into 3D points and their scene flow.
Unlike scene flow with a stereo camera setup \cite{Lai:2019:BSM,Lee:2019:LRF,Wang:2019:UOS}, it is challenging to determine depth on an absolute scale due to the scale ambiguity.
Yet, relating per-pixel correspondence between two images can provide a cue for estimating depth in the monocular setting.
Also, given an optical flow estimate, back-projecting optical flow into 3D yields many possible combinations of depth and scene flow, see \cref{fig:backproj_op}, which makes the problem much more challenging.

\begin{figure*}
\centering	\includegraphics[width=0.9\linewidth]{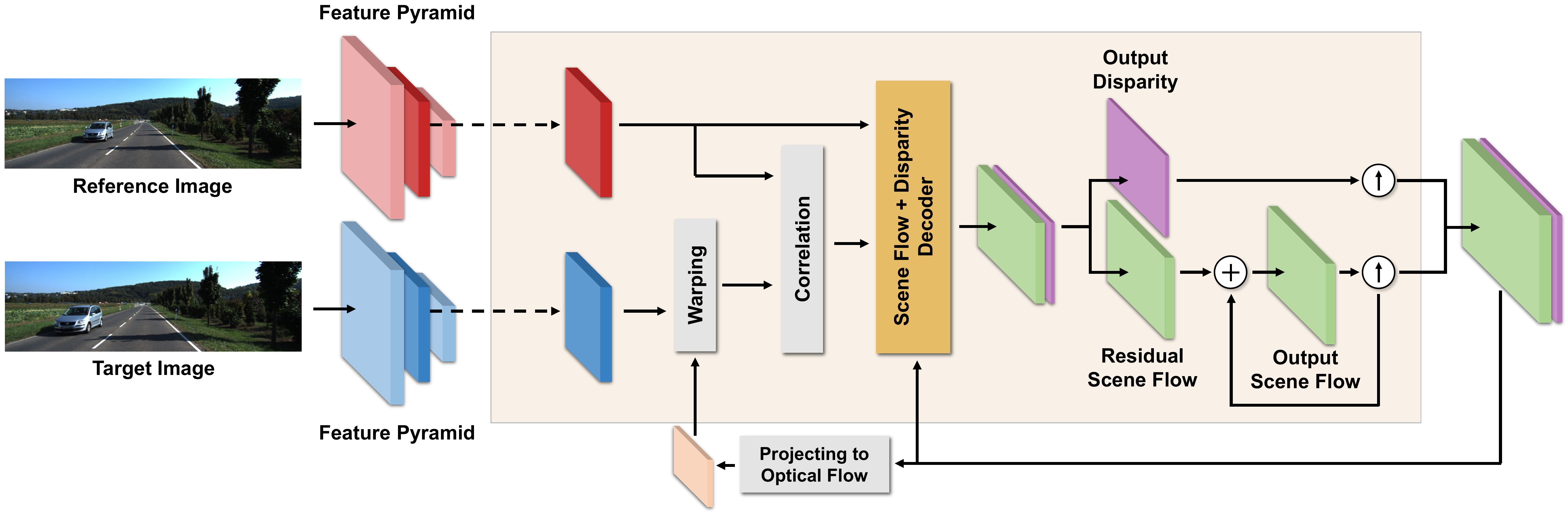}
\caption{\textbf{Our monocular scene flow architecture based on PWC-Net \cite{Sun:2018:PWC}}: 
while maintaining the overall original structure of PWC-Net, we modify the decoder to output \emph{residual scene flow} and (non-residual) \emph{disparity} together. 
After the residual update of scene flow, we project the scene flow back to optical flow using depth. 
Then, the optical flow is used for warping the feature map (only 3 of 7 levels shown for ease of visualization) in the next pyramid level. 
The light-yellow shaded region shows one forward pass for each pyramid level.}
\label{fig:network_architecture}
\vspace{-0.5em}
\end{figure*}

\subsection{Network architecture}
\label{subsec:network_architecture}

In contrast to previous work \cite{Chen:2019:SSL,Luo:2019:EPC,Ranjan:2019:CCJ,Yang:2018:EPC,Yin:2018:GNU,Zou:2018:DFN} that uses separate networks for each task (\eg,~optical flow, depth, and camera motion), our method only uses one single CNN model that outputs both 3D scene flow and disparity\footnote{Even though we do not have stereo images at test time, we still estimate disparity of a hypothetical stereo setup following \cite{Godard:2019:DIS,Godard:2017:UMD}, which can be converted into depth given the assumed stereo configuration.} through a \emph{single decoder}.
We argue that having a single decoder is more sensible in our monocular setting than separate decoders, because when decomposing evidence for 2D correspondence into 3D structure and 3D motion, their interplay need to be taken into account (\cf \cref{fig:backproj_op}). 

The first technical basis of our CNN model is PWC-Net \cite{Sun:2018:PWC}, one of the state-of-the-art optical flow networks, which we modify for our task.
\cref{fig:network_architecture} illustrates our monocular scene flow architecture atop PWC-Net.
PWC-Net has a pyramidal structure that constructs a feature pyramid and incrementally updates the estimation across the pyramid levels.
The yellow-shaded area shows one forward pass for each pyramid level.

While maintaining the original structure, we modify the decoder of each pyramid level to output disparity and scene flow together by increasing the number of output channels from $2$ to $4$ (\ie, $3$ for scene flow and $1$ for disparity).
Following the benefit of residual motion estimation in the context of optical flow \cite{Hui:2018:LFN,Hur:2019:IRR,Sun:2018:PWC}, we estimate residual scene flow at each level.
In contrast, we observe that residual updates hurt disparity estimation, hence we estimate (non-residual) disparity at all levels.
To have more discriminate features, we increase the number of feature channels in the pyramidal feature extractor from $[16, 32, 64, 96, 128, 196]$ to $[32, 64, 96, 128, 192, 256]$.

\subsection{Addressing the scale ambiguity}
\label{subsec:scale_ambiguity}
When resolving the 3D ambiguities, it is not possible to determine the depth scale from a single correspondence in two monocular images.
In order to estimate depth and scene flow on an \emph{absolute scale}, we adopt the monocular depth estimation approach of Godard~\etal \cite{Godard:2019:DIS,Godard:2017:UMD} as our second basis, which utilizes pairs of stereo images with their known stereo configuration and camera intrinsics $\mm{K}$ for training; at test time, only monocular images and known intrinsics are needed.
The images from the right camera guide the CNN to estimate the disparity $d$ on an absolute scale by exploiting semantic and geometric cues indirectly \cite{Dijk:2019:DON} through a self-supervised loss function.
Then the depth $\hat{d}$ can be trivially recovered given the baseline distance of a stereo rig $b$ and the camera focal length $f_{\text{focal}}$ as $\hat{d} = b\cdot f_\text{focal}/d$.
We also use stereo images only for training; at test time our approach is \emph{purely monocular}.
In our context, estimating depth on an absolute scale helps to disambiguate scene flow on an absolute scale as well (\cf \cref{fig:backproj_op}).
Moreover, tightly coupling temporal correspondence and depth actually helps to identify the appropriate absolute scale, which allows us to avoid unrealistic testing settings that other monocular methods rely on (\eg, \cite{Ranjan:2019:CCJ,Yin:2018:GNU,Zou:2018:DFN} use \emph{ground truth} to correctly scale their predictions at test time).

\subsection{A proxy loss for self-supervised learning} 
\label{subsec:proxy_loss}
Similar to previous monocular structure reconstruction methods \cite{Chen:2019:SSL,Luo:2019:EPC,Ranjan:2019:CCJ,Yang:2018:EPC,Yin:2018:GNU,Zhu:2019:RMD,Zou:2018:DFN}, we exploit a view synthesis loss to guide the network to jointly estimate disparity and scene flow.
For better accuracy in both tasks, we exploit occlusion cues through bi-directional estimation \cite{Meister:2018:ULO}, here of disparity and scene flow.
Given a stereo image pair of the reference and target frame \{$\mv{I}^\text{l}_t$, $\mv{I}^\text{l}_{t+1}$, $\mv{I}^\text{r}_t$, $\mv{I}^\text{r}_{t+1}$\}, we input a monocular sequence from the left camera ($\mv{I}^\text{l}_t$ and $\mv{I}^\text{l}_{t+1}$) to the network and obtain a disparity map of each frame ($d^\text{l}_t$ and $d^\text{l}_{t+1}$) as well as forward and backward scene flow ($\mv{s}^\text{l}_\text{fw}$ and $\mv{s}^\text{l}_\text{bw}$) by simply switching the temporal order of the input.
The two images from the right camera ($\mv{I}^\text{r}_t$ and $\mv{I}^\text{r}_{t+1}$) are used only as a guidance in the loss function and are not used at test time.
Our total loss is a weighted sum of a disparity loss $L_\text{d}$ and a scene flow loss $L_\text{sf}$,
\begin{equation}
L_\text{total} = L_\text{d} + \lambda_\text{sf} L_\text{sf}.
\label{eq:total_loss}
\end{equation}

\myparagraph{Disparity loss.} 
Based on the approach of Godard \etal~\cite{Godard:2019:DIS,Godard:2017:UMD}, we propose an occlusion-aware monocular disparity loss, consisting of a photometric loss $L_\text{d\_ph}$ and a smoothness loss $L_\text{d\_sm}$,
\begin{equation}
L_\text{d} = L_\text{d\_ph} + \lambda_\text{d\_sm} L_\text{d\_sm},
\label{eq:disp_loss}
\end{equation}
with regularization parameter $\lambda_\text{d\_sm}=0.1$.
The disparity loss is applied to both disparity maps $d^\text{l}_t$ and $d^\text{l}_{t+1}$. For brevity, we only describes the case of $d^\text{l}_t$.

The \emph{photometric loss} $L_\text{d\_ph}$ penalizes the photometric difference between the left image $\mv{I}^\text{l}_{t}$ and the reconstructed left image $\mv{\tilde{I}}^{\text{l,d}}_{t}$, which is synthesized from the output disparity map $d^\text{l}_{t}$ and the given right image $\mv{I}^\text{r}_{t}$ using bilinear interpolation \cite{Jaderberg:2015:STN}.
Different to \cite{Godard:2019:DIS,Godard:2017:UMD}, we only penalize the photometric loss for non-occluded pixels. 
Following standard practice \cite{Godard:2019:DIS,Godard:2017:UMD}, we use a weighted combination of an $L_1$ loss and the structural similarity index (SSIM) \cite{Wang:2004:IQA}:
\begin{subequations}
\begin{equation}
L_\text{d\_ph} = \frac{\sum_\mv{p} \big(1 - O^\text{l,disp}_t(\mv{p}) \big) \cdot \rho\big(\mv{I}^\text{l}_{t}(\mv{p}), \mv{\tilde{I}}^{\text{l,d}}_{t}(\mv{p})\big)}{\sum_{\mv{q}} \big(1 - O^\text{l,disp}_t(\mv{q})\big)} 
\label{eq:disp_photometric_diff}
\end{equation}
with 
\begin{equation}
\rho(a, b) = \alpha \frac{1-\text{SSIM}(a, b)}{2} + (1-\alpha) {\lVert}a - b{\rVert}_1,
\label{eq:photometric_loss}
\end{equation}%
\label{eq:disp_photometric_loss}%
\end{subequations}%
where $\alpha = 0.85$ and $O^\text{l,disp}_t$ is the disparity occlusion mask ($0$ -- visible, $1$ -- occluded).
To obtain the occlusion mask $O^\text{l,disp}_t$, we feed the right image $\mv{I}^\text{r}_{t}$ into the network to obtain the right disparity $d^\text{r}_{t}$ and take the inverse of its disocclusion map, which is obtained by forward-warping the right disparity map \cite{Hur:2017:MFE,Wang:2018:OAU}.

To encourage locally smooth disparity estimates, we adopt an \emph{edge-aware $2^\text{nd}$-order smoothness} \cite{Liu:2019:ULS,Meister:2018:ULO,Woodford:2008:GSR},
\begin{equation}
L_\text{d\_sm} = \frac{1}{N} \sum_\mv{p} \sum_{i \in \{x, y\}} {\big\lvert} \nabla^2_i d^\text{l}_{t}(\mv{p}) {\big\rvert} \cdot e^{-\beta {\lVert} \nabla_i \mv{I}^\text{l}_{t}(\mv{p}) {\rVert}_1 },
\label{eq:disp_smooth_loss}
\end{equation}
with $\beta=10$ and $N$ being the number of pixels. 

\myparagraph{Scene flow loss.} 
The scene flow loss consists of three terms -- a photometric loss $L_\text{sf\_ph}$, a 3D point reconstruction loss $L_\text{sf\_pt}$, and a scene flow smoothness loss $L_\text{sf\_sm}$,
\begin{equation}
L_\text{sf} = L_\text{sf\_ph} + 
\lambda_\text{sf\_pt} L_\text{sf\_pt} + \lambda_\text{sf\_sm} L_\text{sf\_sm},
\label{eq:sf_loss}
\end{equation}
with regularization parameters $\lambda_\text{sf\_pt}=0.2$ and $\lambda_\text{sf\_sm}=200$.
The scene flow loss is applied to both forward and backward scene flow ($\mv{s}^\text{l}_{\text{fw}}$ and $\mv{s}^\text{l}_{\text{bw}}$). 
Again for brevity, we only describe the case of forward scene flow $\mv{s}^\text{l}_{\text{fw}}$.

\begin{figure}[t]
\centering
\subcaptionbox{Photometric loss.\label{fig:loss_sf_ph}}{\includegraphics[width=0.4\linewidth]{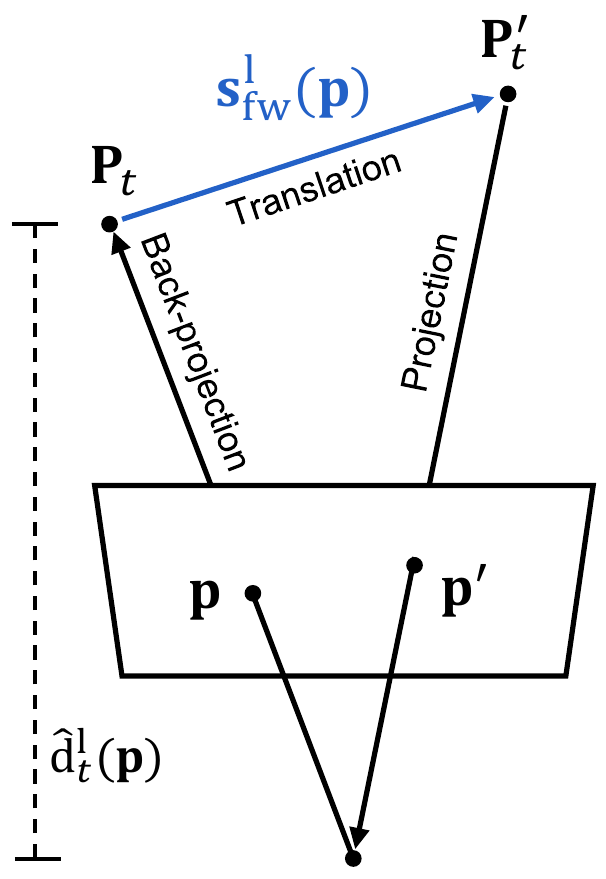}} \quad \quad
\subcaptionbox{3D point reconstruction loss.\label{fig:loss_sf_pt}}{\includegraphics[width=0.41\linewidth]{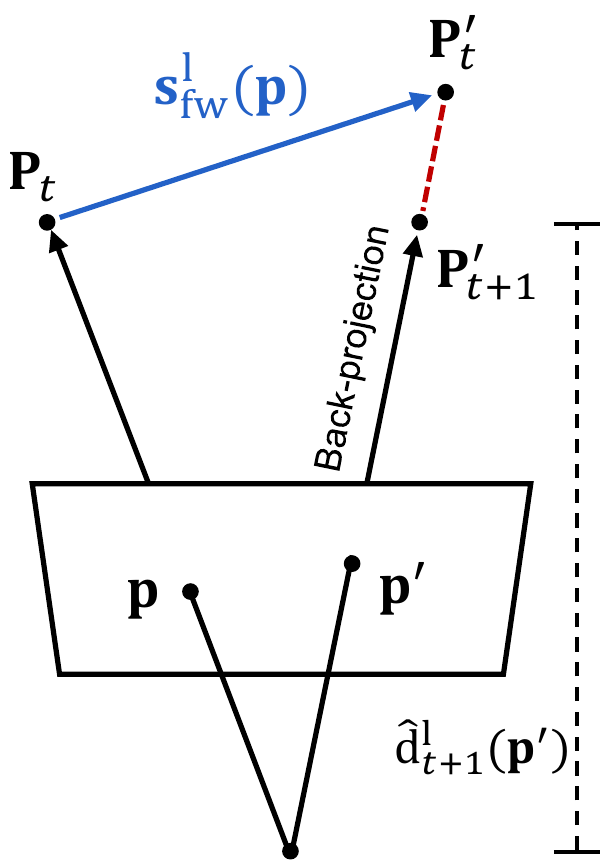}}
\caption{\textbf{Scene flow losses}: \emph{(a)} Finding corresponding pixels given depth and scene flow for the photometric loss $L_\text{sf\_ph}$ (\Eq~\ref{eq:sf_photometric_diff}). \emph{(b)} Penalizing 3D distance (dashed, red) between corresponding 3D points by the point reconstruction loss $L_\text{sf\_pt}$ (\Eq~\ref{eq:sf_3d_dist}).}
\label{fig:loss_sf}
\vspace{-0.5em}
\end{figure}

The \emph{scene flow photometric loss} $L_\text{sf\_ph}$ penalizes the photometric difference between the reference image $\mv{I}^\text{l}_{t}$ and the reconstructed reference image $\mv{\tilde{I}}^{\text{l,sf}}_{t}$, synthesized from the disparity map $d^\text{l}_{t}$, the output scene flow $\mv{s}^\text{l}_\text{fw}$, and the target image $\mv{I}^{\text{l}}_{t+1}$.
To reconstruct the image, the corresponding pixel coordinate $\mv{p'}$ in $\mv{I}^{\text{l}}_{t+1}$ of each pixel $\mv{p}$ in $\mv{I}^{\text{l}}_{t}$ is calculated by back-projecting the pixel $\mv{p}$ into 3D space using the camera intrinsics $\mm{K}$ and estimated depth $\hat{d}^\text{l}_{t}(\mv{p})$, translating the points using the scene flow $\mv{s}^\text{l}_\text{fw}(\mv{p})$, and then re-projecting them to the image plane (\cf \cref{fig:loss_sf_ph}),
\begin{equation}
\mv{p'} = \mm{K} \Big( {\hat{d}}_t^\text{l}(\mv{p}) \cdot \mm{K^{-1}} \mv{p} + \mv{s}^\text{l}_\text{fw}(\mv{p}) \Big),
\label{eq:sf_corresp_point}
\end{equation}
assuming a homogeneous coordinate representation. 
Then, we apply the same occlusion-aware photometric loss as in the disparity case (Eq.~\ref{eq:disp_photometric_diff}),
\begin{equation}
L_\text{sf\_ph} = \frac{\sum_\mv{p} \big(1 - O^\text{l,sf}_t(\mv{p}) \big) \cdot \rho\big(\mv{I}^\text{l}_{t}(\mv{p}), \mv{\tilde{I}}^{\text{l,sf}}_{t}(\mv{p})\big)}{\sum_{\mv{q}} \big(1 - O^\text{l,sf}_t(\mv{q})\big)},
\label{eq:sf_photometric_diff}
\end{equation}
where $O^\text{l,sf}_t$ is the scene flow occlusion mask, obtained by calculating disocclusion using the backward scene flow $\mv{s}^\text{l}_\text{bw}$.

Additionally, we also penalize the Euclidean distance between the two corresponding 3D points, \ie the translated 3D point of pixel $\mv{p}$ from the reference frame and the matched 3D point in the target frame (\cf \cref{fig:loss_sf_pt}):
\begin{subequations}
\label{eq:sf_3d_dist}
\begin{equation}
L_\text{sf\_pt} = \frac{\sum_\mv{p} \big(1 - O^\text{l,sf}_t(\mv{p}) \big) \cdot {\big\lVert} \mv{P}'_t - \mv{P}'_{t+1} {\big\rVert}_2 }{\sum_{\mv{q}} \big(1 - O^\text{l,sf}_t(\mv{q})\big)},
\end{equation}
with 
\begin{eqnarray}
\mv{P}'_t &=& {\hat{d}}_t^\text{l}(\mv{p}) \cdot \mm{K^{-1}} \mv{p} + \mv{s}^\text{l}_\text{fw}(\mv{p})\\
\mv{P}'_{t+1} &=& {\hat{d}}_{t+1}^\text{l}(\mv{p'}) \cdot \mm{K^{-1}} \mv{p'},
\end{eqnarray}
\end{subequations}
and $\mv{p}'$ as defined in \cref{eq:sf_corresp_point}.
Again, this \emph{3D point reconstruction loss} is only applied on visible pixels, where the correspondence should hold.

Analogous to the disparity loss in \cref{eq:disp_smooth_loss}, we also adopt \emph{edge-aware $2^\text{nd}$-order smoothness for scene flow} to encourage locally smooth estimation:
\begin{equation}
{L_\text{sf\_sm} = \frac{1}{N} \sum_\mv{p} \sum_{i \in \{x, y\}} {\big\lvert} \nabla^2_i \mv{s}^\text{l}_{\text{fw}}(\mv{p}) {\big\rvert} \cdot e^{-\beta {\lVert} \nabla_i \mv{I}^\text{l}_{t}(\mv{p}) {\rVert}_1 } }.
\end{equation}

\subsection{Data augmentation}
\label{sec:data_augmentation}

In many prediction tasks, data augmentation is crucial to achieving good accuracy given limited training data.
In our monocular scene flow task, unfortunately, the typical geometric augmentation schemes of the two tasks (\ie, monocular depth estimation, scene flow estimation) conflict each other.
For monocular depth estimation, not performing geometric augmentation is desirable as it enables learning the scene layout under a fixed camera configuration \cite{Dijk:2019:DON,Hu:2019:VCN}.
On the other hand, the scene flow necessitates geometric augmentations to match corresponding pixels better \cite{Jiang:2019:SAS,Mayer:2016:ALD}.

We investigate which type of (geometric) augmentation is suitable for our monocular scene flow task and method.
Similar to previous multi-task approaches \cite{Chen:2019:SSL,Ranjan:2019:CCJ,Zou:2018:DFN}, we prepare a simple data augmentation scheme, consisting of random scales, cropping, resizing, and horizontal image flipping.
Upon the augmentation, we also explore the recent CAM-Convs \cite{Facil:2019:CCC}, which facilitate depth estimation irrespective of the  camera intrinsics. 
After applying augmentations on the input images, we calculate the resulting camera intrinsics and then input them in the format of CAM-Convs (see \cite{Facil:2019:CCC} for technical details).
We conjecture that using geometric augmentation will improve the scene flow accuracy.
Yet, at the same time adopting CAM-Convs \cite{Facil:2019:CCC} could prevent the depth accuracy from dropping due to the changes in camera intrinsics of the augmented images.
We conduct our empirical study on the \emph{KITTI split} \cite{Godard:2017:UMD} of the KITTI raw dataset \cite{Geiger:2013:VMR} (see \cref{subsec:implementation_details} for details).

\begin{table}[t]
\centering
\footnotesize
\setlength\tabcolsep{3.75pt}
\begin{tabular*}{\columnwidth}{@{\extracolsep{\fill}}cc@{\hskip 1.5em}S[table-format=1.3,round-precision=3]S[table-format=1.3,round-precision=3]S[table-format=2.2]S[table-format=2.2]S[table-format=2.2]S[table-format=2.2]@{}}
	\toprule
	& & \multicolumn{2}{c}{Monocular depth} & \multicolumn{4}{c}{Monocular scene flow}  \\\cmidrule(lr){3-4}\cmidrule(l){5-8}
	{Aug.} & {CC. \cite{Facil:2019:CCC}} & {Abs. Rel.} & {Sq. Rel.} & {D$1$-all} & {D$2$-all} & {F$1$-all} & {SF$1$-all}  \\
	\midrule
	       &         & 0.113 & 1.118 & 32.06 & 36.46 & 24.68 & 49.89 \\
	\cmark &         & 0.122 & 1.172 & \bfseries 31.25 & \bfseries 34.86 & \bfseries 23.49 & \bfseries 47.05 \\
	       & \cmark  & \bfseries 0.112 & \bfseries 1.089 & 37.24 & 39.26 & 24.82 & 54.83 \\
	\cmark & \cmark  & 0.121 & 1.155 & 33.25 & 36.21 & 24.73 & 49.12 \\
	\bottomrule
\end{tabular*}
\caption{\textbf{Impact of geometric augmentations (\emph{Aug.}) and CAM-Convs (\emph{CC.}) \cite{Facil:2019:CCC} on monocular depth and scene flow estimation} (on KITTI split, see text): the accuracy of monocular depth estimation improves only when using CAM-Convs while that of monocular scene flow estimation improves when only using augmentation without CAM-Convs.}
\label{table:augmentation_study}
\vspace{-0.5em}
\end{table}

\myparagraph{Empirical study for monocular depth estimation.}
We use a ResNet18-based monocular depth baseline \cite{Godard:2017:UMD} using our proposed occlusion-aware loss.
\Cref{table:augmentation_study} (left hand side) shows the results.
As we can see, geometric augmentations deteriorate the depth accuracy, since they prevent the network from learning a specific camera prior by inputting augmented images with diverse camera intrinsics; this observation holds with and without CAM-Convs.
This likely explains why some multi-task approaches \cite{Lai:2019:BSM, Lee:2019:LRF, Liu:2019:ULS, Wang:2019:UOS} only use minimal augmentation schemes such as image flipping and input temporal-order switching.
Only using CAM-Convs \cite{Facil:2019:CCC} works best as the test dataset contains images with different intrinsics, which CAM-Convs can handle.

\myparagraph{Empirical study for monocular scene flow estimation.} 
We train our full model with the proposed loss from \cref{eq:total_loss}.
Looking at the right side of \cref{table:augmentation_study} yields different conclusions for monocular scene flow estimation: \emph{using augmentation improves the scene flow accuracy in general, but using CAM-Convs \cite{Facil:2019:CCC} actually hurts the accuracy}.
We conjecture that the benefit of CAM-Convs -- introducing a test-time dependence on input camera intrinsics -- may be redundant for correspondence tasks (\ie optical flow, scene flow) and can hurt the accuracy.
We also observe that CAM-Convs lead to slight over-fitting on the training set, yielding marginally lower training loss (\eg, $<$ 1\%) but with higher error on the test set.
Therefore, we apply only geometric augmentation without CAM-Convs in the following.

\section{Experiments}
\label{sec:experiments}

\subsection{Implementation details}
\label{subsec:implementation_details}
\paragraph{Dataset.} 
For evaluation, we use the KITTI raw dataset \cite{Geiger:2013:VMR}, which provides stereo sequences covering \num{61} street scenes.
For the scene flow experiments, we use the \emph{KITTI Split} \cite{Godard:2017:UMD}: we first exclude \num{29} scenes contained in \emph{KITTI Scene Flow Training} \cite{Menze:2015:J3E,Menze:2018:OSF} and split the remaining $32$ scenes into \num{25801} sequences for training and \num{1684} for validation.
For evaluation and the ablation study, we use \emph{KITTI Scene Flow Training} as test set, since it provides ground-truth labels for disparity and scene flow for \num{200} images.

After training on \emph{KITTI Split} in a self-supervised manner, we optionally fine-tune our model using \emph{KITTI Scene Flow Training} \cite{Menze:2015:J3E,Menze:2018:OSF} to see how much accuracy gain can be obtained from annotated data.
We fine-tune our model in a semi-supervised setting by combining a supervised loss with our self-supervised loss (see below for details).

Additionally for evaluating monocular depth accuracy, we also use the \emph{Eigen Split} \cite{Eigen:2014:DMP} by excluding \num{28} scenes that the \num{697} test sequences cover, splitting into \num{20120} training sequences and \num{1338} validation sequences.

\myparagraph{Data augmentation.} 
We adopt photometric augmentations with random gamma, brightness, and color changes.
As discussed in \cref{sec:data_augmentation}, we use geometric augmentations consisting of horizontal flips \cite{Lai:2019:BSM,Lee:2019:LRF,Liu:2019:ULS,Wang:2019:UOS}, random scales, random cropping \cite{Chen:2019:SSL,Ranjan:2019:CCJ,Zou:2018:DFN}, and then resizing into $256 \times 832$ pixels as in previous work \cite{Lee:2019:LRF,Liu:2019:ULS,Luo:2019:EPC,Ranjan:2019:CCJ,Yang:2018:EPC}.

\myparagraph{Self-supervised training.} 
Our network is trained using Adam \cite{Kingma:2015:AAM} with hyper-parameters $\beta_1\!=\!0.9$ and $\beta_2\!=\!0.999$. 
Our initial learning rate is \num{2E-4}, and the mini-batch size is $4$.
We train our network for a total of \num{400}\si{\kilo} iterations.\footnote{Code is available at \href{https://github.com/visinf/self-mono-sf}{https://github.com/visinf/self-mono-sf}.}
In every iteration, the regularization weight $\lambda_\text{sf}$ in \cref{eq:total_loss} is dynamically determined to make the loss of the scene flow and disparity be equal in order to balance the optimization of the two joint tasks \cite{Hur:2019:IRR}.
Our specific learning rate schedule, as well as details on hyper-parameter choice and data augmentation are provided in the supplementary material.

Unlike previous approaches requiring stage-wise pre-training \cite{Lee:2019:LRF,Liu:2019:ULS,Wang:2019:UOS,Zou:2018:DFN} or iterative training \cite{Luo:2019:EPC,Ranjan:2019:CCJ,Yang:2018:EPC} of multiple CNNs due to the instability of joint training, our approach does not need any complex training strategies, but can just be trained from scratch all at once.
\emph{This highlights the practicality of our method.}

\myparagraph{Semi-supervised fine-tuning.}
We optionally fine-tune our trained model in a semi-supervised manner by mixing the two datasets, the KITTI raw dataset \cite{Geiger:2013:VMR} and \emph{KITTI Scene Flow Training} \cite{Menze:2015:J3E,Menze:2018:OSF}, at a ratio of $3:1$ in each batch of $4$.
The latter dataset provides sparse ground truth of the disparity map of the reference image, disparity information at the target image mapped into the reference image, as well as optical flow.
We apply our self-supervised loss to all samples and a supervised loss ($L_2$ for optical flow, $L_1$ for disparity) only for the sample from \emph{KITTI Scene Flow Training} after converting the scene flow into two disparity maps and optical flow.
Through semi-supervised fine-tuning, the proxy loss can guide pixels that the sparse ground truth cannot supervise. 
Moreover, the model can be prevented from heavy over-fitting on the only 200 annotated images by leveraging more data.
We train the network for \num{45}\si{\kilo} iterations with the learning rate starting at \num{4E-5} (see supplemental).

\myparagraph{Evaluation metric.} 
For evaluating the scene flow accuracy, we follow the evaluation metric of \emph{KITTI Scene Flow benchmark} \cite{Menze:2015:J3E,Menze:2018:OSF}.
It evaluates the accuracy of the disparity for the reference frame \emph{(D1-all)} and for the target image mapped into the reference frame \emph{(D2-all)}, as well as of the optical flow \emph{(F1-all)}.
Each pixel that exceeds a threshold of $3$ pixels or 5$\%$ \wrt~the ground-truth disparity or optical flow is regarded as an outlier; the metric reports the outlier ratio (in \%) among all pixels with available ground truth.
Furthermore, if a pixel satisfies all metrics (\ie, D$1$-all, D$2$-all, and F$1$-all), it is regarded as valid scene flow estimate from which the outlier rate for scene flow \emph{(SF1-all)} is calculated. 
For evaluating the depth accuracy, we follow the standard evaluation scheme introduced by Eigen \etal~\cite{Eigen:2014:DMP}.
We assume known test-time camera intrinsics.

\subsection{Ablation study}
\label{sec:ablation}
\begin{table}[t]
\centering
\footnotesize
\begin{tabular*}{\columnwidth}{@{\extracolsep{\fill}}cc@{\hskip 3em}S[table-format=2.2]S[table-format=2.2]S[table-format=2.2]S[table-format=2.2]@{}}
	\toprule
	Occ. & 3D points & {D$1$-all} & {D$2$-all} & {F$1$-all} & {SF$1$-all} \\
	\midrule
	\multicolumn{2}{l}{ \qquad \emph{(Basic)}}  & 33.31 & 51.33 & 24.74 & 64.05 \\
	\cmark &        & \bfseries 30.99 & 50.89 & 23.55 & 62.50 \\ 
	       & \cmark & 32.07 & 36.01 & 27.30 & 49.27 \\
	\cmark & \cmark & 31.25 & \bfseries 34.86 & \bfseries 23.49 & \bfseries 47.05 \\
	\bottomrule	  
\end{tabular*}
\caption{\textbf{Ablation study on the loss function}: based on the \emph{Basic} 2D loss consisting of photometric and smoothness loss, the 3D point reconstruction loss (\emph{3D points}) improves scene flow accuracy, especially when discarding occluded pixels in the loss (\emph{Occ.}).}
\label{table:ablation_loss}
\vspace{-1.0em}	
\end{table}

To confirm the benefit of our various contributions, we conduct ablation studies based on our full model using the \emph{KITTI split} with data augmentation applied.

\myparagraph{Proxy loss for self-supervised learning.}
Our proxy loss consists of three main components: \emph{(i)} \emph{Basic}: a basic combination of 2D photometric and smoothness losses, \emph{(ii)} \emph{3D points}: the 3D point reconstruction loss for scene flow, and \emph{(iii)} \emph{Occ.}: whether applying the photometric and point reconstruction loss only for visible pixels or not.
\cref{table:ablation_loss} shows the contribution of each loss toward the accuracy.

The \emph{3D points} loss significantly contributes to more accurate scene flow by yielding more accurate disparity on the target image (D$2$-all).
This highlights the importance of penalizing the actual 3D Euclidean distance between two corresponding 3D points (\cf \cref{fig:loss_sf_pt}), which typical loss functions in 2D space (\ie \emph{Basic} loss) as in previous work \cite{Luo:2019:EPC,Yang:2018:EPC} cannot.

Taking occlusion into account consistently improves the scene flow accuracy further.
The main objective of our proxy loss is to reconstruct the reference image as closely as possible, which can lead to hallucinating potentially incorrect estimates of disparity and scene flow in the occluded areas.
Thus, discarding occluded pixels in the loss is critical to achieving accurate predictions.

\myparagraph{Single decoder \vs separate decoders.}
To verify the key motivation of \emph{decomposing optical flow cost volumes into depth and scene flow using a single decoder}, we compare against a model with separate decoders for each task, which follows the conventional design of other multi-task methods \cite{Chen:2019:SSL,Liu:2019:ULS,Luo:2019:EPC,Ranjan:2019:CCJ,Yang:2018:EPC,Zou:2018:DFN}.
We also prepare two baselines that estimate either monocular depth or optical flow only, to assess the capacity our modified PWC-Net for each task.

\begin{table}[t]
\centering
\footnotesize
\begin{tabularx}{\columnwidth}{@{}XS[table-format=3.2]S[table-format=2.2]S[table-format=2.2]S[table-format=3.2]@{}}
	\toprule
	Model & {D$1$-all} & {D$2$-all} & {F$1$-all} & {SF$1$-all} \\ 
	\midrule 
	Monocular depth only       & \bfseries 27.59 &  {--}   &  {--}   &  {--}   \\
	Optical flow only          &  {--}   &  {--}   & 24.27 &  {--}   \\
	Scene flow w/ separate decoders & 100 & 97.22 & 27.63  & 100 \\ 
	Scene flow w/ a single decoder & 31.25 & \bfseries 34.86 & \bfseries 23.49 & \bfseries 47.05 \\ 
	\bottomrule	
\end{tabularx}
\caption{\textbf{Single decoder \vs separate decoders}: using a single decoder yields stable training and comparable accuracy on both tasks to models that target each individual task separately.}
\label{table:ablation_decoder}
\vspace{-1.0em}
\end{table}

\cref{table:ablation_decoder} demonstrates our ablation study on the network design.
First, our model with a single decoder achieves comparable or even higher accuracy on the depth and optical flow tasks, compared to using the same network only for each individual task.
We thus conclude that solving monocular scene flow using a single joint network can substitute the two individual tasks given the same amount of training resources and network capacity.

When separating the decoders, we find that the network cannot be trained stably, yielding trivial solutions for disparity.
This is akin to issues observed by previous multi-task approaches, which require pre-training or iterative training for multiple CNNs \cite{Lee:2019:LRF,Liu:2019:ULS,Luo:2019:EPC,Ranjan:2019:CCJ,Wang:2019:UOS,Yang:2018:EPC,Zou:2018:DFN}.
In contrast, having a single decoder resolves the imbalance and stability problem by virtue of joint estimation.
We include a more comprehensive analysis in the supplemental, gradually splitting the decoder to closely analyze its behavior.

\begin{table}[t]
\centering
\footnotesize
\begin{tabularx}{\columnwidth}{@{}XS[table-format=2.2]@{\hskip 0.7em}S[table-format=2.2]@{\hskip 0.7em}S[table-format=2.2]@{\hskip 0.6em}S[table-format=2.2]@{\hskip 0.9em}S[table-format=2.3,table-space-text-post = \si{\s}]@{}}
	\toprule
	Method    &   {D$1$-all}   &  {D$2$-all}  & {F$1$-all}   & {SF$1$-all} & {Runtime} \\  
	\midrule 
	DF-Net \cite{Zou:2018:DFN} & 46.50 & 61.54 & 27.47 & 73.30 & {--} \\ 
	GeoNet \cite{Yin:2018:GNU} & 49.54 & 58.17 & 37.83 & 71.32 & 0.06\si{\s} \\	 
	EPC \cite{Yang:2018:EPC}   & 26.81 & 60.97 & 25.74 & {(\gt 60.97)} & 0.05\si{\s}\\
	EPC++ \cite{Luo:2019:EPC}  & \bfseries 23.84 & 60.32 & \bfseries 19.64 & {(\gt 60.32)} & 0.05\si{\s} \\
	\textbf{Self-Mono-SF (Ours)}   & 31.25 & \bfseries 34.86 & 23.49 & \bfseries 47.05 & 0.09\si{\s}\\
	\midrule 
	Mono-SF \cite{Brickwedde:2019:MSF} & 16.72 & 18.97 & 11.85 & 21.60 & 41\si{\s}\\ 
	\textbf{Self-Mono-SF-ft (Ours)} & (2.89) & (3.91) & (6.19) & (7.53) & 0.09\si{\s}\\
	\bottomrule
\end{tabularx}
\caption{\textbf{Monocular scene flow evaluation on \emph{KITTI Scene Flow Training}}: our self-supervised learning approach significantly outperforms all multi-task CNN methods \emph{(upper rows)} on the scene flow metric, \emph{SF1-all}. Lower rows provide the accuracy of a semi-supervised method \cite{Brickwedde:2019:MSF} and our fine-tuned model.}
\label{table:eval_monosf_kitti_train}
\end{table}

\begin{table}[t]
\centering
\footnotesize
\begin{tabularx}{\columnwidth}{@{}XS[table-format=2.2]@{\hskip 0.6em}S[table-format=2.2]@{\hskip 0.6em}S[table-format=2.2]@{\hskip 0.6em}S[table-format=2.2]@{\hskip 0.9em}S[table-format=2.3,table-space-text-post = \si{\s}]@{}}
	\toprule
	Method & {D$1$-all} & {D$2$-all} & {F$1$-all} & {SF$1$-all} & {Runtime} \\  
	\midrule 
	DRISF \cite{Ma:2019:DRI} & 2.55 & \bfseries 4.04 & \bfseries 4.73 & \bfseries 6.31 & 0.75\si{\s}\\
	SENSE \cite{Jiang:2019:SAS} & \bfseries 2.22 & 5.89 & 7.64 & 9.55  & 0.32\si{\s}\\
	PWOC-3D \cite{Saxena:2019:PDO} & 5.13 & 8.46 & 12.96 & 15.69 & 0.13\si{\s}\\
	UnOS \cite{Wang:2019:UOS} & 6.67 & 12.05 & 18.00 & 22.32 & 0.08\si{\s}\\
	\midrule 
	Mono-SF \cite{Brickwedde:2019:MSF} & \bfseries 16.32 & \bfseries 19.59 & \bfseries 12.77 & \bfseries 23.08 & 41\si{\s} \\
	\textbf{Self-Mono-SF (Ours)}   & 34.02 & 36.34 & 23.54 & 49.54 & 0.09\si{\s}\\
	\textbf{Self-Mono-SF-ft (Ours)} & 22.16 & 25.24 & 15.91 & 33.88 & 0.09\si{\s}\\
	\bottomrule	
\end{tabularx}
\caption{\textbf{Scene flow evaluation on \emph{KITTI Scene Flow Test}}: we compare our method with stereo \emph{(top)} and monocular \emph{(bottom)} scene flow methods. Despite the difficult setting, our fine-tuned model demonstrates encouraging results in real-time.}
\label{table:eval_sf_kitti_test}
\vspace{-1.0em}
\end{table}

\begin{table*}[t]
\centering
\begin{minipage}[b]{0.58\textwidth}
\centering
\scriptsize
\setlength\tabcolsep{4.15pt}
\begin{tabularx}{\linewidth}{@{}c@{\hskip 0.5em}XS[table-format=1.3,round-precision=3]@{\hskip 0.6em}S[table-format=1.3,round-precision=3]@{\hskip 0.6em}S[table-format=1.3,round-precision=3]@{\hskip 0.6em}S[table-format=1.3,round-precision=3]S[table-format=1.3,round-precision=3]@{\hskip 0.6em}S[table-format=1.3,round-precision=3]@{\hskip 0.6em}S[table-format=1.3,round-precision=3]@{}}
	\toprule
	 & & \multicolumn{4}{c}{(lower is better)} & \multicolumn{3}{c}{(higher is better)}  \\ \cmidrule(lr){3-6} \cmidrule(lr){7-9}
	{Split} & {Method} & {Abs Rel} & {Sq Rel} & {RMSE} & {RMSE log} & {\tiny$\delta\!\lt\!1.25$} & {\tiny$\delta\!\lt\!1.25^2$} & {\tiny$\delta\!\lt\!1.25^3$}\\ \midrule 
	\multirow{4}{*}{\rotatebox[origin=c]{90}{KITTI}}
	& DF-Net \cite{Zou:2018:DFN} & 0.150 & 1.124 &  5.507 & 0.223 & 0.806 & 0.933 & 0.973 \\
	& EPC$^{\S}$ \cite{Yang:2018:EPC} & 0.109 & 1.004 & 6.232 & 0.203 & 0.853 & 0.937 & 0.975 \\	
	& Liu~\etal~$^{\S}$ \cite{Liu:2019:ULS} & 0.108 & 1.020 & 5.528 & 0.195 & 0.863 & 0.948 & 0.980 \\ 
	& \textbf{Self-Mono-SF (Ours)}$^{\S}$ & \bfseries 0.106 & \bfseries 0.888 & \bfseries 4.853 & \bfseries 0.175 & \bfseries 0.879 & \bfseries 0.965 & \bfseries 0.987 \\ \midrule
	\multirow{5}{*}{\rotatebox[origin=c]{90}{Eigen}}
	& GeoNet \cite{Yin:2018:GNU} & 0.155 & 1.296 & 5.857 & 0.233 & 0.793 & 0.931 & 0.973 \\
	& CC \cite{Ranjan:2019:CCJ} & 0.140 & 1.070 & 5.326 & 0.217 & 0.826 & 0.941 & 0.975 \\
	& GLNet(-ref.) \cite{Chen:2019:SSL} & 0.135 & 1.070 & 5.230 & 0.210 & 0.841 & 0.948 & \bfseries 0.980 \\	
	& EPC$^{\S}$ \cite{Yang:2018:EPC} & 0.127 & 1.239 & 6.247 & 0.214 & 0.847 & 0.926 & 0.969 \\ 	
	& EPC++$^{\S}$ \cite{Luo:2019:EPC} & 0.127 & \bfseries 0.936 & 5.008 & 0.209 & 0.841 & 0.946 & 0.979 \\
	& \textbf{Self-Mono-SF (Ours)}$^{\S}$ & \bfseries 0.125 & 0.978 & \bfseries 4.877 & \bfseries 0.208 & \bfseries 0.851 & \bfseries 0.950 & 0.978 \\
	\bottomrule
\end{tabularx}
\caption{\textbf{Monocular depth comparison}: our method demonstrates superior accuracy on the \emph{KITTI split} and competitive accuracy on the \emph{Eigen split} compared to all published multi-task methods. $^{\S}$method using stereo sequences for training.}
\label{table:eval_monodepth}
\end{minipage} \quad 
\begin{minipage}[b]{0.39\textwidth}
\centering
\scriptsize
\setlength\tabcolsep{4.15pt}
\begin{tabularx}{\linewidth}{@{}c|@{\hskip 0.4em}XS[table-format=2.2]S[table-format=2.2]S[table-format=2.2]@{}}
	\toprule
	\multicolumn{2}{l}{} & \multicolumn{2}{c}{Train} & {Test} \\ \cmidrule(lr){3-4} \cmidrule(l){5-5} 
	\multicolumn{2}{@{}l}{Method} & {EPE} & {F1-all} & {F1-all} \\ \midrule 
	\multirow{3}{*}{\rotatebox[origin=c]{90}{Stereo}}
	& Lai~\etal~\cite{Lai:2019:BSM} & 7.13 & 27.13 & {--} \\
	& Lee~\etal~\cite{Lee:2019:LRF} & 8.74 & 20.88 & {--} \\
	& UnOS~\cite{Wang:2019:UOS} & \bfseries 5.58 & {--} & \bfseries 18.00 \\  \midrule 
	\multirow{7}{*}{\rotatebox[origin=c]{90}{Monocular}}
	& GeoNet \cite{Yin:2018:GNU} & 10.81 & {--} & {--} \\ 
	& DF-Net \cite{Zou:2018:DFN}& 8.98 & 26.01 & 25.70 \\ 
	& GLNet \cite{Chen:2019:SSL}& 8.35 & {--} & {--} \\ 
	& EPC$^{\S}$ \cite{Yang:2018:EPC}  & {--} & 25.74 & {--} \\ 
	& EPC++$^{\S}$ \cite{Luo:2019:EPC} & \bfseries 5.43 & \bfseries 19.64 & \bfseries 20.52 \\ 
	& Liu~\etal~$^{\S}$ \cite{Liu:2019:ULS}& 5.74 & {--} & {--}  \\ 
	& \textbf{Self-Mono-SF (Ours)}$^{\S}$ & 7.51 & 23.49 & 23.54 \\ \bottomrule	
\end{tabularx}
\caption{\textbf{Optical flow estimation on the \emph{KITTI split}}: our method demonstrates comparable accuracy to both monocular and stereo-based multi-task methods.}
\label{table:eval_flow}
\end{minipage}
\vspace{-0.5em}
\end{table*}

{
\begin{figure*}[!ht]
\centering
\scriptsize
\setlength\tabcolsep{0.1pt}
\renewcommand{\arraystretch}{0.2}
\begin{tabular}{>{\centering\arraybackslash}m{.140\textwidth} @{\hskip 0.2em} >{\centering\arraybackslash}m{.278\textwidth} @{\hskip 0.2em} >{\centering\arraybackslash}m{.278\textwidth} @{\hskip 0.2em} >{\centering\arraybackslash}m{.30\textwidth}}
	\includegraphics[width=\linewidth]{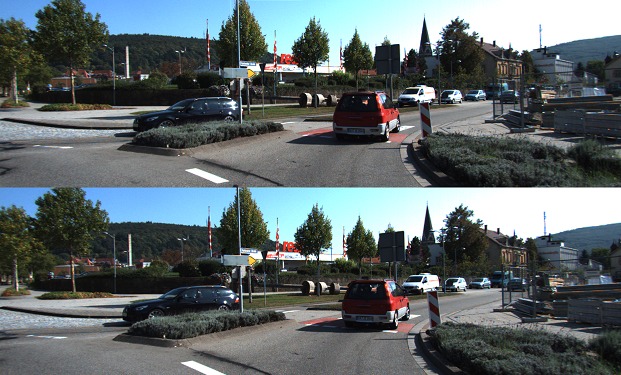} & 
	\includegraphics[width=\linewidth]{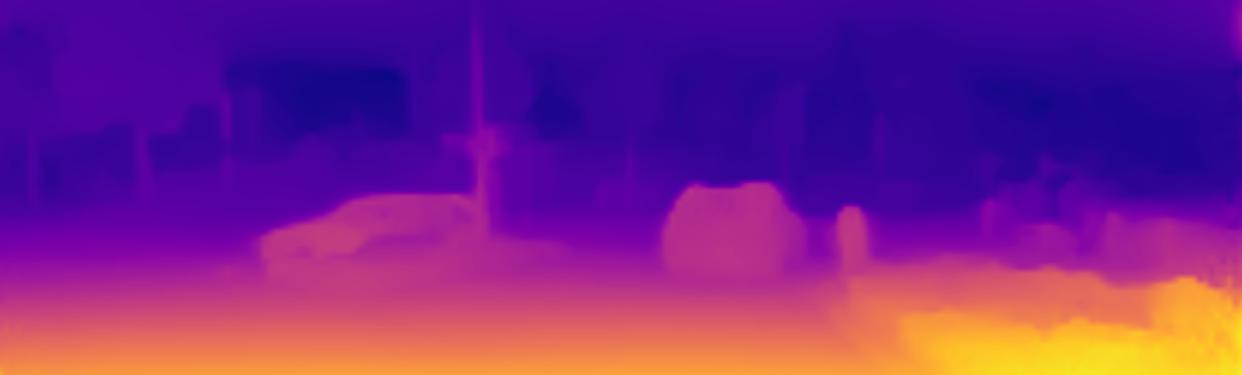} & 
	\includegraphics[width=\linewidth]{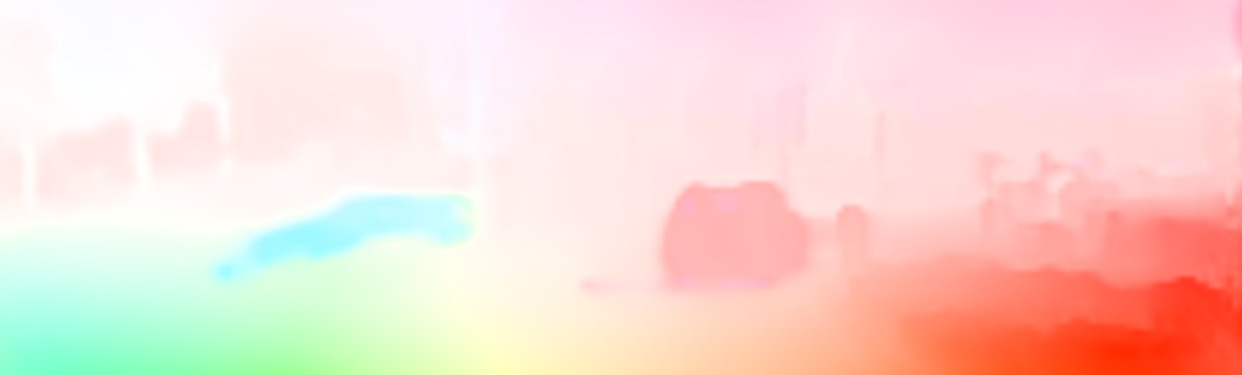} & 
	\includegraphics[width=\linewidth]{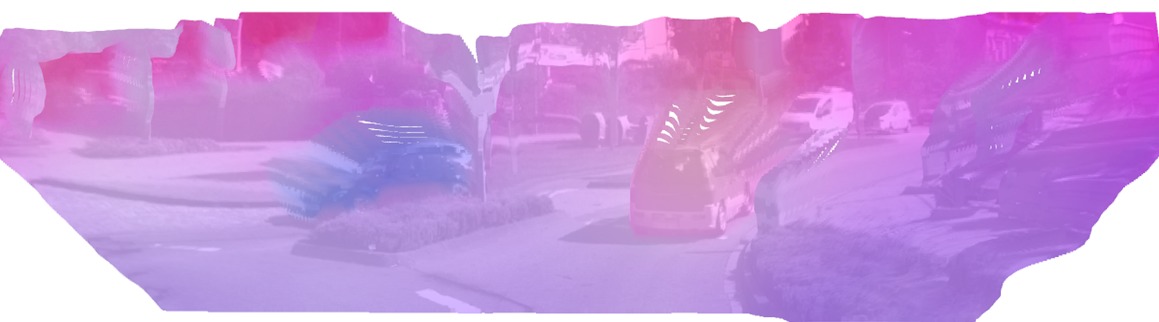} \\

	\includegraphics[width=\linewidth]{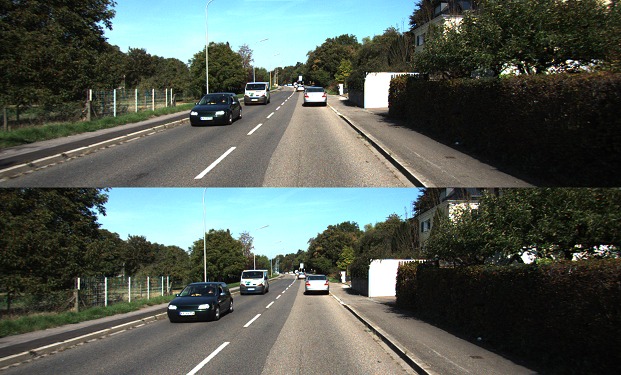} & 
	\includegraphics[width=\linewidth]{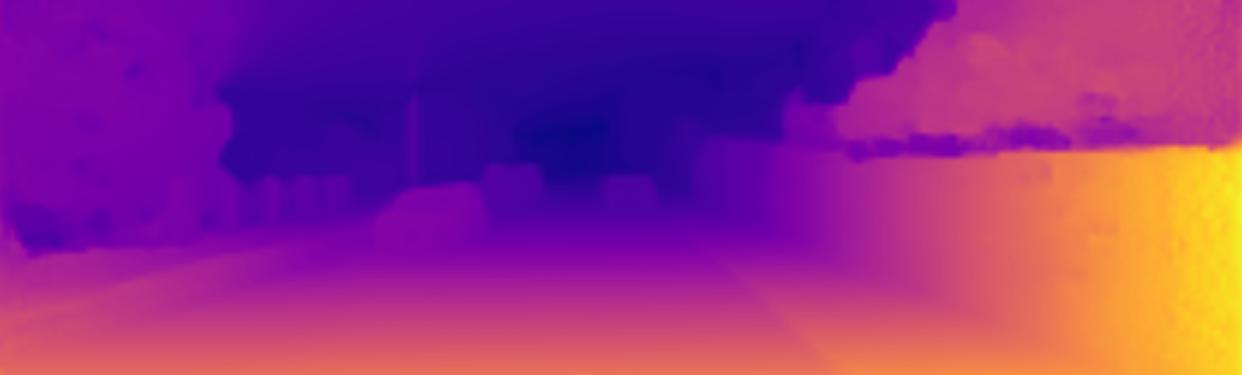} & 
	\includegraphics[width=\linewidth]{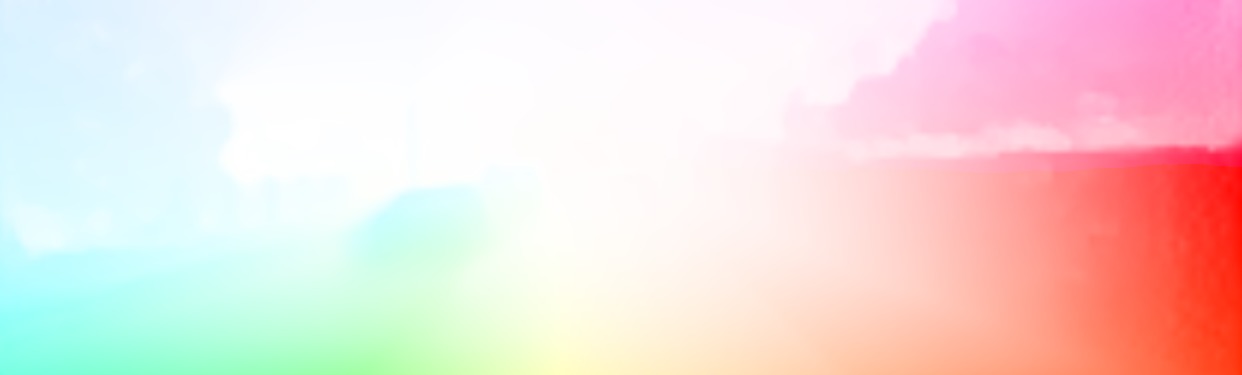} & 
	\includegraphics[width=\linewidth]{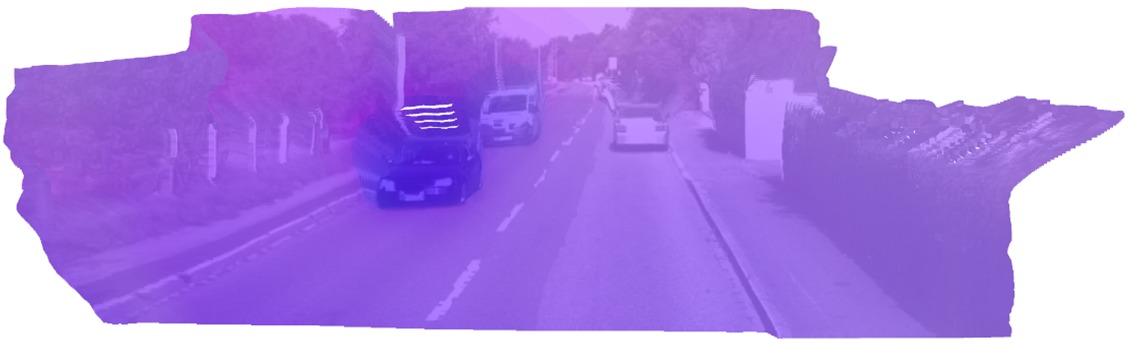} \\

	\includegraphics[width=\linewidth]{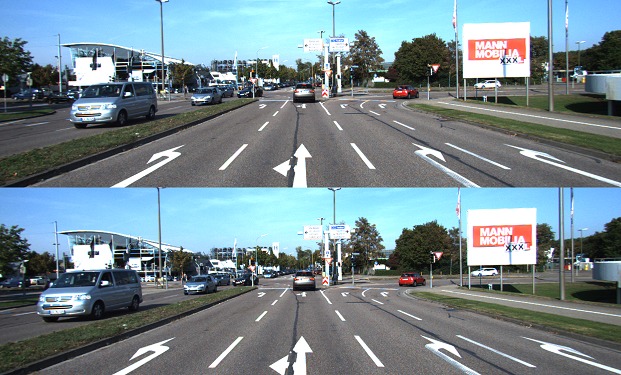} & 
	\includegraphics[width=\linewidth]{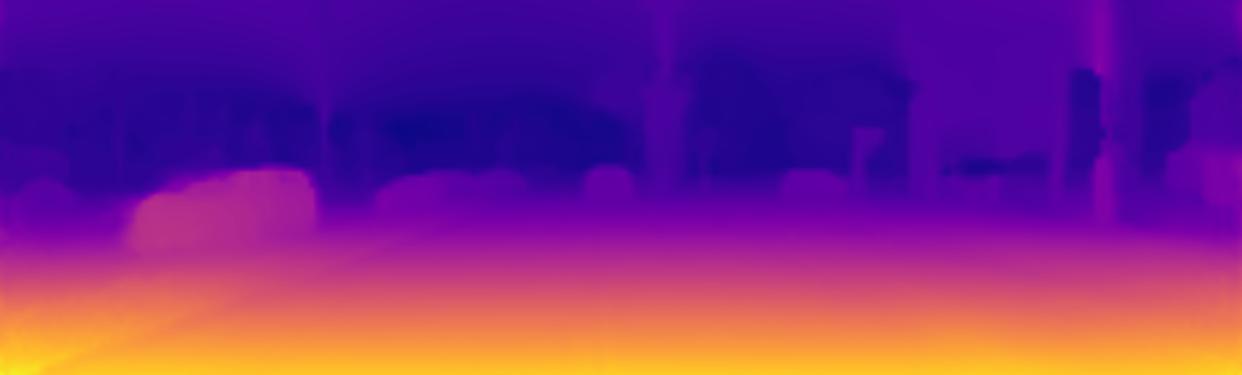} & 
	\includegraphics[width=\linewidth]{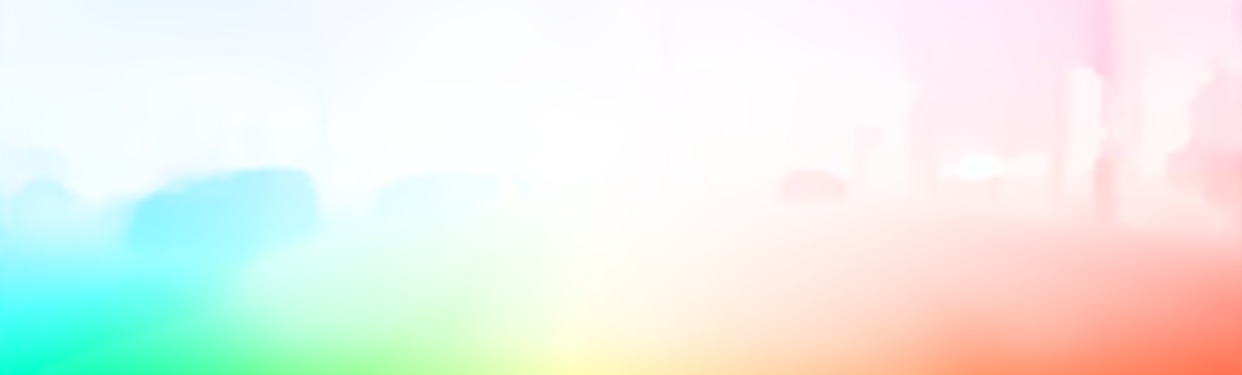} & 
	\includegraphics[width=\linewidth]{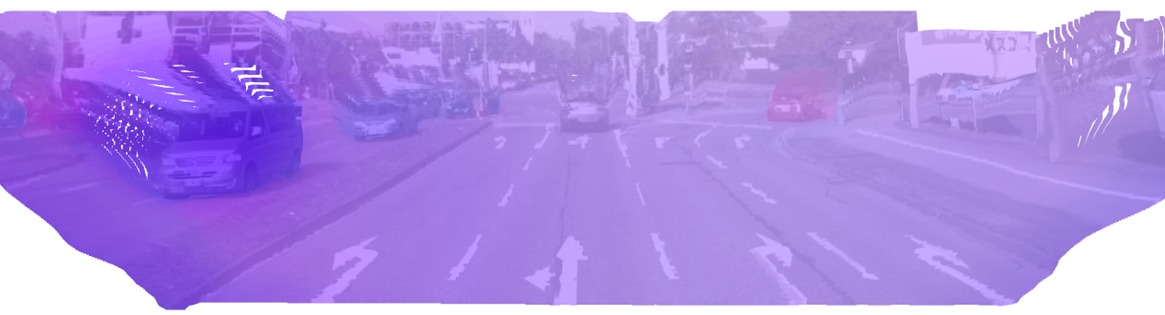} \\ [0.5em] \\
	(a) Input images & (b) Monocular depth & (c) Optical flow & (d) 3D visualization of scene flow \\
\end{tabular}
\caption{\textbf{Qualitative results of our monocular scene flow results (Self-Mono-SF-ft) on KITTI 2015 Scene Flow Test}: each scene shows \emph{(a)} two input images, \emph{(b)} monocular depth, \emph{(c)} optical flow, and \emph{(d)} a 3D visualization of estimated depth, overlayed with the reference image, and colored with the $(x,z)$-coordinates of the 3D scene flow using the standard optical flow color coding.}
\label{fig:qualitative_img}
\vspace{-1.0em}
\end{figure*}
}

\subsection{Monocular scene flow}

\cref{table:eval_monosf_kitti_train} demonstrates the comparison to existing monocular scene flow methods on \emph{KITTI Scene Flow Training}.
We compare against state-of-the-art multi-task CNN methods \cite{Luo:2019:EPC,Yang:2018:EPC,Yin:2018:GNU,Zou:2018:DFN} on the scene flow evaluation metric.
Our model significantly outperforms these methods by a large margin, confirming our method as the most accurate monocular scene flow method using CNNs to date.
For example, our method yields more than $40.1\%$ accuracy gain for estimating the disparity on the target image (D$2$-all).
Though the two methods, EPC \cite{Yang:2018:EPC} and EPC++ \cite{Luo:2019:EPC}, do not provide scene flow accuracy numbers (SF$1$-all), we can conclude that our method clearly outperforms all four methods in SF$1$-all, since SF$1$-all is lower-bounded by D$2$-all.

Our self-supervised learning approach \emph{(Self-Mono-SF)} is outperformed only by Mono-SF \cite{Brickwedde:2019:MSF}, which is a semi-supervised method using pseudo labels, semantic instance knowledge, and an additional dataset (Cityscapes \cite{Cordts:2016:CDS}).
However, our method runs more than two orders of magnitude faster.
We also provide the accuracy of our fine-tuned model \emph{(Self-Mono-SF-ft)} on the training set for reference.

\cref{table:eval_sf_kitti_test} shows the comparison with stereo and monocular scene flow methods on the KITTI Scene Flow 2015 benchmark.
\cref{fig:qualitative_img} provides a visualization.
Our semi-supervised fine-tuning further improves the accuracy, going toward that of Mono-SF \cite{Brickwedde:2019:MSF}, but with a more than $400\times$ faster run-time.
For further accuracy improvements, \eg rigidity refinement \cite{Jiang:2019:SAS,Liu:2019:ULS}, exploiting an external dataset \cite{Cordts:2016:CDS} for pre-training, or pseudo ground truth \cite{Brickwedde:2019:MSF} can be applied on top of our self-supervised learning and semi-supervised fine-tuning pipeline without affecting run-time.

\subsection{Monocular depth and optical flow}

Finally, we provide a comparison to unsupervised multi-task CNN approaches \cite{Chen:2019:SSL,Liu:2019:ULS,Luo:2019:EPC,Ranjan:2019:CCJ,Yang:2018:EPC,Yin:2018:GNU,Zou:2018:DFN} regarding the accuracy of depth and optical flow.
We do not report methods that use extra datasets (\eg, the Cityscapes dataset \cite{Cordts:2016:CDS}) for pre-training or online fine-tuning \cite{Chen:2019:SSL}, which is known to give an accuracy boost.

For monocular depth estimation in \cref{table:eval_monodepth}, our monocular scene flow method outperforms all published multi-task methods on the \emph{KITTI Split} \cite{Godard:2017:UMD} and demonstrates competitive accuracy on the \emph{Eigen split} \cite{Eigen:2014:DMP}.
Note that some of the methods \cite{Ranjan:2019:CCJ,Yin:2018:GNU,Zou:2018:DFN} use \emph{ground truth} to correctly scale their predictions at test time, which gives them an unfair advantage, but are still outperformed by ours.

For optical flow estimation in \cref{table:eval_monodepth}, our method demonstrates comparable accuracy to existing state-of-the-art monocular \cite{Chen:2019:SSL,Yang:2018:EPC,Yin:2018:GNU,Zou:2018:DFN} and stereo methods \cite{Lai:2019:BSM,Lee:2019:LRF}, in part outperforming them.

One reason why our flow accuracy may not surpass all previous methods is that we use a 3D scene flow regularizer and not a 2D optical flow regularizer.
This is consistent with our goal of estimating 3D scene flow, but it is known that using a regularizer in the target space is critical for achieving best accuracy \cite{Vogel:2015:3SF}.
While our choice of 3D regularizer is not ideal for optical flow estimation, its benefits manifest in 3D.
For example, while we do not outperform EPC++ \cite{Luo:2019:EPC} in terms of 2D flow accuracy, we clearly surpass it in terms of scene flow accuracy (see \cref{table:eval_monosf_kitti_train}).
Consequently, our approach is not only the first CNN approach to monocular scene flow estimation that directly predicts the 3D scene flow, but also outperforms existing multi-task CNNs.

\section{Conclusion}
\label{sec:conclusion}
We proposed a CNN-based monocular scene flow estimation approach based on PWC-Net that predicts 3D scene flow directly.
A crucial feature is our single joint decoder for depth and scene flow, which allows to overcome the limitations of existing multi-task approaches such as complex training schedules or lacking occlusion handling.
We take a self-supervised approach, where our 3D loss function and occlusion reasoning significantly improve the accuracy.
Moreover, we show that a suitable augmentation scheme is critical for competitive accuracy.
Our model achieves state-of-the-art scene flow accuracy among un-/self-supervised monocular methods, and our semi-supervised fine-tuned model approaches the accuracy of the best monocular scene flow method to date, while being orders of magnitude faster.
With competitive accuracy and real-time performance, our method provides a solid foundation for CNN-based monocular scene flow estimation as well as follow-up work.

{\small
\bibliographystyle{ieee_fullname}

}
\flushcolsend


\title{Self-Supervised Monocular Scene Flow Estimation \\ {\large -- Supplementary Material --}}
\author{Junhwa Hur \qquad\qquad Stefan Roth \\ Department of Computer Science, TU Darmstadt}
\maketitle
\appendix


In this supplementary material, we provide further details on the learning rate schedules, data augmentation, and the hyper-parameter settings.
Afterwards, we provide a more comprehensive study of the decoder design, qualitative examples for the loss ablation study, and a qualitative comparison with the state-of-the-art Mono-SF approach \cite{Brickwedde:2019:MSF}.

\section{Learning Rate Schedule}
\label{sec:lr_schedule}

\begin{figure}[t]
\centering
\subcaptionbox{Learning rate schedule for self-supervised learning.\label{fig:lr_unsup}}{\includegraphics[width=0.97\linewidth]{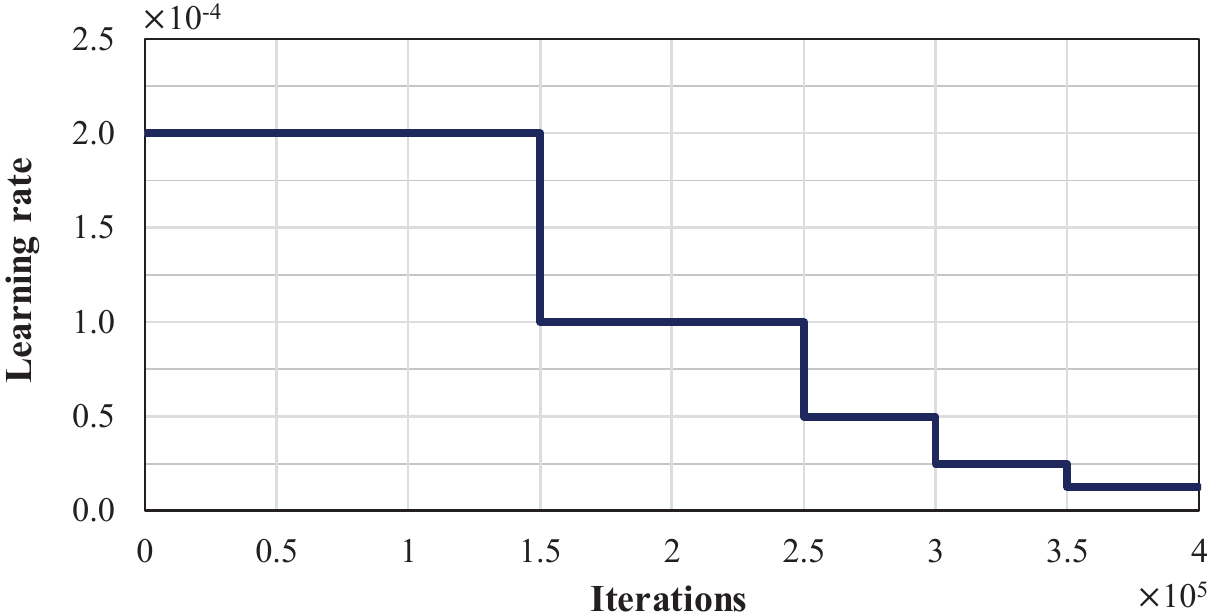}} \\[2mm]
\subcaptionbox{Learning rate schedule for fine-tuning.\label{fig:lr_ft}}{\includegraphics[width=0.97\linewidth]{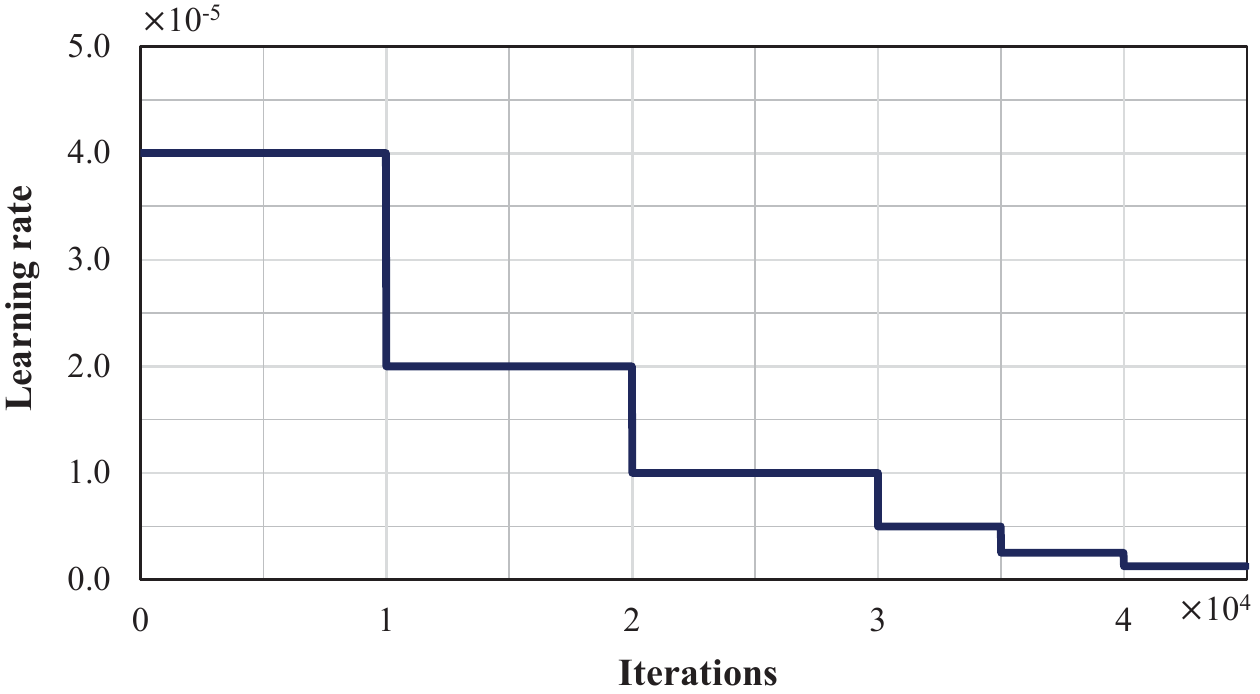}}
\caption{\textbf{Learning rate schedules} for \emph{(a)} self-supervised learning and \emph{(b)} semi-supervised fine-tuning.}
\label{fig:lr_schedules}
\end{figure}

\cref{fig:lr_schedules} illustrates the learning rate schedules for both self-supervised learning and semi-supervised fine-tuning. 
When first training our model in a self-supervised manner for \num{400}k iterations, the initial learning rate starts from \num{2E-4} and is halved at \num{150}k, \num{250}k, \num{300}k, and \num{350}k iteration steps.
When fine-tuning in a semi-supervised manner afterwards, the training schedule consists of \num{45}k iterations; the initial learning rate starts from \num{4E-5} and is halved at \num{10}k, \num{20}k, \num{30}k, \num{35}k, and \num{40}k iteration steps.

\section{Details on Data Augmentation}
\label{sec:detail_data_augmentation}

As discussed in the main paper, we perform photometric and geometric augmentations at training time. 
Here we provide more details on our augmentation setup for both self-supervised training and semi-supervised fine-tuning.

\myparagraph{Augmentations for self-supervised training.}
We apply photometric augmentations with $50\%$ probability.
Specifically, we adopt random gamma adjustments, uniformly sampled from $[0.8, 1.2]$, brightness changes with a multiplication factor that is uniformly sampled in $[0.5, 2.0]$, and random color changes with a multiplication factor that is uniformly sampled in $[0.8, 1.2]$ for each color channel.

For geometric augmentations, we first randomly crop the input images with a random scale factor uniformly sampled in $[93\%, 100\%]$ and apply random translations uniformly sampled from $[-3.5\%, 3.5\%]$ \wrt the input image size.
Then we resize the cropped image to $256 \times 832$ pixels as in previous work \cite{Lee:2019:LRF,Liu:2019:ULS,Luo:2019:EPC,Ranjan:2019:CCJ,Yang:2018:EPC}.
We also apply a horizontal flip \cite{Lai:2019:BSM,Lee:2019:LRF,Liu:2019:ULS,Wang:2019:UOS} with $50\%$ probability.
Because the geometric augmentations have an effect on the camera intrinsics, we adjust the intrinsic camera matrix accordingly by calculating the corresponding camera center and focal length of each augmented image.
At testing time, we only resize the input image to $256 \times 832$ pixels without photometric augmentation.

\myparagraph{Augmentations for semi-supervised fine-tuning.}
Likewise, we also apply the same photometric augmentations with $50\%$ probability.
For geometric augmentations, we only apply random cropping without scaling and then resize to $256 \times 832$ pixels.
Not performing scaling is to avoid changes to the ground truth, which may happen if zooming and interpolating the sparse ground truth.
The crop size $s \cdot h_{0} \times s \cdot w_{0}$ is determined by the cropping factor $s$ that is uniformly sampled in $[94\%, 100\%]$, where $h_{0}$ and $w_{0}$ is height and width of the original input resolution.
At testing time, the same augmentation scheme as during self-supervised training applies: resizing the input images to $256 \times 832$ pixels without photometric augmentation.
However, we note that better augmentation protocols can likely be discovered with further investigation \cite{Aviram:2020:SFD}.

\section{Hyper-Parameter Settings}
\label{hyperparam_setting}

Our self-supervised proxy loss in \cref{eq:total_loss} of the main paper has a total of $6$ hyper-parameters, which could make it difficult to achieve satisfactory results without careful tuning.
In this section, we thus discuss how we choose the hyper-parameters and provide an analysis on how sensitive the scene flow accuracy is depending on the hyper-parameter choices.

First, as discussed in the main paper, the balancing weight $\lambda_\text{sf}$ between the two joint tasks in \cref{eq:total_loss} is dynamically determined to make the loss of the scene flow and disparity be equal in every iteration \cite{Hur:2019:IRR}.
For the disparity loss, we simply adopt the same hyper-parameters (\ie, $\lambda_\text{d\_sm}$, $\alpha$, and $\beta$ in \cref{eq:disp_loss,eq:photometric_loss,eq:disp_smooth_loss}, respectively) as in previous work \cite{Godard:2017:UMD}, which leaves only two hyper-parameters, $\lambda_\text{sf\_sm}$ and $\lambda_\text{sf\_pt}$, to tune in the scene flow loss, \cref{eq:sf_loss}.
We perform grid search on the two parameters.

\begin{table}[t]
\centering
\footnotesize
\setlength\tabcolsep{4.5pt}
\begin{tabular*}{\columnwidth}{@{}cc@{\hskip 2em}S[table-format=1.3,round-precision=3]S[table-format=1.3,round-precision=3]S[table-format=2.2]S[table-format=2.2]S[table-format=2.2]S[table-format=2.2]@{}}
    \toprule
     & & {Depth} & {Flow} & \multicolumn{4}{c}{Scene Flow} \\
    \cmidrule(r){3-3} \cmidrule(lr){4-4} \cmidrule(l){5-8}
    {$\lambda_\text{sf\_sm}$} & {$\lambda_\text{sf\_pt}$} & {Abs Rel} & {EPE} & {D$1$-all} & {D$2$-all} & {F$1$-all} & {SF$1$-all} \\
    \midrule 
    \multirow{4}{*}{\tablenum[table-format=3.1]{1}} & \tablenum[table-format=1.3,round-precision=3]{0.005} & \bfseries 0.104 & 7.118 & \bfseries 30.50 & 51.48 & \bfseries 22.32 & 62.97 \\
        & \tablenum[table-format=1.3,round-precision=2]{0.05} & 0.107 & \bfseries 7.057 & 32.56 & 49.45 & 22.33 & 61.27 \\
        & \tablenum[table-format=1.3,round-precision=1]{0.1} & 0.109 & 7.319 & 33.65 & \bfseries 35.57 & 22.58 & \bfseries 47.46 \\
        & \tablenum[table-format=1.3,round-precision=1]{0.5} & 0.117 & 8.259 & 33.91 & 36.24 & 25.18 & 48.72 \\ \midrule
                                    
    \multirow{4}{*}{\tablenum[table-format=3.1]{10}} & \tablenum[table-format=1.3,round-precision=3]{0.005} & \bfseries 0.105 & \bfseries 6.934 & \bfseries 31.18 & 52.29 & \bfseries 22.15 & 63.47 \\
        & \tablenum[table-format=1.3,round-precision=1]{0.2} & 0.108 & 7.421 & 31.37 & \bfseries 34.39 & 22.73 & \bfseries 46.08 \\
        & \tablenum[table-format=1.3,round-precision=1]{0.3} & 0.110 & 7.379 & 31.91 & 34.42 & 23.79 & 47.10 \\
        & \tablenum[table-format=1.3,round-precision=1]{0.4} & 0.113 & 7.773 & 32.79 & 35.53 & 23.98 & 47.63 \\ \midrule
                                    
    \multirow{4}{*}{\tablenum[table-format=3.1]{200}} & \tablenum[table-format=1.3,round-precision=3]{0.005} & \bfseries 0.103 & \bfseries 6.883 & \bfseries 30.48 & 50.05 & \bfseries 22.65 & 61.47 \\
        & \tablenum[table-format=1.3,round-precision=1]{0.1} & 0.108 & 7.525 & 31.49 & 46.50 & 23.38 & 59.17 \\
        & \tablenum[table-format=1.3,round-precision=1]{0.2} & 0.107 & 7.197 & 31.40 & \bfseries 34.75 & 23.02 & \bfseries 46.95 \\
        & \tablenum[table-format=1.3,round-precision=1]{0.4} & 0.114 & 7.435 & 33.35 & 35.56 & 24.30 & 48.25 \\ \midrule[\heavyrulewidth]
     
    \tablenum[table-format=3.1,round-precision=1]{0.1} & {\multirow{4}{*}{\tablenum[table-format=1.3,round-precision=3]{0.005}}} & 0.106 & 6.839 & 31.47 & 52.20 & 22.39 & 63.70 \\
    \tablenum[table-format=3.1]{1}   &     & \bfseries 0.104 & 7.118 & \bfseries 30.50 & 51.48 & 22.32 & 62.97 \\
    \tablenum[table-format=3.1]{10}  &     & 0.105 & 6.934 & 31.18 & 52.29 & \bfseries 22.15 & 63.47 \\
    \tablenum[table-format=3.1]{100} &     & 0.105 & \bfseries 6.723 & 31.15 & \bfseries 51.05 & 22.18 & \bfseries 62.55 \\ \midrule
                                
    \tablenum[table-format=3.1]{1}   & {\multirow{4}{*}{\tablenum[table-format=1.3,round-precision=1]{0.2}}} & 0.109 & \bfseries 7.118 & 31.81 & 34.95 & 23.01 & 46.82 \\
    \tablenum[table-format=3.1]{10}  &      & 0.108 & 7.421 & 31.37 & \bfseries 34.39 & \bfseries 22.73 & \bfseries 46.08 \\
    \tablenum[table-format=3.1]{100} &      & 0.108 & 7.386 & \bfseries 31.05 & 34.95 & 22.88 & 47.08 \\
    \tablenum[table-format=3.1]{200} &      & \bfseries 0.107 & 7.197 & 31.40 & 34.75 & 23.02 & 46.95 \\ \midrule
                                 
    \tablenum[table-format=3.1]{10}  & {\multirow{4}{*}{\tablenum[table-format=1.3,round-precision=1]{0.4}}} & 0.113 & 7.773 & 32.79 & 35.53 & 23.98 & 47.63 \\
    \tablenum[table-format=3.1]{100} &      & \bfseries 0.111 & \bfseries 7.365 & 32.97 & \bfseries 34.63 & \bfseries 23.92 & \bfseries 47.29 \\
    \tablenum[table-format=3.1]{200} &      & 0.114 & 7.435 & 33.35 & 35.56 & 24.30 & 48.25 \\
    \tablenum[table-format=3.1]{300} &      & 0.112 & 7.833 & \bfseries 31.97 & 35.20 & 25.39 & 48.48 \\
    \bottomrule 
\end{tabular*}
\caption{\textbf{Grid search results on the two hyper-parameters, $\lambda_\text{sf\_sm}$ and $\lambda_\text{sf\_pt}$} based on the accuracy of monocular depth, optical flow, and scene flow. The 3D point reconstruction parameter $\lambda_\text{sf\_pt}$ contributes to more accurate disparity information on the target frame, \emph{D2-all}, yielding more accurate scene flow \emph{SF1-all} in the end. The overall results are not very sensitive to the choice of the 3D smoothness parameter $\lambda_\text{sf\_sm}$.}
\label{table:hyperparameter}
\end{table}

\cref{table:hyperparameter} gives the grid search results regarding the two hyper-parameters, reporting the accuracy for monocular depth, optical flow, and scene flow. 
In the upper half of the table, we fix the smoothness parameter $\lambda_\text{sf\_sm}$ and control the 3D point reconstruction loss parameter $\lambda_\text{sf\_pt}$ to see its effect on the accuracy. 
The bottom half of the table is set up the other way around.
Note that the lower the better for all metrics.

We find that $\lambda_\text{sf\_pt}$ is important for best scene flow accuracy, specifically settings that yield accurate disparity information on the target frame, \emph{D2-all}.
This observation follows our design of the 3D point reconstruction loss, which penalizes the 3D distance between corresponding points, encouraging more accurate 3D scene flow in 3D space. 
However, as a trade-off, having a higher value of $\lambda_\text{sf\_pt}$ leads to lower accuracy for 2D estimation, \ie of depth and optical flow. 
On the other hand, we find that the parameter for the 3D smoothness loss, $\lambda_\text{sf\_sm}$, does not strongly affect the accuracy in general.
That is, once $\lambda_\text{sf\_pt}$ is in the right range, the results are not particularly sensitive to the parameter choice.

\section{In-Depth Analysis of the Decoder Design}
\label{sec:splitting_decoder}

With the decoder ablation study in \cref{table:ablation_decoder} of the main paper, we demonstrate that having separate decoders for disparity and scene flow yields instable, unbalanced outputs in contrast to having our proposed single decoder design.
For a more comprehensive analysis, we conduct an empirical study by gradually splitting the decoder consisting of 5 convolution layers and studying the behavior of the networks for each configuration. 
Our backbone network, PWC-Net \cite{Sun:2018:PWC}, has context networks at the end of the decoder, which are fed the output and the last feature map from the decoder as input and perform post-processing for better accuracy.
In our splitting study, we also separate the context networks for each separated decoder so that the two decoders at the end of the networks do not share information.

\begin{figure*}[t]
\centering
\subcaptionbox*{}{\includegraphics[width=0.15\linewidth]{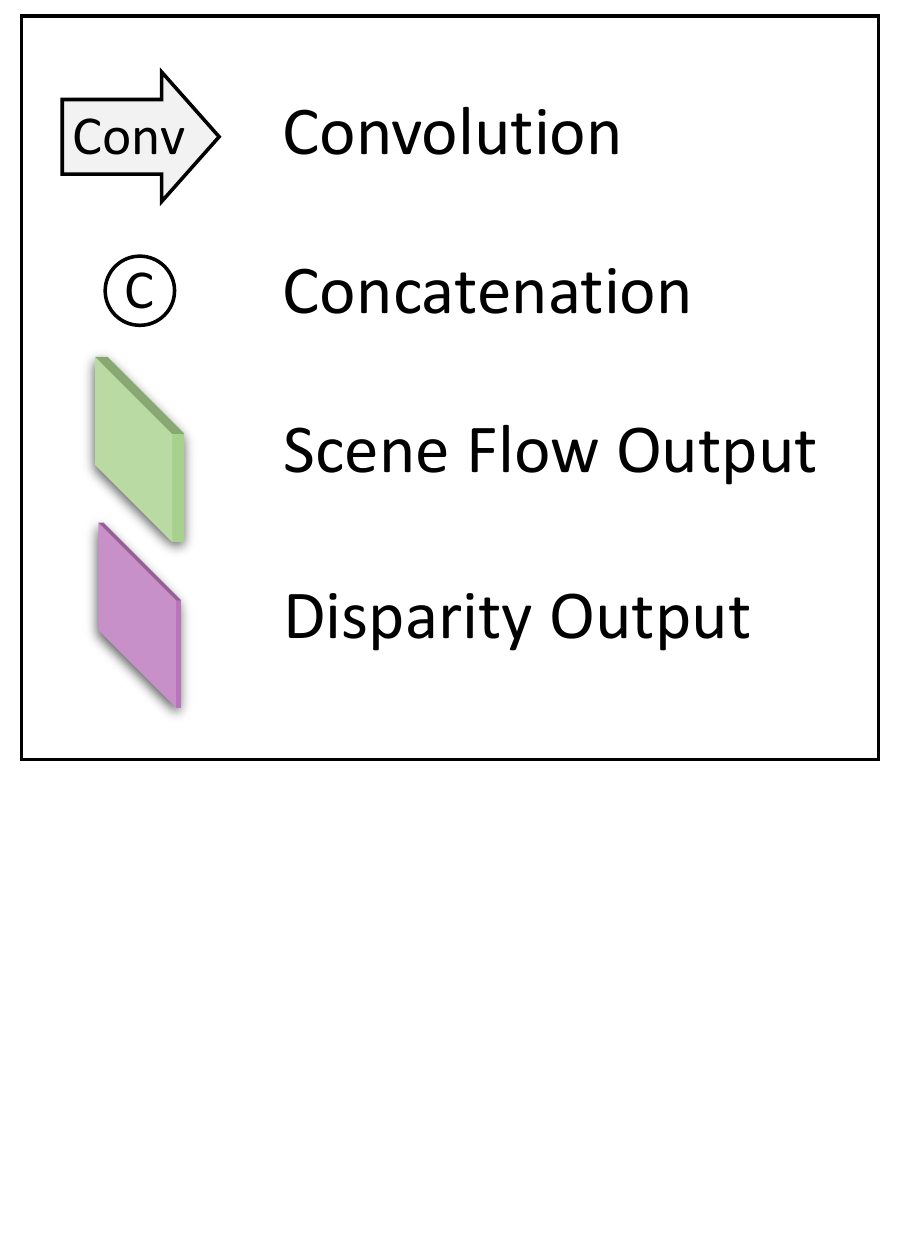}} \quad
\subcaptionbox{Our single decoder design. \label{fig:decoder_split_single}}{\includegraphics[width=0.365\linewidth]{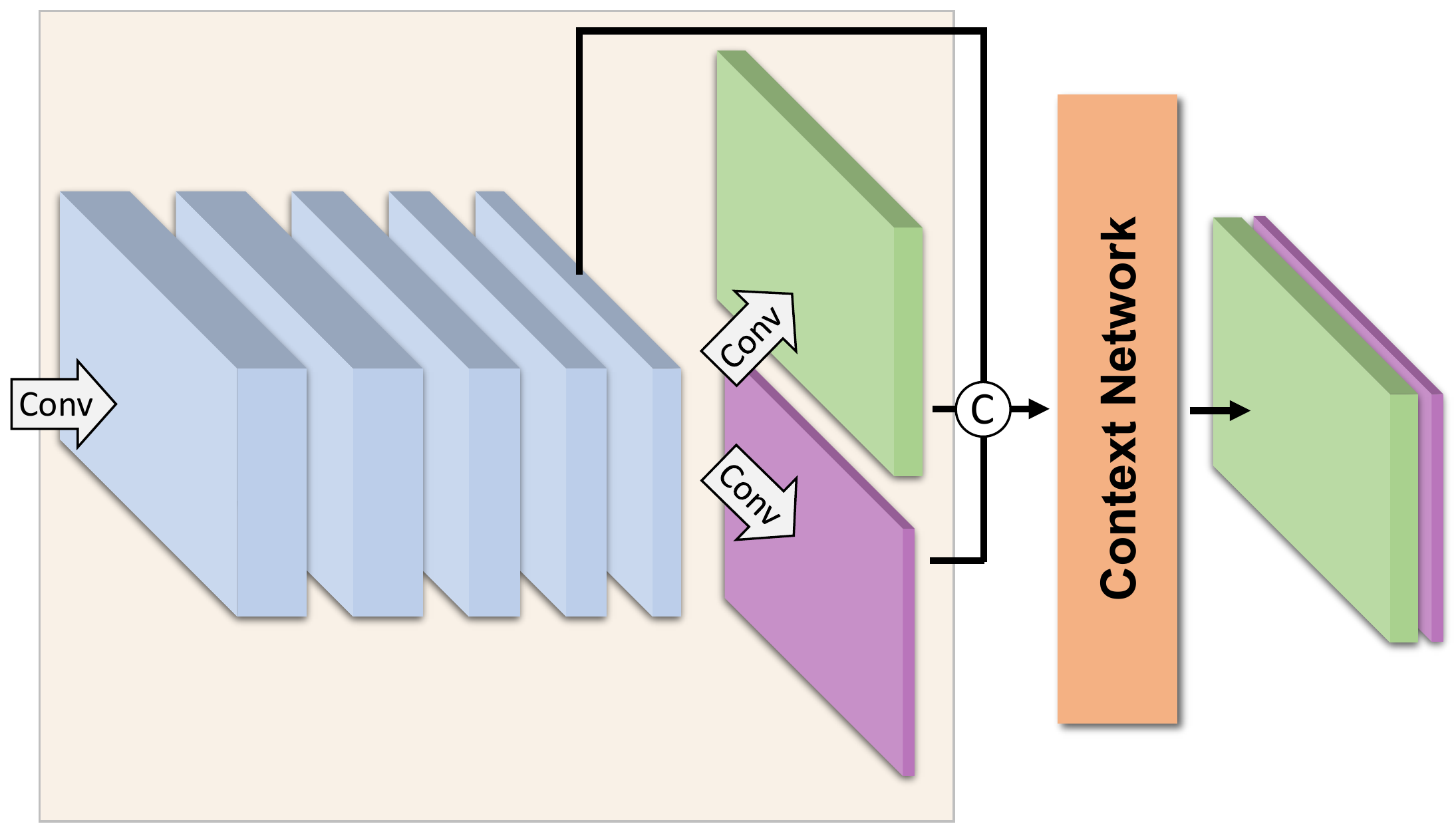}} \quad
\subcaptionbox{Splitting the context network. \label{fig:decoder_split_cont}}{\includegraphics[width=0.34\linewidth]{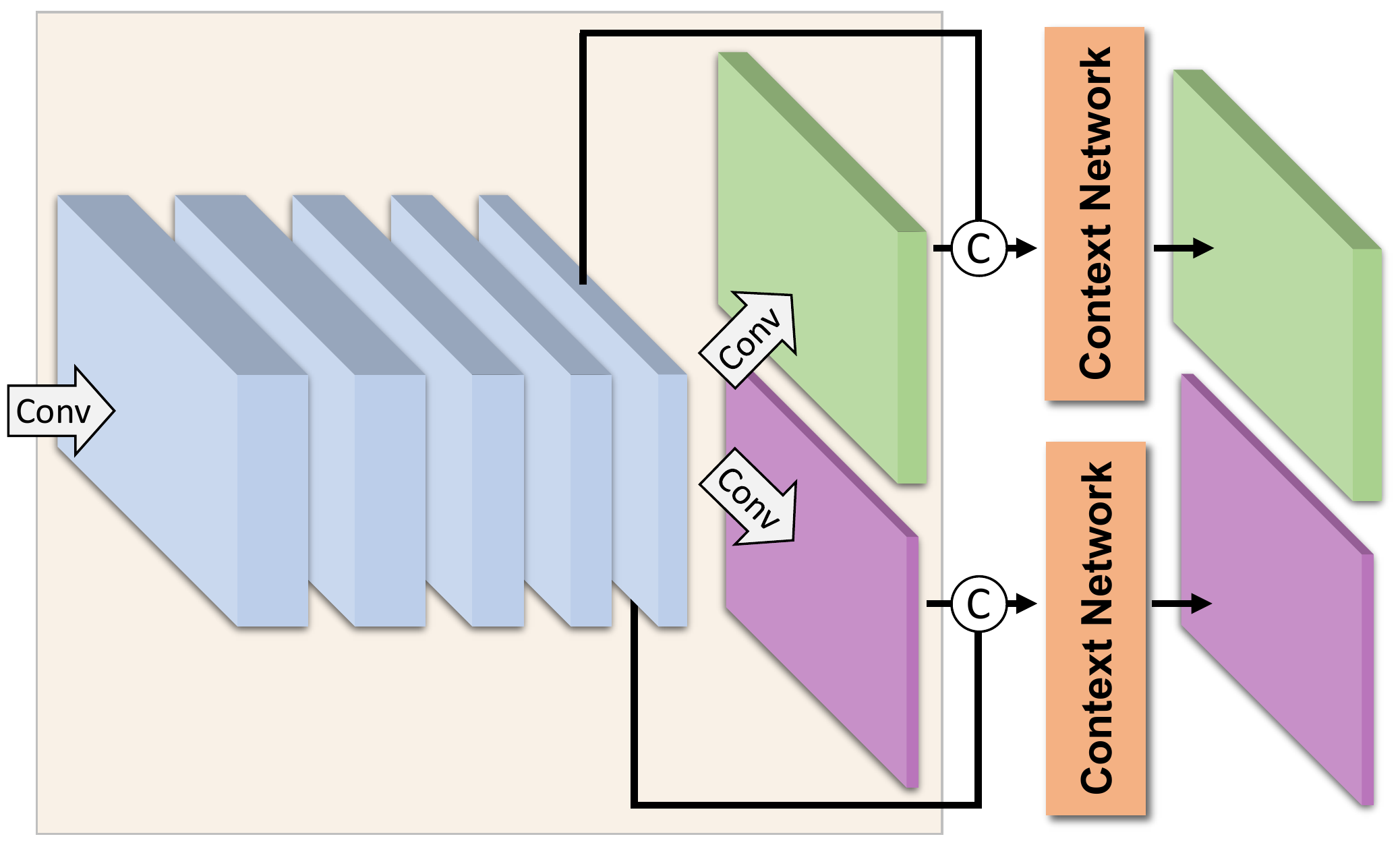}} \\ [1em]
\subcaptionbox{Splitting from the last layer. \label{fig:decoder_split1}}{\includegraphics[width=0.33\linewidth]{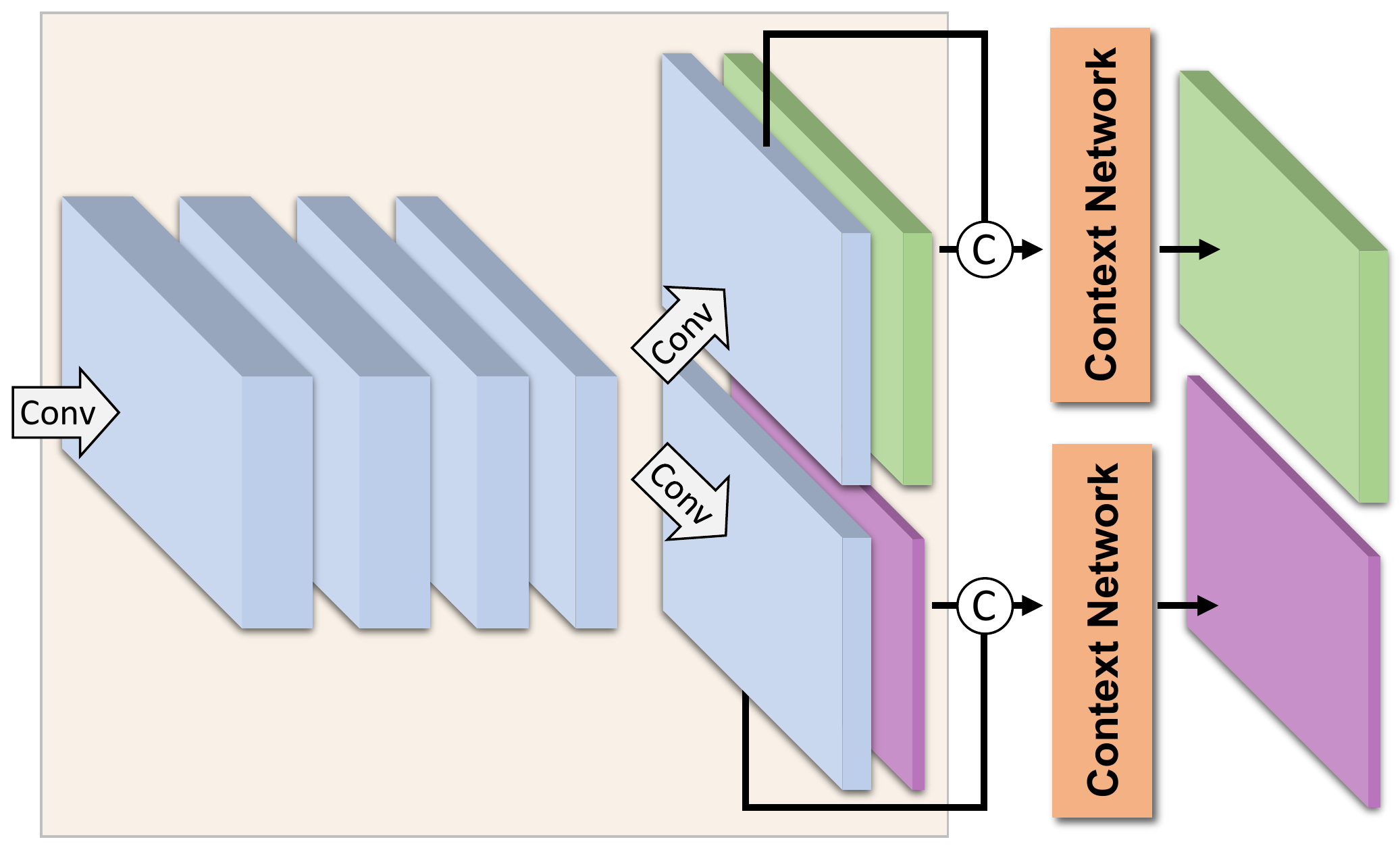}} \quad
\subcaptionbox{Splitting from the \nth{2}-to-last layer. \label{fig:decoder_split2}}{\includegraphics[width=0.33\linewidth]{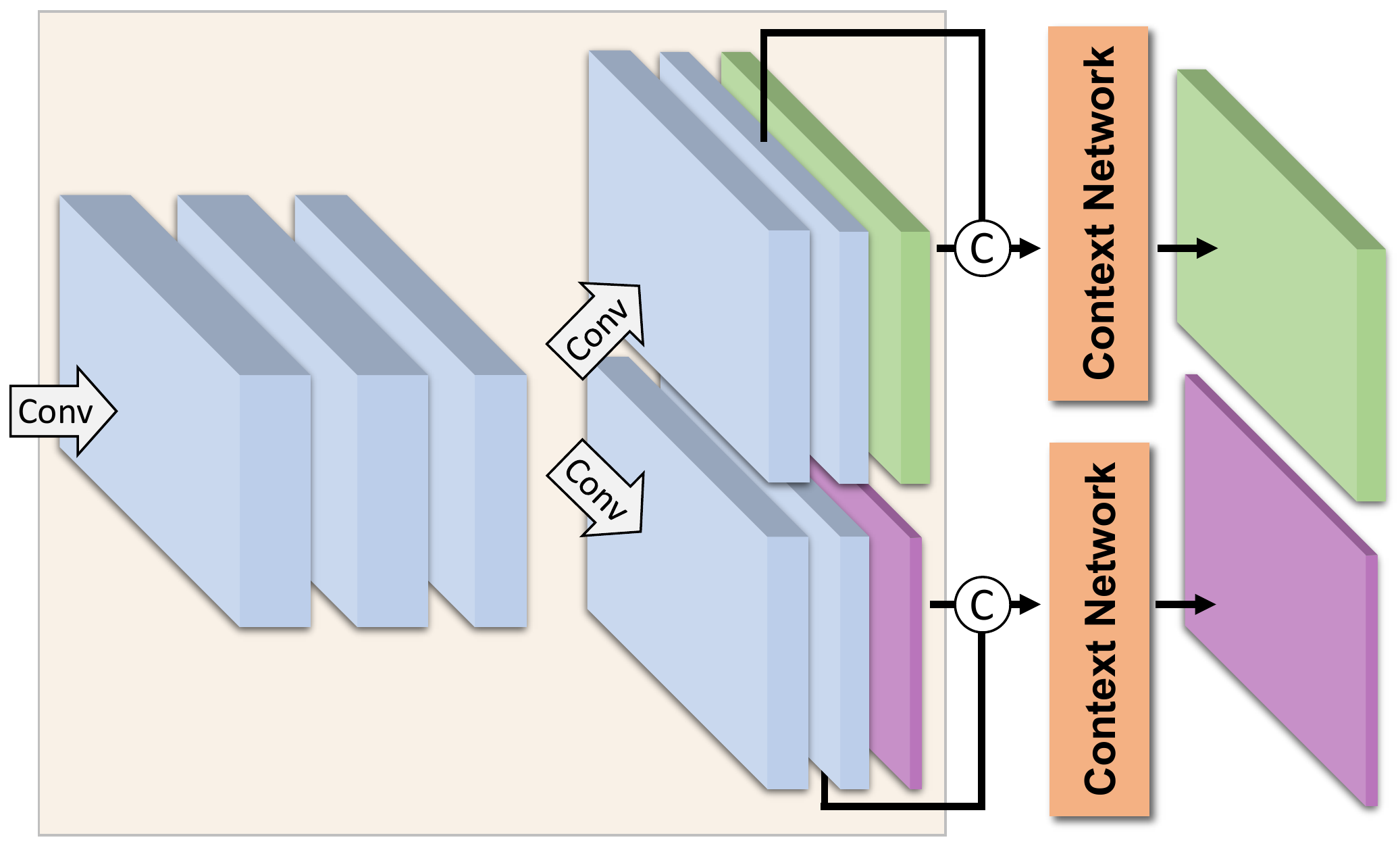}} \quad
\subcaptionbox{Splitting into two separate decoders. \label{fig:decoder_split5}}{\includegraphics[width=0.275\linewidth]{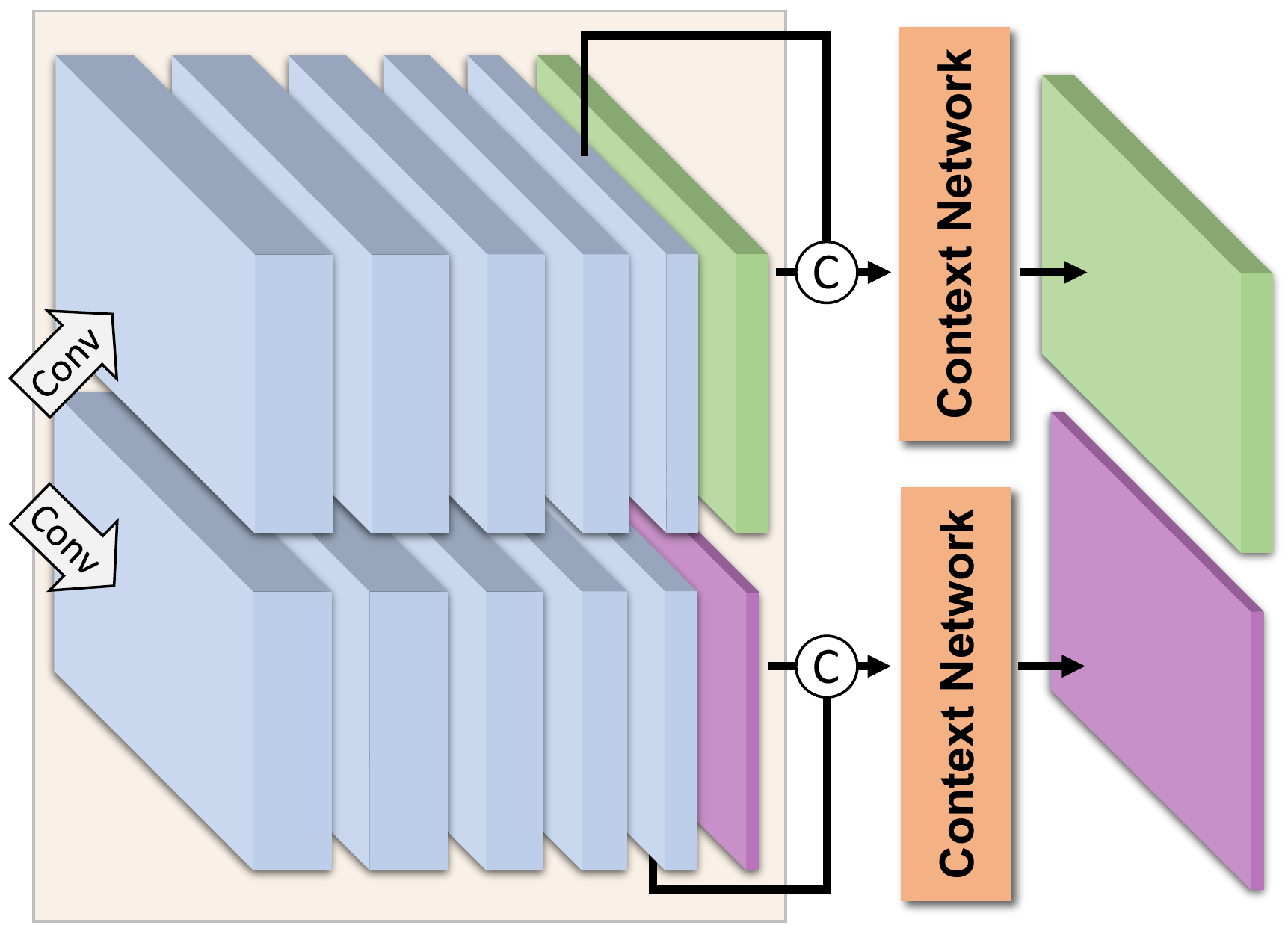}} 
\caption{\textbf{Gradually splitting the single decoder into two separate decoders}: we gradually split the single decoder \emph{(a)} by first splitting the context network \emph{(b)}, and then splitting from the last layer of the decoder \emph{(c)}, the \nth{2}-to-last layer \emph{(d)}, and so on until completely splitting into two separate decoders\emph{(e)}. For ease of visualization, we omit showing the \emph{convolution} operation between the neighboring feature maps in the decoder.}
\label{fig:splitting}
\end{figure*}

\cref{fig:splitting} illustrates each configuration.
From our single decoder design in \cref{fig:decoder_split_single}, we first split the context network for disparity and scene flow respectively, as shown in \cref{fig:decoder_split_cont}.
Then, we begin to split the decoder from the last convolution layer (\ie, \cref{fig:decoder_split1}), the \nth{2}-to-last layer (\ie, \cref{fig:decoder_split2}), and so on until eventually completely splitting into two separate decoders (\ie, \cref{fig:decoder_split5}).
To ensure the same network capacity, we adjust the number of filters so that all configurations have network parameter numbers in a similar range.
All configurations are trained on the \emph{KITTI Split} of KITTI raw \cite{Geiger:2013:VMR} in our self-supervised manner.

\cref{table:spliting_decoders} shows the disparity, optical flow, and scene flow accuracy of each configuration on \emph{KITTI Scene Flow Training} \cite{Menze:2015:J3E,Menze:2018:OSF}. 
We first observe that splitting the context network yields a significant \num{32.73}\% decrease in scene flow accuracy (\ie, SF$1$-all), which mainly stems from the less accurate disparity estimates (\ie, D$1$-all and D$2$-all) although the optical flow accuracy remains almost the same.
This provides an important outlook: given the same optical flow accuracy, the scene flow accuracy depends crucially on how well one can decompose the optical flow cost volume into depth and scene flow, where using the single decoder model works better.
When further splitting the decoder starting from the last convolution layer, the networks \emph{(i)} cannot be trained stably anymore, \emph{(ii)} output trivial solutions for the disparity, and \emph{(iii)} even decrease the optical flow accuracy.
This observation again confirms the benefits of using our proposed single decoder design in terms of both accuracy and training stability.

\begin{table}[t]
\centering
\footnotesize
\begin{tabularx}{\columnwidth}{@{}XS[table-format=3.2]@{\hskip 0.6em}S[table-format=3.2]@{\hskip 0.6em}S[table-format=3.2]@{\hskip 0.6em}S[table-format=3.2]@{}}
	\toprule
	Configuration & {D$1$-all} & {D$2$-all} & {F$1$-all} & {SF$1$-all} \\  
	\midrule 
	Single decoder & \bfseries 31.25 & \bfseries 34.86 & \bfseries 23.49 & \bfseries 47.05 \\
	\midrule
	Splitting the context network & 44.19 & 45.02 & 23.51 & 62.45 \\
	Splitting at the last layer & 100 & 97.22 & 26.46 & 100 \\
	Splitting at the \nth{2}-to-last layer & 100 & 97.22 & 26.39 & 100 \\
	Splitting at the \nth{3}-to-last layer & 100 & 97.22 & 26.94 & 100 \\
	Splitting at the \nth{4}-to-last layer & 100 & 97.22 & 28.68 & 100 \\
	\midrule
	Splitting into two separate decoders & 100 & 97.22 & 27.63 & 100 \\
	\bottomrule	
\end{tabularx}
\caption{\textbf{Scene flow accuracy of each decoder configuration}: splitting the context network already decreases the scene flow accuracy by \num{32.73}\%. Further splitting the decoder yields training instability with trivial solutions for the disparity output.}
\label{table:spliting_decoders}
\vspace{-0.5em}
\end{table}

\section{Qualitative Analysis of Loss Ablation Study}
\label{sec:qualitative_loss_ablation}

\cref{table:ablation_loss} in the main paper provides an ablation study of our self-supervised proxy loss. 
For better understanding of how each loss term affects the results, we provide qualitative examples of disparity, optical flow, and scene flow estimation.
\cref{fig:qualitative_ablation} displays the results for each loss configuration: \emph{(a)} the basic loss where only the brightness and smoothness terms are active; \emph{(b)} with occlusion handling, which discards occluded pixels in the loss; \emph{(c)} with the 3D point reconstruction loss; and \emph{(d)} the full loss.
Each configuration is trained in the proposed self-supervised manner using the \emph{KITTI Split} and evaluated on \emph{KITTI Scene Flow Training} \cite{Menze:2015:J3E,Menze:2018:OSF}.

{
\begin{figure*}[b]
\centering
\scriptsize
\setlength\tabcolsep{0.3pt}
\renewcommand{\arraystretch}{0.2}
\begin{tabular}{c@{\hskip 0.5em} >{\centering\arraybackslash}m{.19\textwidth} >{\centering\arraybackslash}m{.19\textwidth} >{\centering\arraybackslash}m{.19\textwidth} >{\centering\arraybackslash}m{.19\textwidth} >{\centering\arraybackslash}m{.19\textwidth}}
	& Reference image & Target image \\ 
	& \includegraphics[width=\linewidth]{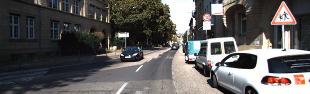} & 
	\includegraphics[width=\linewidth]{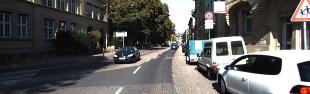} \\ \\[0.3em]
	& \emph{(a) Basic} & \emph{(b) With occlusion handling} & \emph{(c) With 3D point loss} & \textbf{\emph{(d) Full loss (Self-Mono-SF)}} & \emph{(e) Ground truth} \\

	\rotatebox[origin=c]{90}{D1} &
	\begin{tikzpicture}
    \node[inner sep=0] (img) {\includegraphics[width=\linewidth]{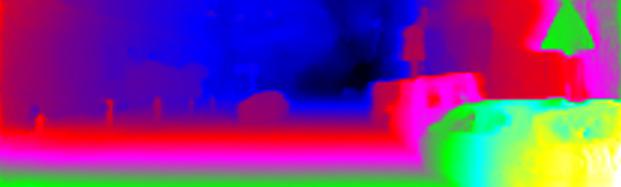}};
	\end{tikzpicture}&
	\begin{tikzpicture}
    \node[inner sep=0] (img) {\includegraphics[width=\linewidth]{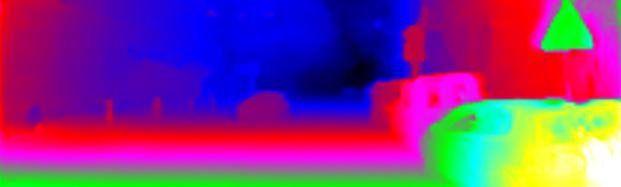}};
	\end{tikzpicture}&
	\begin{tikzpicture}
    \node[inner sep=0] (img) {\includegraphics[width=\linewidth]{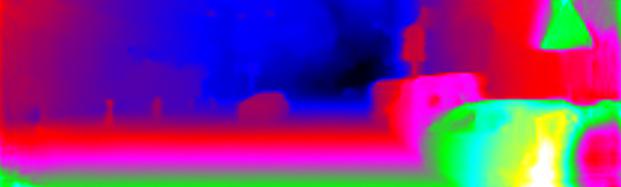}};
	\end{tikzpicture}&
	\begin{tikzpicture}
    \node[inner sep=0] (img) {\includegraphics[width=\linewidth]{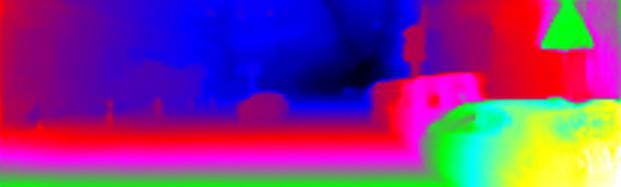}};
	\end{tikzpicture}&
	\begin{tikzpicture}
    \node[inner sep=0] (img) {\includegraphics[width=\linewidth]{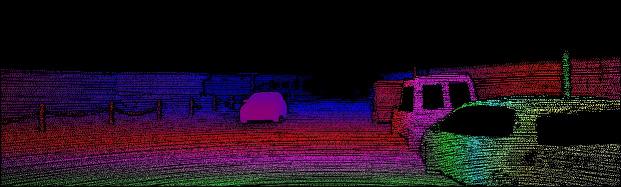}};
	\end{tikzpicture}\\

	\rotatebox[origin=c]{90}{D1 Error} &
	\begin{tikzpicture}
    \node[inner sep=0] (img) {\includegraphics[width=\linewidth]{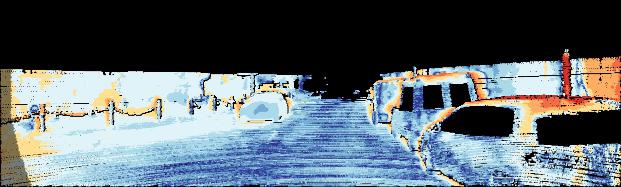}};
    \node[anchor=north east] at (img.north east){ \tiny \color{white} \textbf{11.76 \%}};
	\end{tikzpicture}&
	\begin{tikzpicture}
    \node[inner sep=0] (img) {\includegraphics[width=\linewidth]{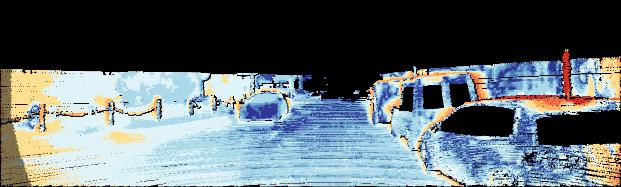}};
    \node[anchor=north east] at (img.north east){ \tiny \color{white} \textbf{12.66 \%}};
	\end{tikzpicture}&
	\begin{tikzpicture}
    \node[inner sep=0] (img) {\includegraphics[width=\linewidth]{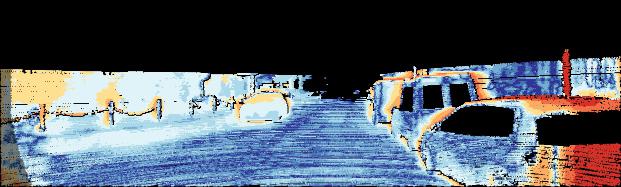}};
    \node[anchor=north east] at (img.north east){ \tiny \color{white} \textbf{12.48 \%}};
	\end{tikzpicture}&
	\begin{tikzpicture}
    \node[inner sep=0] (img) {\includegraphics[width=\linewidth]{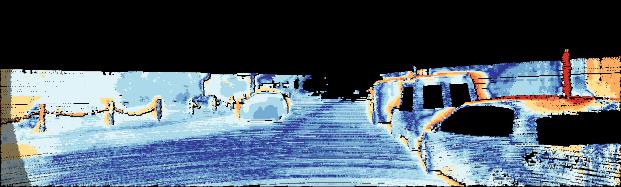}};
    \node[anchor=north east] at (img.north east){ \tiny \color{white} \textbf{7.99 \%}};
	\end{tikzpicture} \\
	
	\rotatebox[origin=c]{90}{D2} &
	\begin{tikzpicture}
    \node[inner sep=0] (img) {\includegraphics[width=\linewidth]{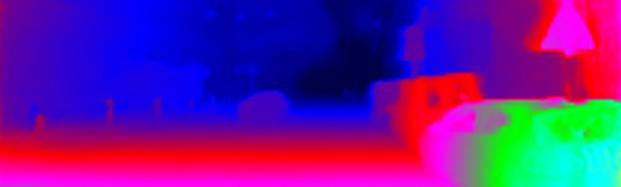}};
	\end{tikzpicture}&
	\begin{tikzpicture}
    \node[inner sep=0] (img) {\includegraphics[width=\linewidth]{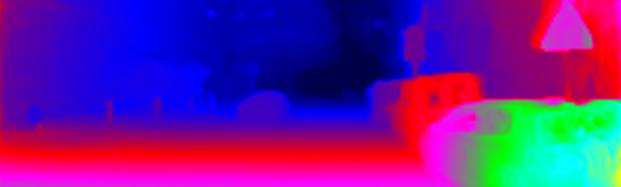}};
	\end{tikzpicture}&
	\begin{tikzpicture}
    \node[inner sep=0] (img) {\includegraphics[width=\linewidth]{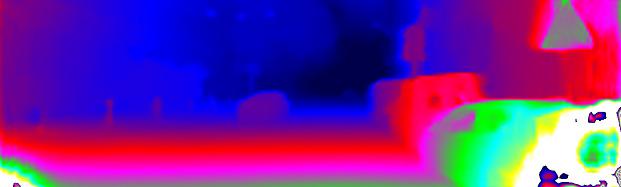}};
	\end{tikzpicture}&
	\begin{tikzpicture}
    \node[inner sep=0] (img) {\includegraphics[width=\linewidth]{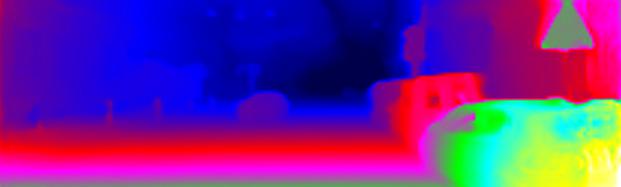}};
	\end{tikzpicture}&
	\begin{tikzpicture}
	\node[inner sep=0] (img) {\includegraphics[width=\linewidth]{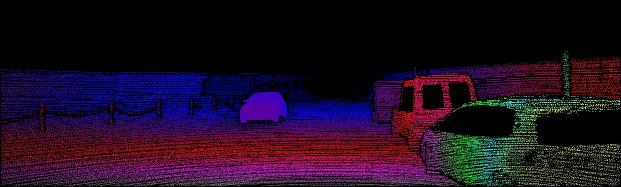}};
	\end{tikzpicture} \\

	\rotatebox[origin=c]{90}{D2 Error} &
	\begin{tikzpicture}
    \node[inner sep=0] (img) {\includegraphics[width=\linewidth]{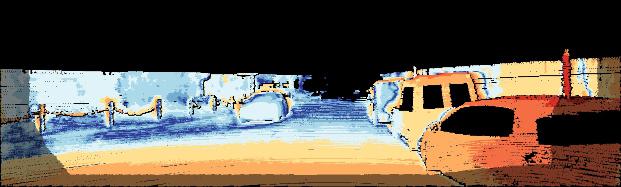}};
    \node[anchor=north east] at (img.north east){ \tiny \color{white} \textbf{49.99 \%}};
	\end{tikzpicture}&
	\begin{tikzpicture}
    \node[inner sep=0] (img) {\includegraphics[width=\linewidth]{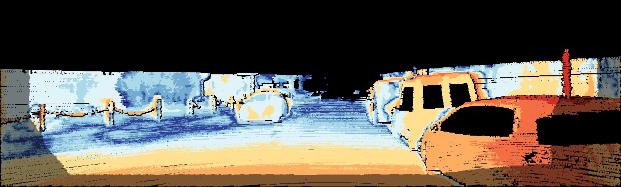}};
    \node[anchor=north east] at (img.north east){ \tiny \color{white} \textbf{46.13 \%}};
	\end{tikzpicture}&
	\begin{tikzpicture}
    \node[inner sep=0] (img) {\includegraphics[width=\linewidth]{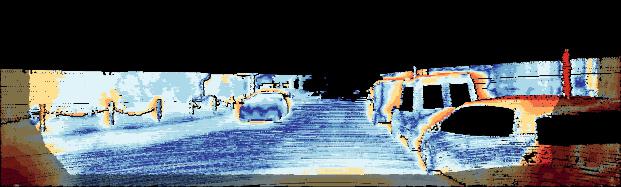}};
    \node[anchor=north east] at (img.north east){ \tiny \color{white} \textbf{19.55 \%}};
	\end{tikzpicture}&
	\begin{tikzpicture}
    \node[inner sep=0] (img) {\includegraphics[width=\linewidth]{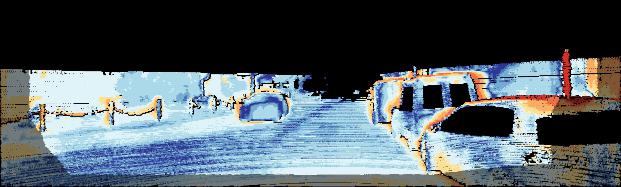}};
    \node[anchor=north east] at (img.north east){ \tiny \color{white} \textbf{11.30 \%}};
	\end{tikzpicture} \\

	\rotatebox[origin=c]{90}{F1} &
	\begin{tikzpicture}
    \node[inner sep=0] (img) {\includegraphics[width=\linewidth]{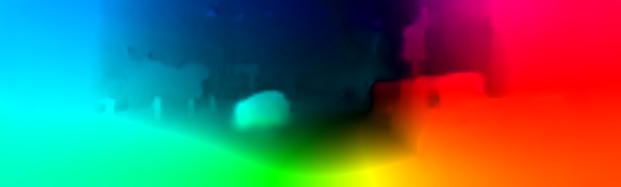}};
	\end{tikzpicture}&
	\begin{tikzpicture}
    \node[inner sep=0] (img) {\includegraphics[width=\linewidth]{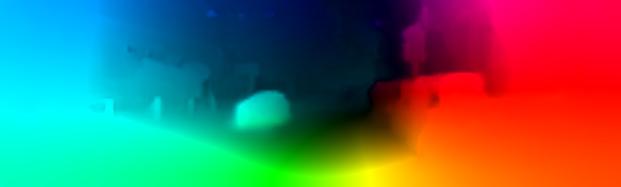}};
	\end{tikzpicture}&
	\begin{tikzpicture}
    \node[inner sep=0] (img) {\includegraphics[width=\linewidth]{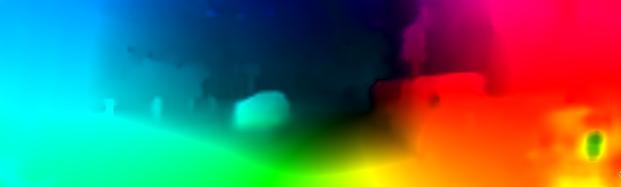}};
	\end{tikzpicture}&
	\begin{tikzpicture}
    \node[inner sep=0] (img) {\includegraphics[width=\linewidth]{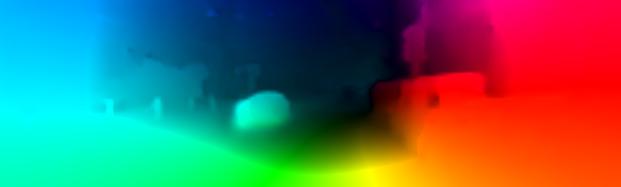}};
	\end{tikzpicture}&
	\begin{tikzpicture}
    \node[inner sep=0] (img) {\includegraphics[width=\linewidth]{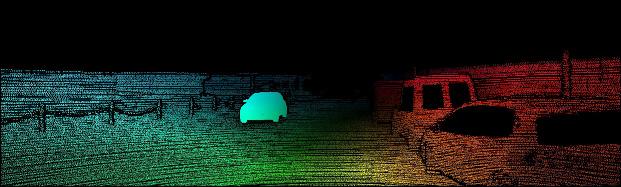}};
	\end{tikzpicture} \\

	\rotatebox[origin=c]{90}{F1 Error} &
	\begin{tikzpicture}
    \node[inner sep=0] (img) {\includegraphics[width=\linewidth]{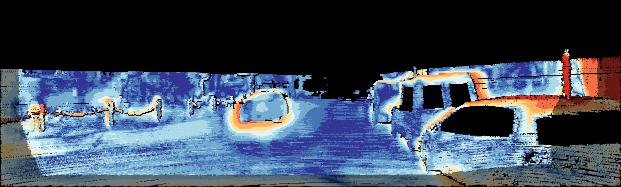}};
    \node[anchor=north east] at (img.north east){ \tiny \color{white} \textbf{8.52 \%}};
	\end{tikzpicture}&
	\begin{tikzpicture}
    \node[inner sep=0] (img) {\includegraphics[width=\linewidth]{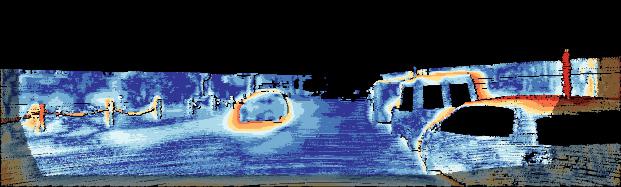}};
    \node[anchor=north east] at (img.north east){ \tiny \color{white} \textbf{8.28 \%}};
	\end{tikzpicture}&
	\begin{tikzpicture}
    \node[inner sep=0] (img) {\includegraphics[width=\linewidth]{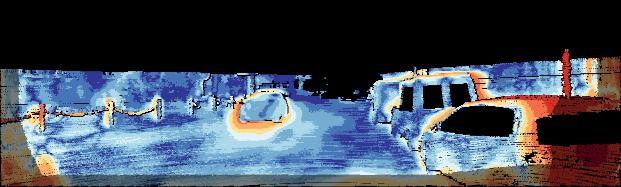}};
    \node[anchor=north east] at (img.north east){ \tiny \color{white} \textbf{14.41 \%}};
	\end{tikzpicture}&
	\begin{tikzpicture}
    \node[inner sep=0] (img) {\includegraphics[width=\linewidth]{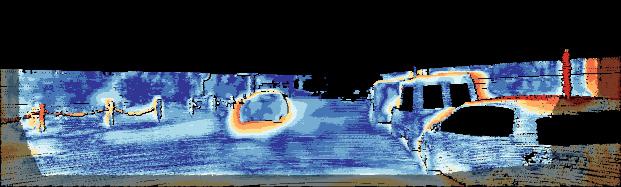}};
    \node[anchor=north east] at (img.north east){ \tiny \color{white} \textbf{10.33 \%}};
	\end{tikzpicture} \\

	\rotatebox[origin=c]{90}{SF1 Error} &
	\begin{tikzpicture}
    \node[inner sep=0] (img) {\includegraphics[width=\linewidth]{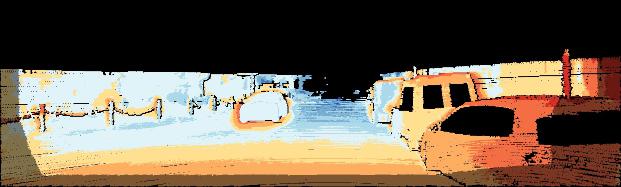}};
    \node[anchor=north east] at (img.north east){ \tiny \color{white} \textbf{55.87 \%}};
	\end{tikzpicture}&
	\begin{tikzpicture}
    \node[inner sep=0] (img) {\includegraphics[width=\linewidth]{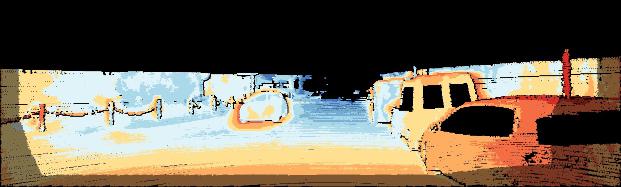}};
    \node[anchor=north east] at (img.north east){ \tiny \color{white} \textbf{54.50 \%}};
	\end{tikzpicture}&
	\begin{tikzpicture}
    \node[inner sep=0] (img) {\includegraphics[width=\linewidth]{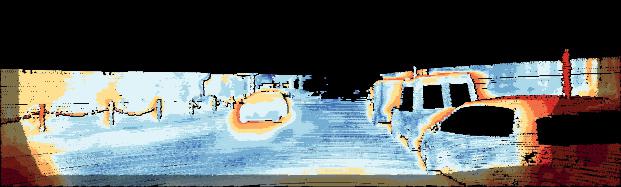}};
    \node[anchor=north east] at (img.north east){ \tiny \color{white} \textbf{24.32 \%}};
	\end{tikzpicture}&
	\begin{tikzpicture}
    \node[inner sep=0] (img) {\includegraphics[width=\linewidth]{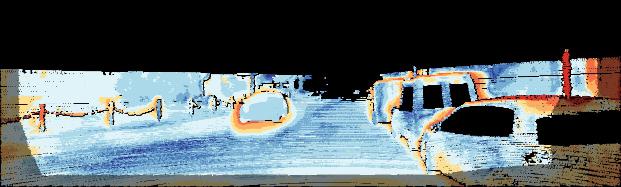}};
    \node[anchor=north east] at (img.north east){ \tiny \color{white} \textbf{15.07 \%}};
	\end{tikzpicture}  \\ 

	\\[0.8em]

	& Reference image & Target image \\ 
	& \includegraphics[width=\linewidth]{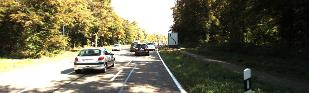} & 
	\includegraphics[width=\linewidth]{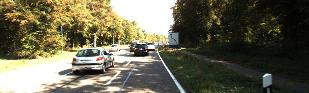} \\ \\[0.3em]
	& \emph{(a) Basic} & \emph{(b) With occlusion handling} & \emph{(c) With 3D point loss} & \textbf{\emph{(d) Full loss (Self-Mono-SF)}} & \emph{(e) Ground truth} \\
	\rotatebox[origin=c]{90}{D1} &
	\begin{tikzpicture}
    \node[inner sep=0] (img) {\includegraphics[width=\linewidth]{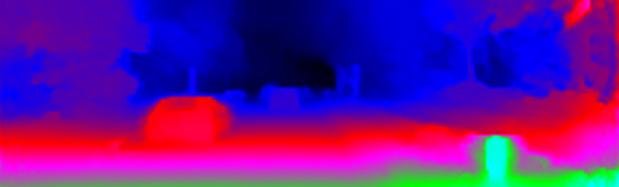}};
	\end{tikzpicture}&
	\begin{tikzpicture}
    \node[inner sep=0] (img) {\includegraphics[width=\linewidth]{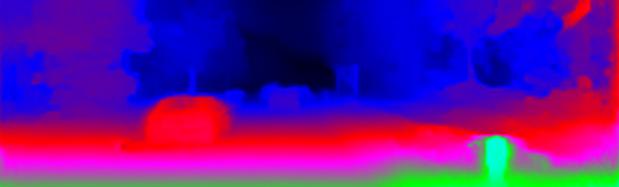}};
	\end{tikzpicture}&
	\begin{tikzpicture}
    \node[inner sep=0] (img) {\includegraphics[width=\linewidth]{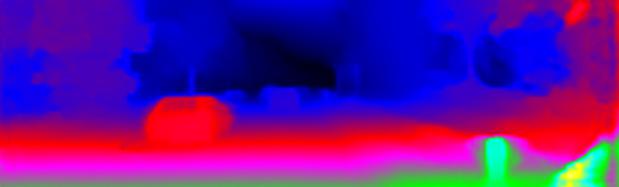}};
	\end{tikzpicture}&
	\begin{tikzpicture}
    \node[inner sep=0] (img) {\includegraphics[width=\linewidth]{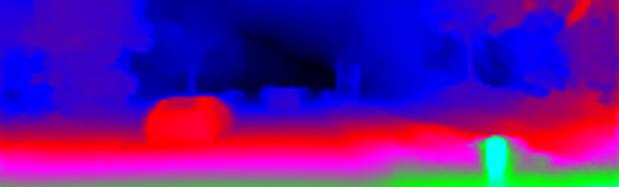}};
	\end{tikzpicture}&
	\begin{tikzpicture}
    \node[inner sep=0] (img) {\includegraphics[width=\linewidth]{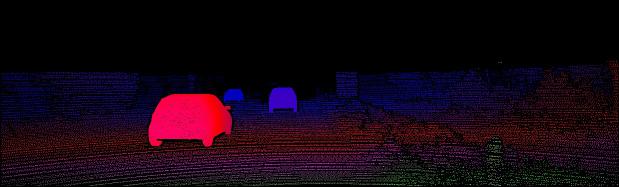}};
	\end{tikzpicture}\\

	\rotatebox[origin=c]{90}{D1 Error} &
	\begin{tikzpicture}
    \node[inner sep=0] (img) {\includegraphics[width=\linewidth]{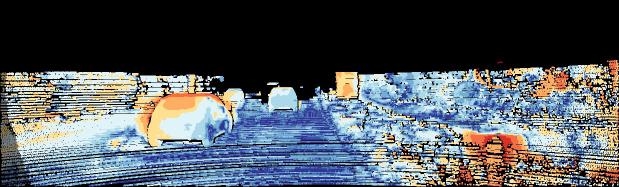}};
    \node[anchor=north east] at (img.north east){ \tiny \color{white} \textbf{22.73 \%}};
	\end{tikzpicture}&
	\begin{tikzpicture}
    \node[inner sep=0] (img) {\includegraphics[width=\linewidth]{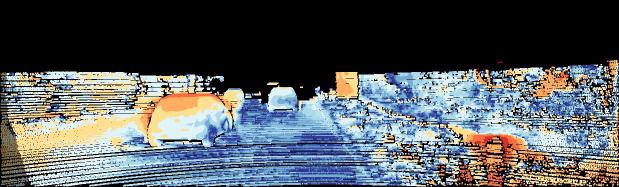}};
    \node[anchor=north east] at (img.north east){ \tiny \color{white} \textbf{26.45 \%}};
	\end{tikzpicture}&
	\begin{tikzpicture}
    \node[inner sep=0] (img) {\includegraphics[width=\linewidth]{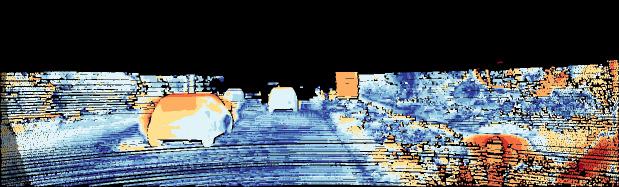}};
    \node[anchor=north east] at (img.north east){ \tiny \color{white} \textbf{25.07 \%}};
	\end{tikzpicture}&
	\begin{tikzpicture}
    \node[inner sep=0] (img) {\includegraphics[width=\linewidth]{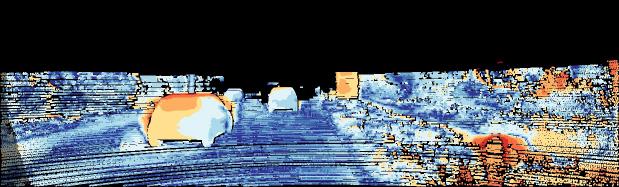}};
    \node[anchor=north east] at (img.north east){ \tiny \color{white} \textbf{22.41 \%}};
	\end{tikzpicture} \\
	
	\rotatebox[origin=c]{90}{D2} &
	\begin{tikzpicture}
    \node[inner sep=0] (img) {\includegraphics[width=\linewidth]{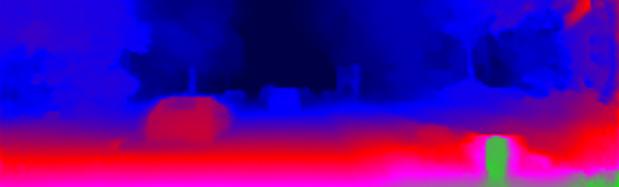}};
	\end{tikzpicture}&
	\begin{tikzpicture}
    \node[inner sep=0] (img) {\includegraphics[width=\linewidth]{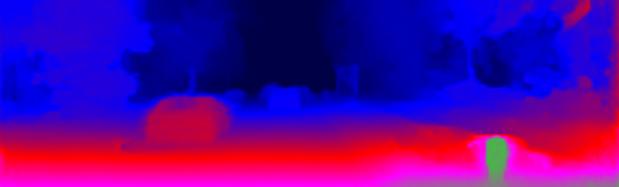}};
	\end{tikzpicture}&
	\begin{tikzpicture}
    \node[inner sep=0] (img) {\includegraphics[width=\linewidth]{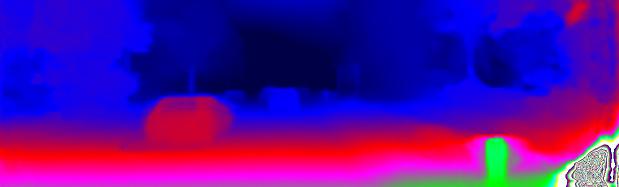}};
	\end{tikzpicture}&
	\begin{tikzpicture}
    \node[inner sep=0] (img) {\includegraphics[width=\linewidth]{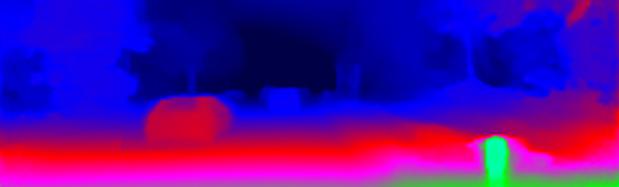}};
	\end{tikzpicture}&
	\begin{tikzpicture}
	\node[inner sep=0] (img) {\includegraphics[width=\linewidth]{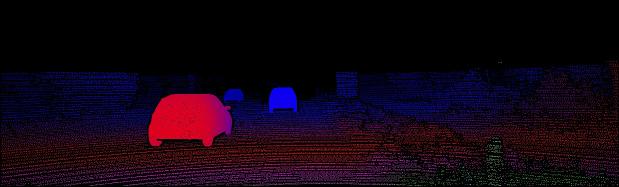}};
	\end{tikzpicture} \\

	\rotatebox[origin=c]{90}{D2 Error} &
	\begin{tikzpicture}
    \node[inner sep=0] (img) {\includegraphics[width=\linewidth]{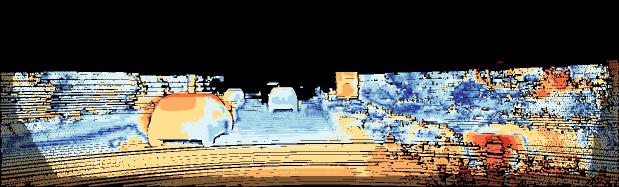}};
    \node[anchor=north east] at (img.north east){ \tiny \color{white} \textbf{47.15 \%}};
	\end{tikzpicture}&
	\begin{tikzpicture}
    \node[inner sep=0] (img) {\includegraphics[width=\linewidth]{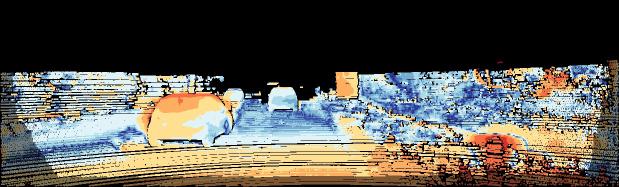}};
    \node[anchor=north east] at (img.north east){ \tiny \color{white} \textbf{46.66 \%}};
	\end{tikzpicture}&
	\begin{tikzpicture}
    \node[inner sep=0] (img) {\includegraphics[width=\linewidth]{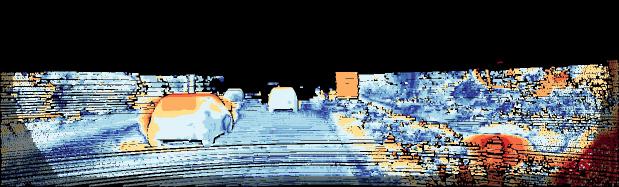}};
    \node[anchor=north east] at (img.north east){ \tiny \color{white} \textbf{22.21 \%}};
	\end{tikzpicture}&
	\begin{tikzpicture}
    \node[inner sep=0] (img) {\includegraphics[width=\linewidth]{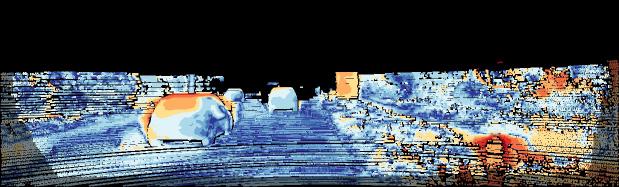}};
    \node[anchor=north east] at (img.north east){ \tiny \color{white} \textbf{18.19 \%}};
	\end{tikzpicture} \\

	\rotatebox[origin=c]{90}{F1} &
	\begin{tikzpicture}
    \node[inner sep=0] (img) {\includegraphics[width=\linewidth]{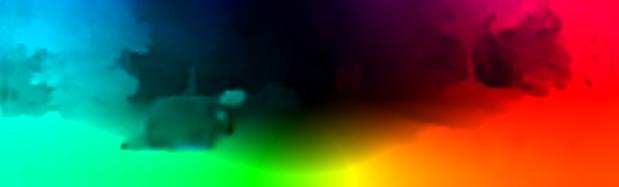}};
	\end{tikzpicture}&
	\begin{tikzpicture}
    \node[inner sep=0] (img) {\includegraphics[width=\linewidth]{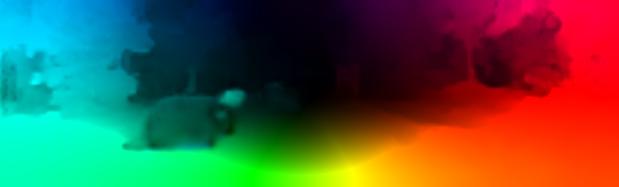}};
	\end{tikzpicture}&
	\begin{tikzpicture}
    \node[inner sep=0] (img) {\includegraphics[width=\linewidth]{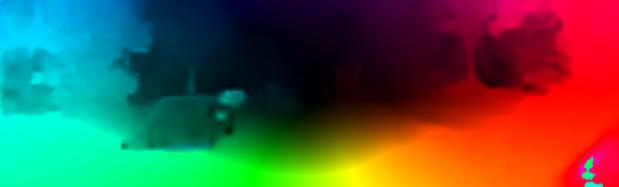}};
	\end{tikzpicture}&
	\begin{tikzpicture}
    \node[inner sep=0] (img) {\includegraphics[width=\linewidth]{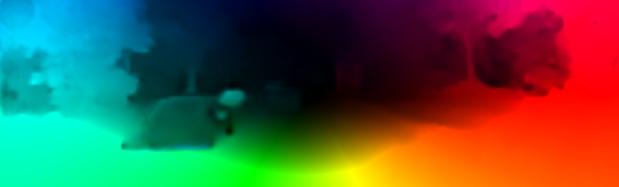}};
	\end{tikzpicture}&
	\begin{tikzpicture}
    \node[inner sep=0] (img) {\includegraphics[width=\linewidth]{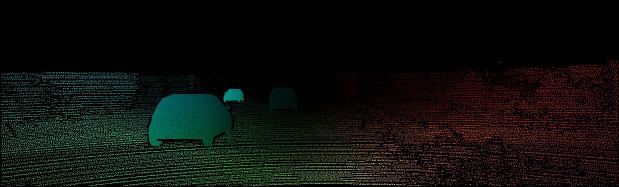}};
	\end{tikzpicture} \\

	\rotatebox[origin=c]{90}{F1 Error} &
	\begin{tikzpicture}
    \node[inner sep=0] (img) {\includegraphics[width=\linewidth]{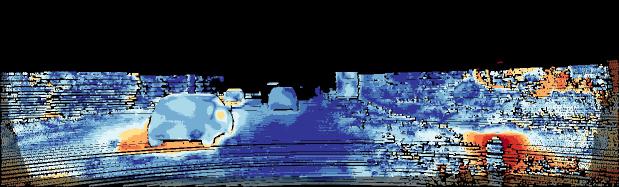}};
    \node[anchor=north east] at (img.north east){ \tiny \color{white} \textbf{9.08 \%}};
	\end{tikzpicture}&
	\begin{tikzpicture}
    \node[inner sep=0] (img) {\includegraphics[width=\linewidth]{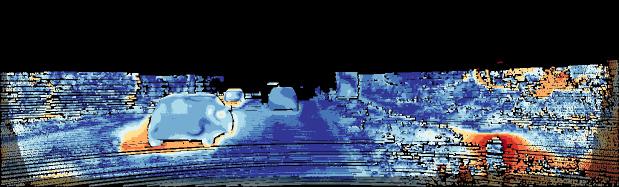}};
    \node[anchor=north east] at (img.north east){ \tiny \color{white} \textbf{8.62 \%}};
	\end{tikzpicture}&
	\begin{tikzpicture}
    \node[inner sep=0] (img) {\includegraphics[width=\linewidth]{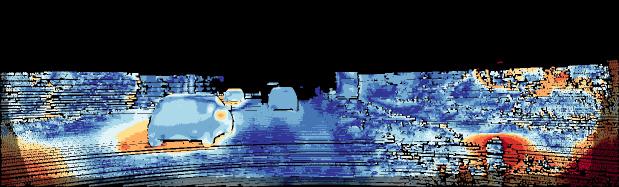}};
    \node[anchor=north east] at (img.north east){ \tiny \color{white} \textbf{11.80 \%}};
	\end{tikzpicture}&
	\begin{tikzpicture}
    \node[inner sep=0] (img) {\includegraphics[width=\linewidth]{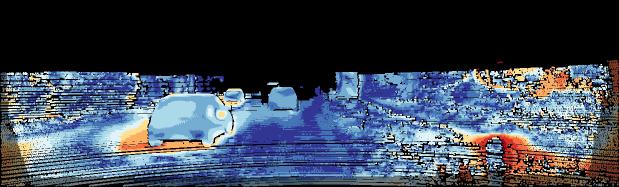}};
    \node[anchor=north east] at (img.north east){ \tiny \color{white} \textbf{9.14 \%}};
	\end{tikzpicture} \\

	\rotatebox[origin=c]{90}{SF1 Error} &
	\begin{tikzpicture}
    \node[inner sep=0] (img) {\includegraphics[width=\linewidth]{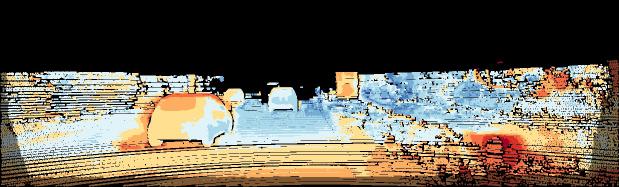}};
    \node[anchor=north east] at (img.north east){ \tiny \color{white} \textbf{52.21 \%}};
	\end{tikzpicture}&
	\begin{tikzpicture}
    \node[inner sep=0] (img) {\includegraphics[width=\linewidth]{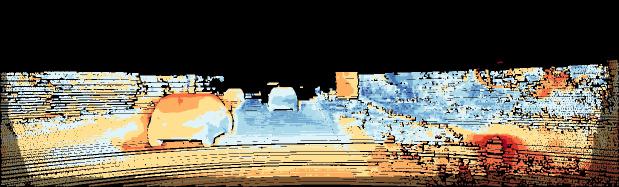}};
    \node[anchor=north east] at (img.north east){ \tiny \color{white} \textbf{54.73 \%}};
	\end{tikzpicture}&
	\begin{tikzpicture}
    \node[inner sep=0] (img) {\includegraphics[width=\linewidth]{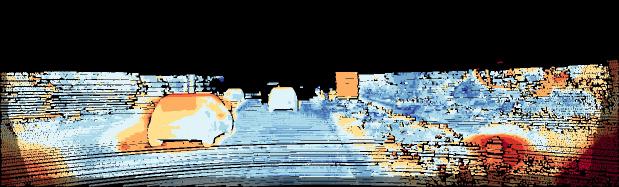}};
    \node[anchor=north east] at (img.north east){ \tiny \color{white} \textbf{31.71 \%}};
	\end{tikzpicture}&
	\begin{tikzpicture}
    \node[inner sep=0] (img) {\includegraphics[width=\linewidth]{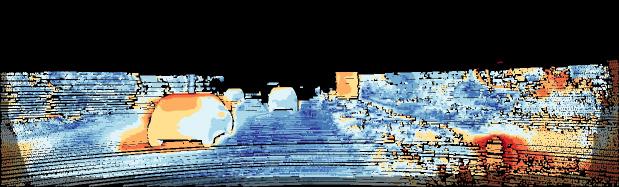}};
    \node[anchor=north east] at (img.north east){ \tiny \color{white} \textbf{27.41 \%}};
	\end{tikzpicture}  \\

\end{tabular}
\caption {\textbf{Qualitative examples on the loss ablation study}. For each scene in the first row we show two input images, the reference and the target image. From the second to the last row, we show a qualitative comparison of each loss configuration: \emph{(a)} basic loss, \emph{(b)} with occlusion handling, \emph{(c)} with 3D point reconstruction loss, and the \emph{(d)} our full loss.
Each row visualizes the disparity map of the reference image \emph{(D1)} with its error map \emph{(D1 Error)}, disparity estimation at the target image mapped into the reference frame \emph{(D2)} along with its error map \emph{(D2 Error)}, optical flow \emph{(F1)} with its error map \emph{(F1 Error)}, and the scene flow error map \emph{(SF1 Error)}. 
The outlier rates are overlayed on each error map. The last column shows \emph{(e)} the ground truth for each estimate.}
\label{fig:qualitative_ablation}
\end{figure*}
}

Without the 3D point reconstruction loss for scene flow (\ie, columns \emph{(a)} and \emph{(b)} in \cref{fig:qualitative_ablation}), the networks output inaccurate disparity information for the target frame \emph{(D2)} especially in the road area, which yields inaccurate scene flow results \emph{(SF1)} in the end.
Applying the 3D point reconstruction loss but without occlusion handling (\ie, column \emph{(c)} in \cref{fig:qualitative_ablation}) results in inaccurate estimates and some artifacts appearing on out-of-bound pixels, still leading to an unsatisfactory final scene flow accuracy. 
These artifacts happen when the 3D point reconstruction loss tries to minimize the 3D Euclidean distance between incorrect pixel correspondences, such as for occlusions or out-of-bound pixels.
Discarding those occluded regions in the proxy loss eventually yields better estimates in the occluded region as well.

\section{Qualitative Comparison}
\label{sec:qualitative_comparison}

We provide some qualitative examples of our monocular scene flow estimation by comparing with the state-of-the-art Mono-SF method \cite{Brickwedde:2019:MSF}, which uses an integrated pipeline of CNNs and an energy-based model. 
\cref{fig:qualitative_successful,fig:qualitative_unsuccessful} show successful qualitative results as well as some failure cases of our fine-tuned model on the KITTI 2015 Scene Flow public benchmark \cite{Menze:2015:J3E,Menze:2018:OSF}, respectively.

In \cref{fig:qualitative_successful}, our model outputs more accurate disparity and optical flow estimation results than Mono-SF \cite{Brickwedde:2019:MSF} without using an explicit planar surface representation or a rigid motion assumption, which would be beneficial for achieving better accuracy on the KITTI 2015 Scene Flow public benchmark. 

\cref{fig:qualitative_unsuccessful}, in contrast, shows some of the failure cases, where our model outputs less accurate results for scene flow estimation than Mono-SF \cite{Brickwedde:2019:MSF}. 
Although our model can estimate optical flow with an accuracy comparable to Mono-SF, inaccurate disparity estimation eventually leads to less accurate scene flow.
The gap in terms of the disparity accuracy of ours \vs Mono-SF \cite{Brickwedde:2019:MSF} can be explained by the fact that Mono-SF exploits over \num{20000} instances of pseudo ground-truth depth data to train their monocular depth model, while our method uses only \num{200} images for fine-tuning.

{
\begin{figure*}[!b]
\centering
\footnotesize
\setlength\tabcolsep{0.3pt}
\renewcommand{\arraystretch}{0.2}
\begin{tabular}{c@{\hskip 0.5em} >{\centering\arraybackslash}m{.24\textwidth} >{\centering\arraybackslash}m{.24\textwidth} @{\hskip 0.4em} >{\centering\arraybackslash}m{.24\textwidth} >{\centering\arraybackslash}m{.24\textwidth}}
	& Reference image & Target image & Reference image & Target image \\ 
	& \includegraphics[width=\linewidth]{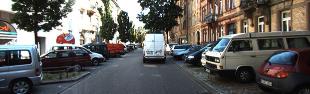} & 
	\includegraphics[width=\linewidth]{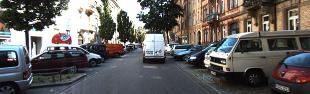} & 
	\includegraphics[width=\linewidth]{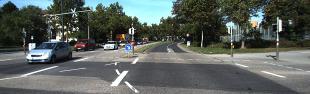} & 
	\includegraphics[width=\linewidth]{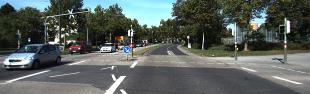} \\ \\[0.3em]
	& \textbf{Self-Mono-SF-ft (Ours)} & Mono-SF \cite{Brickwedde:2019:MSF} & \textbf{Self-Mono-SF-ft (Ours)} & Mono-SF \cite{Brickwedde:2019:MSF} \\

	\rotatebox[origin=c]{90}{D1} &
	\begin{tikzpicture}
    \node[inner sep=0] (img) {\includegraphics[width=\linewidth]{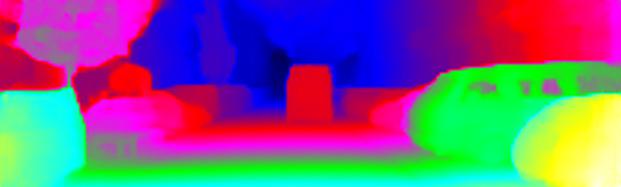}};
	\end{tikzpicture}&
	\begin{tikzpicture}
    \node[inner sep=0] (img) {\includegraphics[width=\linewidth]{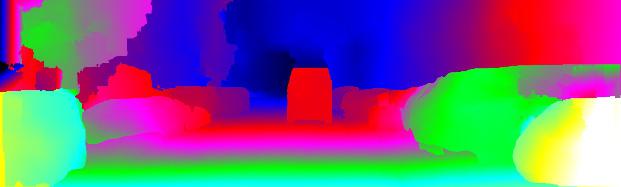}};
	\end{tikzpicture}&
	\begin{tikzpicture}
    \node[inner sep=0] (img) {\includegraphics[width=\linewidth]{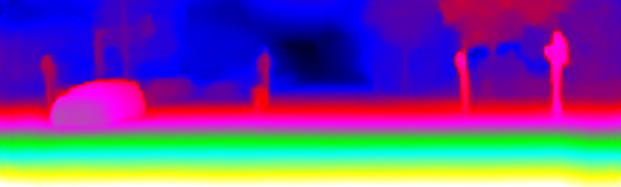}};
	\end{tikzpicture}&
	\begin{tikzpicture}
    \node[inner sep=0] (img) {\includegraphics[width=\linewidth]{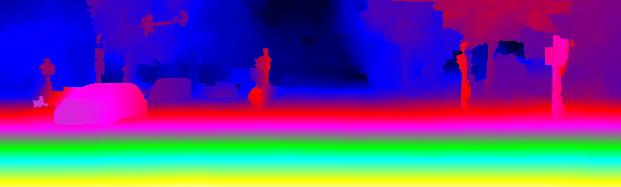}};
	\end{tikzpicture} \\

	\rotatebox[origin=c]{90}{D1 Error}  &
	\begin{tikzpicture}
    \node[inner sep=0] (img) {\includegraphics[width=\linewidth]{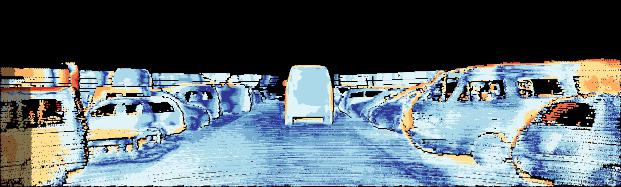}};
    \node[anchor=north east] at (img.north east){ \tiny \color{white} \textbf{11.38 \%}};
	\end{tikzpicture}&
	\begin{tikzpicture}
    \node[inner sep=0] (img) {\includegraphics[width=\linewidth]{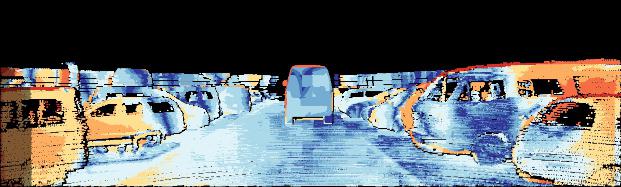}};
    \node[anchor=north east] at (img.north east){ \tiny \color{white} \textbf{20.88 \%}};
	\end{tikzpicture}&
	\begin{tikzpicture}
    \node[inner sep=0] (img) {\includegraphics[width=\linewidth]{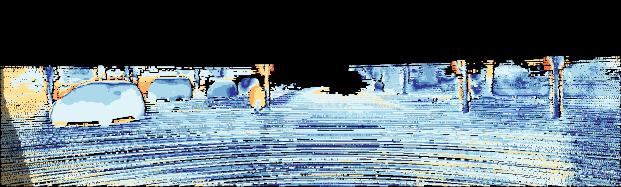}};
    \node[anchor=north east] at (img.north east){ \tiny \color{white} \textbf{6.90 \%}};
	\end{tikzpicture}&
	\begin{tikzpicture}
    \node[inner sep=0] (img) {\includegraphics[width=\linewidth]{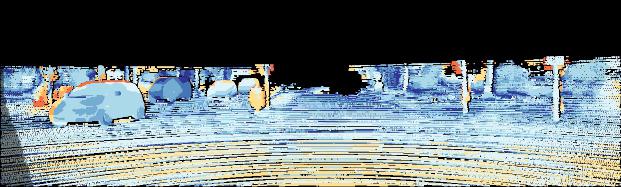}};
    \node[anchor=north east] at (img.north east){ \tiny \color{white} \textbf{13.31 \%}};
	\end{tikzpicture} \\
	
	\rotatebox[origin=c]{90}{D2} &
	\begin{tikzpicture}
    \node[inner sep=0] (img) {\includegraphics[width=\linewidth]{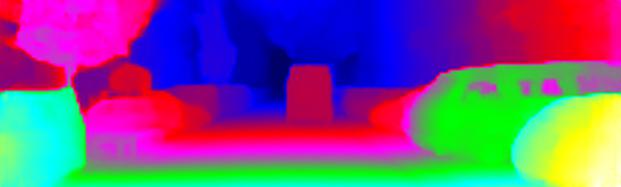}};
	\end{tikzpicture}&
	\begin{tikzpicture}
    \node[inner sep=0] (img) {\includegraphics[width=\linewidth]{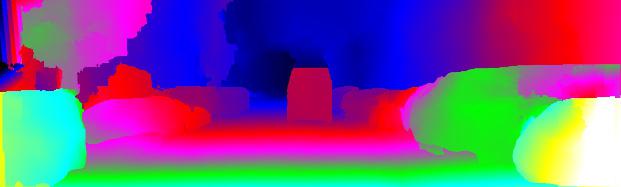}};
	\end{tikzpicture}&
	\begin{tikzpicture}
    \node[inner sep=0] (img) {\includegraphics[width=\linewidth]{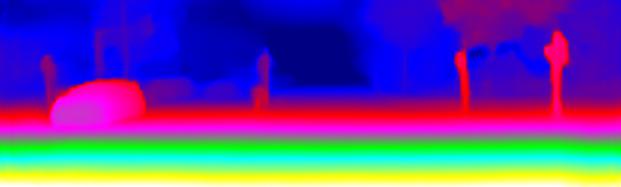}};
	\end{tikzpicture}&
	\begin{tikzpicture}
    \node[inner sep=0] (img) {\includegraphics[width=\linewidth]{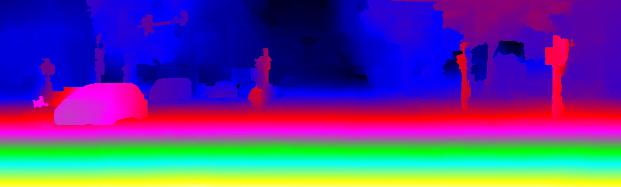}};
	\end{tikzpicture} \\

	\rotatebox[origin=c]{90}{D2 Error}  &
	\begin{tikzpicture}
    \node[inner sep=0] (img) {\includegraphics[width=\linewidth]{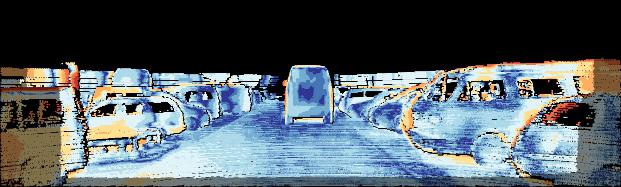}};
    \node[anchor=north east] at (img.north east){ \tiny \color{white} \textbf{13.56 \%}};
	\end{tikzpicture}&
	\begin{tikzpicture}
    \node[inner sep=0] (img) {\includegraphics[width=\linewidth]{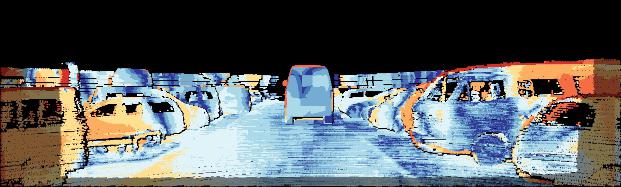}};
    \node[anchor=north east] at (img.north east){ \tiny \color{white} \textbf{22.75 \%}};
	\end{tikzpicture}&
	\begin{tikzpicture}
    \node[inner sep=0] (img) {\includegraphics[width=\linewidth]{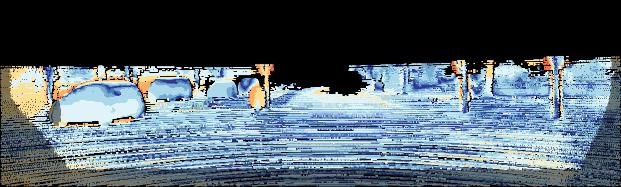}};
    \node[anchor=north east] at (img.north east){ \tiny \color{white} \textbf{7.62 \%}};
	\end{tikzpicture}&
	\begin{tikzpicture}
    \node[inner sep=0] (img) {\includegraphics[width=\linewidth]{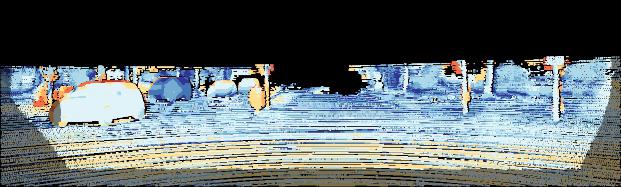}};
    \node[anchor=north east] at (img.north east){ \tiny \color{white} \textbf{16.72 \%}};
	\end{tikzpicture} \\

	\rotatebox[origin=c]{90}{F1} &
	\begin{tikzpicture}
    \node[inner sep=0] (img) {\includegraphics[width=\linewidth]{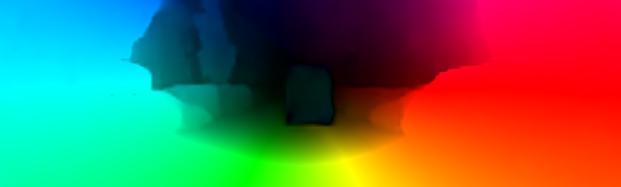}};
	\end{tikzpicture}&
	\begin{tikzpicture}
    \node[inner sep=0] (img) {\includegraphics[width=\linewidth]{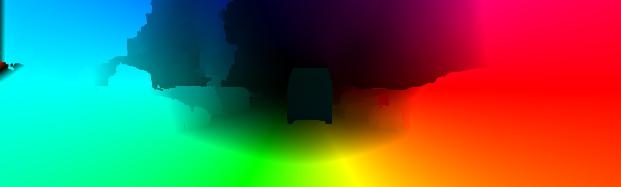}};
	\end{tikzpicture}&
	\begin{tikzpicture}
    \node[inner sep=0] (img) {\includegraphics[width=\linewidth]{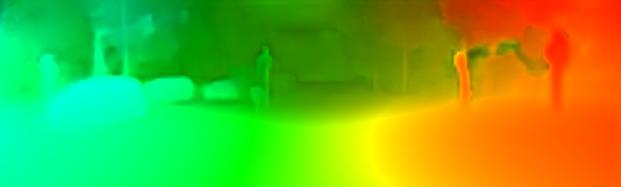}};
	\end{tikzpicture}&
	\begin{tikzpicture}
    \node[inner sep=0] (img) {\includegraphics[width=\linewidth]{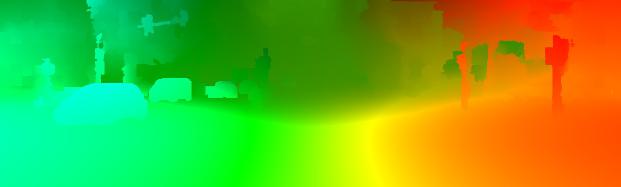}};
	\end{tikzpicture} \\

	\rotatebox[origin=c]{90}{F1 Error}  &
	\begin{tikzpicture}
    \node[inner sep=0] (img) {\includegraphics[width=\linewidth]{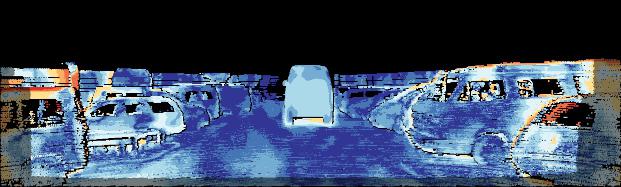}};
    \node[anchor=north east] at (img.north east){ \tiny \color{white} \textbf{4.92 \%}};
	\end{tikzpicture}&
	\begin{tikzpicture}
    \node[inner sep=0] (img) {\includegraphics[width=\linewidth]{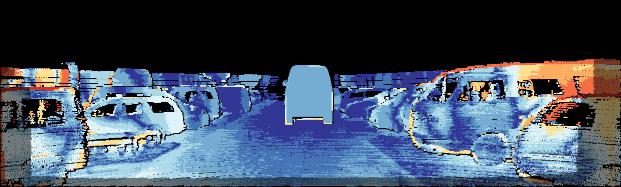}};
    \node[anchor=north east] at (img.north east){ \tiny \color{white} \textbf{8.54 \%}};
	\end{tikzpicture}&
	\begin{tikzpicture}
    \node[inner sep=0] (img) {\includegraphics[width=\linewidth]{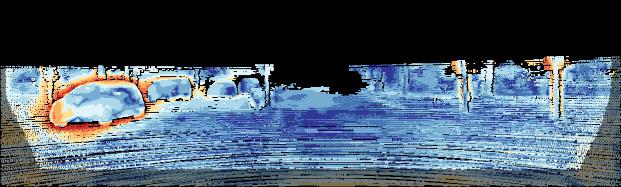}};
    \node[anchor=north east] at (img.north east){ \tiny \color{white} \textbf{8.89 \%}};
	\end{tikzpicture}&
	\begin{tikzpicture}
    \node[inner sep=0] (img) {\includegraphics[width=\linewidth]{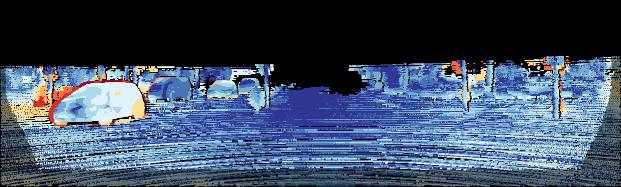}};
    \node[anchor=north east] at (img.north east){ \tiny \color{white} \textbf{5.68 \%}};
	\end{tikzpicture} \\

	\rotatebox[origin=c]{90}{SF1 Error}  &
	\begin{tikzpicture}
    \node[inner sep=0] (img) {\includegraphics[width=\linewidth]{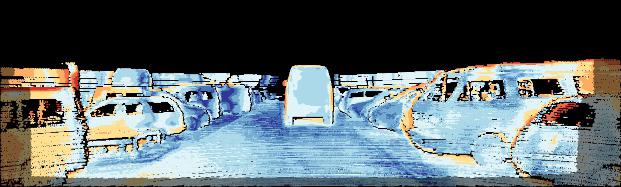}};
    \node[anchor=north east] at (img.north east){ \tiny \color{white} \textbf{14.63 \%}};
	\end{tikzpicture}&
	\begin{tikzpicture}
    \node[inner sep=0] (img) {\includegraphics[width=\linewidth]{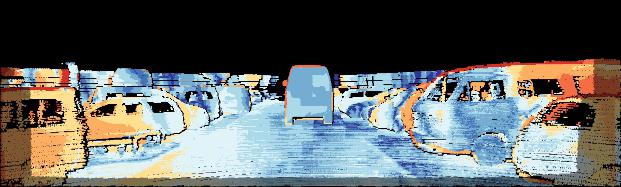}};
    \node[anchor=north east] at (img.north east){ \tiny \color{white} \textbf{23.31 \%}};
	\end{tikzpicture}&
	\begin{tikzpicture}
    \node[inner sep=0] (img) {\includegraphics[width=\linewidth]{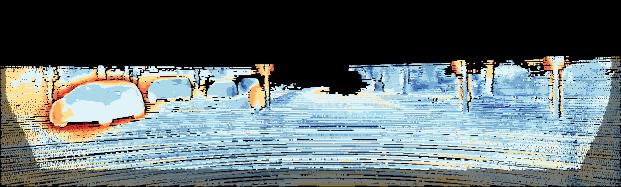}};
    \node[anchor=north east] at (img.north east){ \tiny \color{white} \textbf{13.80 \%}};
	\end{tikzpicture}&
	\begin{tikzpicture}
    \node[inner sep=0] (img) {\includegraphics[width=\linewidth]{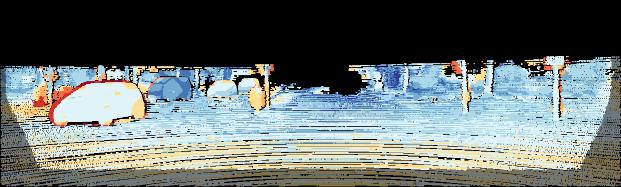}};
    \node[anchor=north east] at (img.north east){ \tiny \color{white} \textbf{19.29 \%}};
	\end{tikzpicture} \\ 

\end{tabular}
\caption {\textbf{Some successful cases and qualitative comparison with the state of the art on the KITTI 2015 Scene Flow public benchmark \cite{Menze:2015:J3E,Menze:2018:OSF}}. In the first row, we show two input images, the reference and target image. From the second to the last row, we give a qualitative comparison with Mono-SF \cite{Brickwedde:2019:MSF}: the disparity map of the reference image \emph{(D1)} with its error map \emph{(D1 Error)}, disparity estimation at the target image mapped into the reference frame \emph{(D2)} along with its error map \emph{(D2 Error)}, optical flow \emph{(F1)} with its error map \emph{(F1 Error)}, and the scene flow error map \emph{(SF1 Error)}. The outlier rates are overlayed on each error map.} 
\label{fig:qualitative_successful}
\end{figure*}
}

{
\begin{figure*}[!b]
\centering
\footnotesize
\setlength\tabcolsep{0.3pt}
\renewcommand{\arraystretch}{0.2}
\begin{tabular}{c@{\hskip 0.5em} >{\centering\arraybackslash}m{.24\textwidth} >{\centering\arraybackslash}m{.24\textwidth} @{\hskip 0.4em} >{\centering\arraybackslash}m{.24\textwidth} >{\centering\arraybackslash}m{.24\textwidth}}
	& Reference image & Target image & Reference image & Target image \\ 
	& \includegraphics[width=\linewidth]{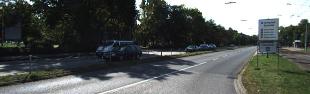} & 
	\includegraphics[width=\linewidth]{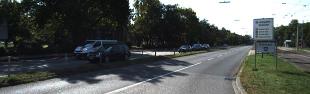} & 
	\includegraphics[width=\linewidth]{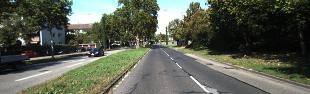} & 
	\includegraphics[width=\linewidth]{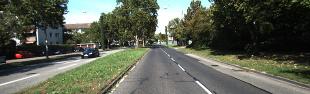} \\ \\[0.3em]
	& \textbf{Self-Mono-SF-ft (Ours)} & Mono-SF \cite{Brickwedde:2019:MSF} & \textbf{Self-Mono-SF-ft (Ours)} & Mono-SF \cite{Brickwedde:2019:MSF} \\

	\rotatebox[origin=c]{90}{D1} &
	\begin{tikzpicture}
    \node[inner sep=0] (img) {\includegraphics[width=\linewidth]{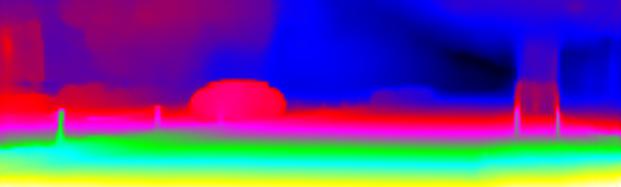}};
	\end{tikzpicture}&
	\begin{tikzpicture}
    \node[inner sep=0] (img) {\includegraphics[width=\linewidth]{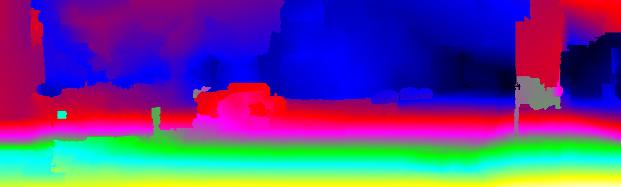}};
	\end{tikzpicture}&
	\begin{tikzpicture}
    \node[inner sep=0] (img) {\includegraphics[width=\linewidth]{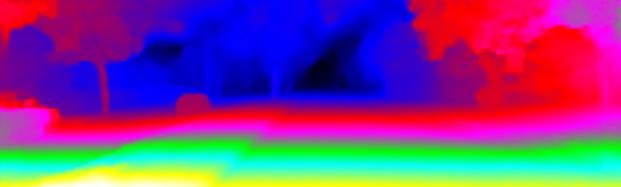}};
	\end{tikzpicture}&
	\begin{tikzpicture}
    \node[inner sep=0] (img) {\includegraphics[width=\linewidth]{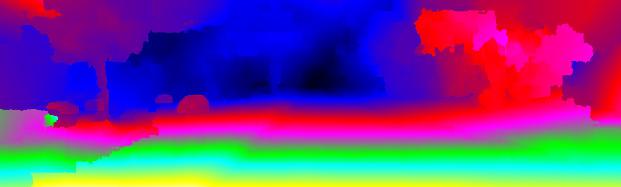}};
	\end{tikzpicture} \\

	\rotatebox[origin=c]{90}{D1 Error}  &
	\begin{tikzpicture}
    \node[inner sep=0] (img) {\includegraphics[width=\linewidth]{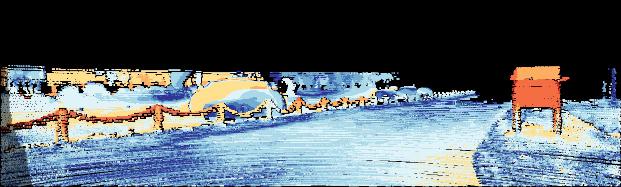}};
    \node[anchor=north east] at (img.north east){ \tiny \color{white} \textbf{14.41 \%}};
	\end{tikzpicture}&
	\begin{tikzpicture}
    \node[inner sep=0] (img) {\includegraphics[width=\linewidth]{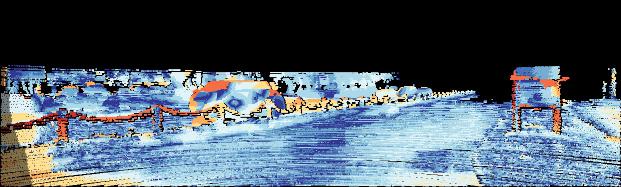}};
    \node[anchor=north east] at (img.north east){ \tiny \color{white} \textbf{9.08 \%}};
	\end{tikzpicture}&
	\begin{tikzpicture}
    \node[inner sep=0] (img) {\includegraphics[width=\linewidth]{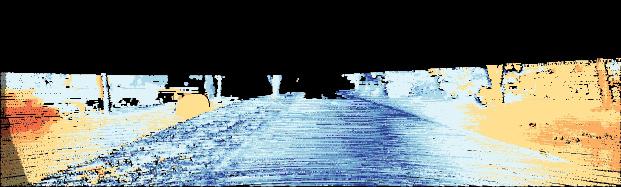}};
    \node[anchor=north east] at (img.north east){ \tiny \color{white} \textbf{27.64 \%}};
	\end{tikzpicture}&
	\begin{tikzpicture}
    \node[inner sep=0] (img) {\includegraphics[width=\linewidth]{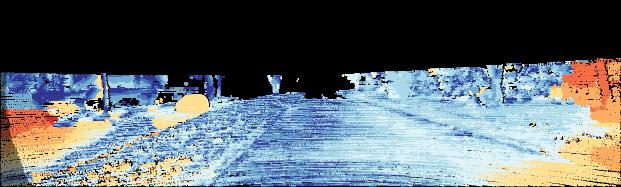}};
    \node[anchor=north east] at (img.north east){ \tiny \color{white} \textbf{15.90 \%}};
	\end{tikzpicture} \\
	
	\rotatebox[origin=c]{90}{D2} &
	\begin{tikzpicture}
    \node[inner sep=0] (img) {\includegraphics[width=\linewidth]{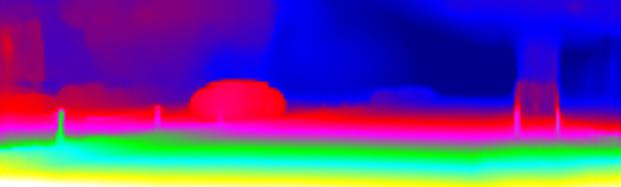}};    
	\end{tikzpicture}&
	\begin{tikzpicture}
    \node[inner sep=0] (img) {\includegraphics[width=\linewidth]{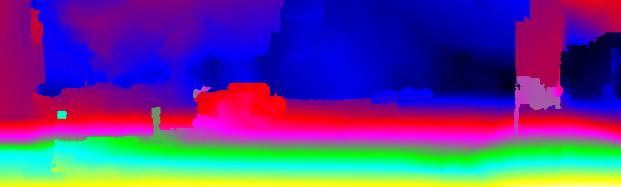}};
	\end{tikzpicture}&
	\begin{tikzpicture}
    \node[inner sep=0] (img) {\includegraphics[width=\linewidth]{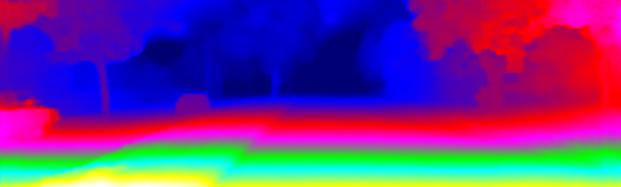}};
	\end{tikzpicture}&
	\begin{tikzpicture}
    \node[inner sep=0] (img) {\includegraphics[width=\linewidth]{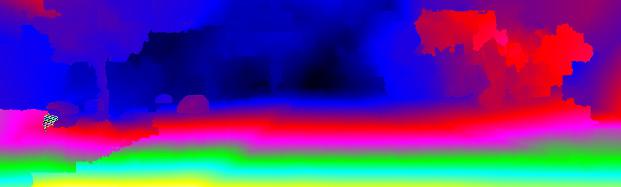}};
	\end{tikzpicture} \\

	\rotatebox[origin=c]{90}{D2 Error}  &
	\begin{tikzpicture}
    \node[inner sep=0] (img) {\includegraphics[width=\linewidth]{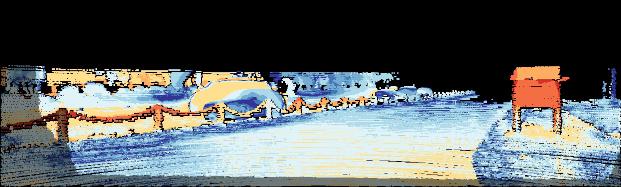}};
    \node[anchor=north east] at (img.north east){ \tiny \color{white} \textbf{20.42 \%}};
	\end{tikzpicture}&
	\begin{tikzpicture}
    \node[inner sep=0] (img) {\includegraphics[width=\linewidth]{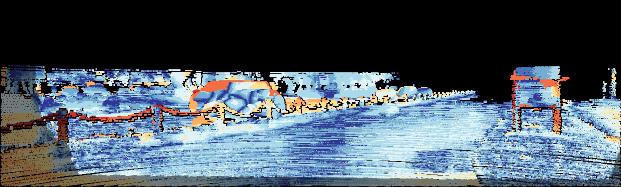}};
    \node[anchor=north east] at (img.north east){ \tiny \color{white} \textbf{10.02 \%}};
	\end{tikzpicture}&
	\begin{tikzpicture}
    \node[inner sep=0] (img) {\includegraphics[width=\linewidth]{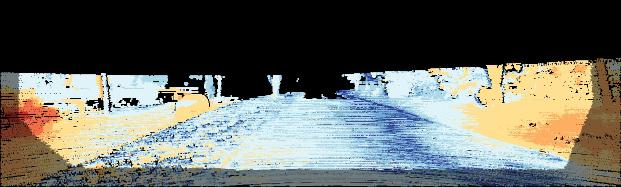}};
    \node[anchor=north east] at (img.north east){ \tiny \color{white} \textbf{29.30 \%}};
	\end{tikzpicture}&
	\begin{tikzpicture}
    \node[inner sep=0] (img) {\includegraphics[width=\linewidth]{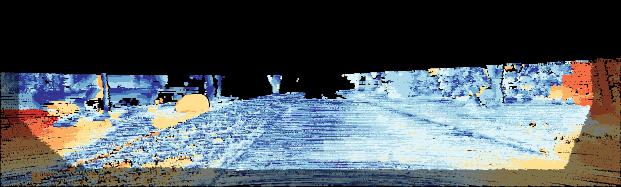}};
    \node[anchor=north east] at (img.north east){ \tiny \color{white} \textbf{18.90 \%}};
	\end{tikzpicture} \\

	\rotatebox[origin=c]{90}{F1} &
	\begin{tikzpicture}
    \node[inner sep=0] (img) {\includegraphics[width=\linewidth]{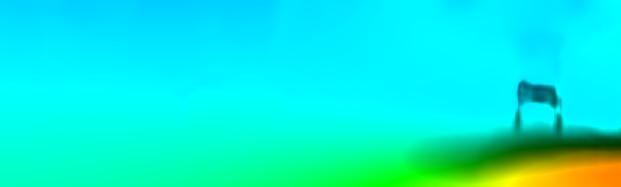}};
	\end{tikzpicture}&
	\begin{tikzpicture}
    \node[inner sep=0] (img) {\includegraphics[width=\linewidth]{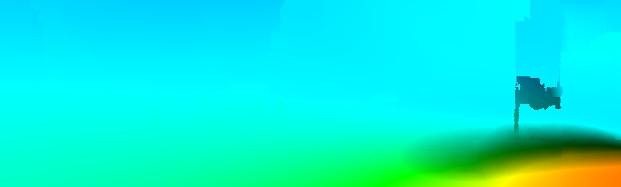}};
	\end{tikzpicture}&
	\begin{tikzpicture}
    \node[inner sep=0] (img) {\includegraphics[width=\linewidth]{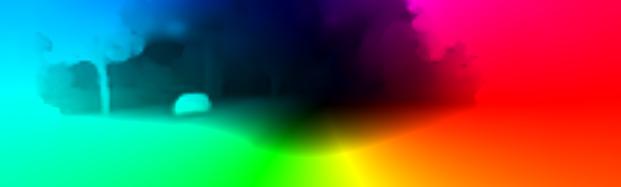}};
	\end{tikzpicture}&
	\begin{tikzpicture}
    \node[inner sep=0] (img) {\includegraphics[width=\linewidth]{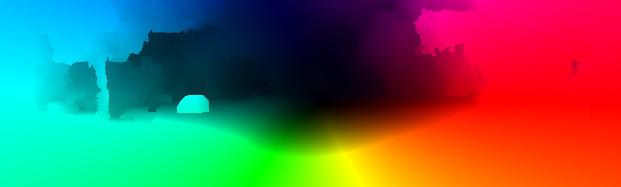}};
	\end{tikzpicture} \\

	\rotatebox[origin=c]{90}{F1 Error}  &
	\begin{tikzpicture}
    \node[inner sep=0] (img) {\includegraphics[width=\linewidth]{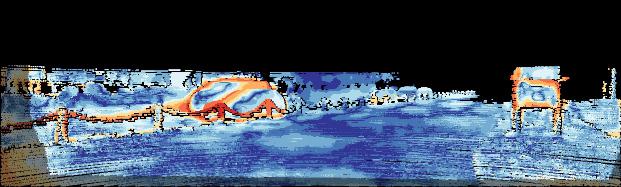}};
    \node[anchor=north east] at (img.north east){ \tiny \color{white} \textbf{8.41 \%}};
	\end{tikzpicture}&
	\begin{tikzpicture}
    \node[inner sep=0] (img) {\includegraphics[width=\linewidth]{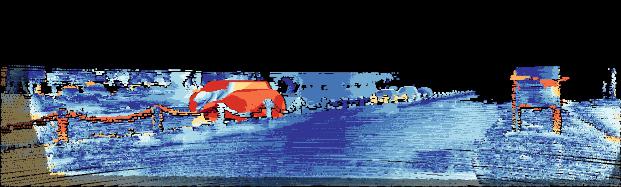}};
    \node[anchor=north east] at (img.north east){ \tiny \color{white} \textbf{12.32 \%}};
	\end{tikzpicture}&
	\begin{tikzpicture}
    \node[inner sep=0] (img) {\includegraphics[width=\linewidth]{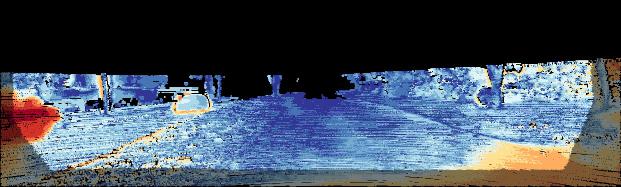}};
    \node[anchor=north east] at (img.north east){ \tiny \color{white} \textbf{13.24 \%}};
	\end{tikzpicture}&
	\begin{tikzpicture}
    \node[inner sep=0] (img) {\includegraphics[width=\linewidth]{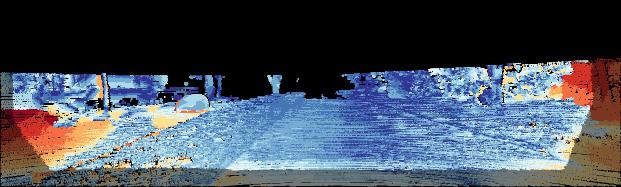}};
    \node[anchor=north east] at (img.north east){ \tiny \color{white} \textbf{13.86 \%}};
	\end{tikzpicture} \\

	\rotatebox[origin=c]{90}{SF1 Error}  &
	\begin{tikzpicture}
    \node[inner sep=0] (img) {\includegraphics[width=\linewidth]{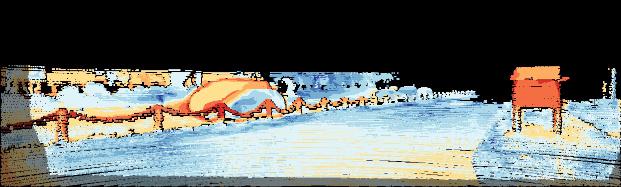}};
    \node[anchor=north east] at (img.north east){ \tiny \color{white} \textbf{25.25 \%}};
	\end{tikzpicture}&
	\begin{tikzpicture}
    \node[inner sep=0] (img) {\includegraphics[width=\linewidth]{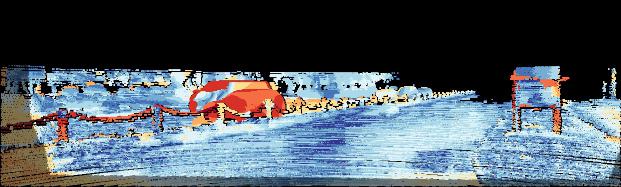}};
    \node[anchor=north east] at (img.north east){ \tiny \color{white} \textbf{15.60 \%}};
	\end{tikzpicture}&
	\begin{tikzpicture}
    \node[inner sep=0] (img) {\includegraphics[width=\linewidth]{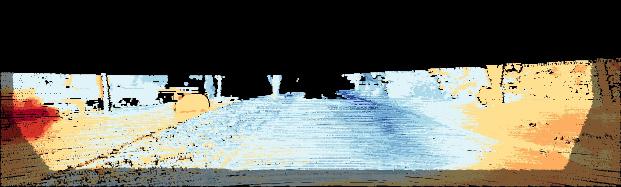}};
    \node[anchor=north east] at (img.north east){ \tiny \color{white} \textbf{37.65 \%}};
	\end{tikzpicture}&
	\begin{tikzpicture}
    \node[inner sep=0] (img) {\includegraphics[width=\linewidth]{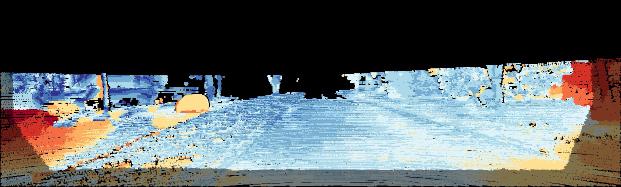}};
    \node[anchor=north east] at (img.north east){ \tiny \color{white} \textbf{20.33 \%}};
	\end{tikzpicture} \\ 
	
\end{tabular}
\caption {\textbf{Failure cases and qualitative comparison with the state of the art on the KITTI 2015 Scene Flow public benchmark \cite{Menze:2015:J3E,Menze:2018:OSF}}. In the first row, we show two input images, the reference and target image. From the second to the last row, we give a qualitative comparison with Mono-SF \cite{Brickwedde:2019:MSF}: the disparity map of the reference image \emph{(D1)} with its error map \emph{(D1 Error)}, disparity estimation at the target image mapped into the reference frame \emph{(D2)} with its error map \emph{(D2 Error)}, optical flow \emph{(F1)} with its error map \emph{(F1 Error)}, and the scene flow error map \emph{(SF1 Error)}. The outlier rates are overlayed on each error map.} 
\label{fig:qualitative_unsuccessful}
\end{figure*}
}

{\small

\let\oldthebibliography=\thebibliography
\let\oldendthebibliography=\endthebibliography
\renewenvironment{thebibliography}[1]{%
     \oldthebibliography{#1}%
     \setcounter{enumiv}{ 64 }%
}{\oldendthebibliography}

\bibliographystyle{ieee_fullname}

}

\end{document}